\documentclass{article} % For LaTeX2e
\usepackage{iclr2026_conference,times}

\usepackage{amsmath,amsfonts,bm}

\def\eqref#1{equation~\ref{#1}}
\def\1{\bm{1}}

\DeclareMathAlphabet{\mathsfit}{\encodingdefault}{\sfdefault}{m}{sl}
\SetMathAlphabet{\mathsfit}{bold}{\encodingdefault}{\sfdefault}{bx}{n}

\usepackage{hyperref}
\usepackage{url}
\usepackage{booktabs}
\usepackage{tabularx}
\usepackage{tcolorbox}
\usepackage{caption}
\usepackage{multirow}
\usepackage{adjustbox}
\usepackage{xcolor}
\usepackage{graphicx}
\usepackage{subcaption}
\usepackage{wrapfig}
\usepackage{mdframed}
\usepackage{enumitem}
\usepackage{amssymb}

\iclrpreprint
\title{On Code-Induced Reasoning in LLMs}

\author{
  Abdul Waheed\thanks{Equal contribution}\,
  ~~Zhen Wu\footnotemark[1]\,
  ~~Carolyn Ros\'e\,
  ~~Daphne Ippolito \\
  Carnegie Mellon University \\
  \texttt{\normalsize \{abdulw,zhenwu,cprose,dippolit\}@cs.cmu.edu}
}

\begin{document}

\maketitle

\begin{abstract}
% Code data has been shown to enhance the reasoning capabilities of large language models (LLMs), yet it remains unclear why this effect occurs and which aspects of code are most responsible. We investigate this question from a data-centric perspective by introducing \textbf{code-induced reasoning} as the improvement in reasoning performance attributable to the code training data. We systematically construct and perturb code training data to isolate structural and semantic factors, fine-tune LLMs on each variant, and evaluate them across general, code reasoning, and mathematical reasoning tasks. We find that [\textcolor{red}{FINDINGS GOES HERE}]. Through this, we aim to provide insight into the inductive biases introduced by code and inform the design of training data for enhancing reasoning capabilities.

% Daphne's proposed version. I don't think it's necessary to introduce a new term in bold (code-induced reasoning) since this isn't a novel method you are coming up with.
Code data has been shown to enhance the reasoning capabilities of large language models (LLMs), but it remains unclear which aspects of code are most responsible. We investigate this question with a systematic, data-centric framework. We construct parallel instruction datasets in ten programming languages and apply controlled perturbations that selectively disrupt structural or semantic properties of code. We then finetune LLMs from five model families and eight scales on each variant and evaluate their performance on natural language, math, and code tasks. Across 3,331 experiments, our results show that LLMs are more vulnerable to structural perturbations than semantic ones, particularly on math and code tasks. Appropriate abstractions like pseudocode and flowcharts can be as effective as code, while encoding the same information with fewer tokens without adhering to original syntax can often retain or even improve performance. Remarkably, even corrupted code with misleading signals remains competitive when surface-level regularities persist. Finally, syntactic styles also shape task-specific gains with Python favoring natural language reasoning and lower-level languages such as Java and Rust favoring math. Through our systematic framework, we aim to provide insight into how different properties of code influence reasoning and inform the design of training data for enhancing LLM reasoning capabilities.

\end{abstract}

% 1. Introduction
\section{Introduction}\label{sec:introduction}

There has been substantial interest in the last several years in engineering language models that can tackle challenging reasoning tasks \citep{huang-chang-2023-towards}.
Language reasoning tasks, such as math word problems or logic puzzles, tend to require multi-step, structured ``thinking'' in order to produce the correct answer.
Recent work has found that training the language model on code, either during pre-training~\citep{fu2022gptroadmap, ma2023training} or during post-training~\citep{zhang2024unveilingimpactcodingdata}, can improve its skill at reasoning tasks, even ones that are unrelated to programming.
These prior works have hypothesized that the properties of code data, such as its logical consistency, compositional structure, and reduced ambiguity compared to natural language, provide effective signals that benefit reasoning. Despite the broad effectiveness of code data in training, we still lack a systematic understanding of which aspects of code drive these improvements: is it the its syntactic regularity, structural abstractions, or linguistic styles?

In this work, we aim to provide such an account by systematically investigating which aspects of code serve as effective training signals. To this end, we construct parallel instruction datasets in both natural language and code, and further expand the code dataset into language-specific variants by generating responses in ten widely used programming languages. This design allows us to examine how structural differences across languages affect downstream reasoning. In addition, we introduce controlled perturbations to the code data to isolate contributing factors: (1) \emph{rule-based} transformations such as whitespace removal or comment shuffling, and (2) \emph{generative} transformations where GPT-4o-mini rewrites or reformats the code (e.g., with augmented comments, pseudocode, or flowcharts). We then fine-tune language models on each dataset variant, and evaluate them across natural language and general knowledge, math, as well as code understanding and generation tasks. Our contributions are:
\begin{itemize}
    \item We introduce a systematic framework to disentangle what aspects of code data improve reasoning, combining parallel instruction data construction, controlled perturbations, and large-scale evaluation across five model families and eight scales.
    \item We design a comprehensive and controlled suite of perturbations spanning rule-based edits and generative rewritings.
    \item We provide new insights into the role of code in reasoning to inspire guidance on leveraging its structural and linguistic properties in future training data design.
\end{itemize}

\section{Related Work}\label{sec:related_work}

\paragraph{Code data for LLM reasoning} Recent work has increasingly demonstrated that incorporating code data can substantially improve the reasoning abilities of LLMs. Prior studies show that adding code during pretraining or instruction tuning consistently improves model performance across reasoning tasks, domains, model scales and architectures \citep{Ma2023-zn, Zhang2024-bh, Yang2025-ks, Aryabumi2024ToCO}. Several works further explore the synergy between code and reasoning and highlight how code’s structured and verifiable properties support logical decomposition and intermediate step generation \citep{BiUnknown-hm, Yang2024-ks}. This effect has been observed in multilingual contexts as well, where code-augmented training improves structured reasoning in under-resourced languages \citep{li-etal-2024-eliciting-better}. Complementary research focuses on code’s impact for alignment and reward modeling, where pretraining with code-preference pairs or code-based intermediate steps can improve model calibration for reasoning-intensive tasks \citep{Yu2024-pc}. The closest line of research to our work explores stress-testing LLMs with structural and semantic code perturbations \citep{Lam2025-ck}, which shows that small corruptions can significantly reduce reasoning performance.

\paragraph{Data impact on LLM performance}

The performance of LLMs are tied to the vast amounts of training data, but the quality, composition, and characteristics of this data greatly shape their abilities~\citep{wang2024survey, li2023quantity, lee2022deduplicatingtrainingdatamakes}. For example, extensive analyses by \cite{longpre-etal-2024-pretrainers} have shown that pretraining data curation decisions for dataset age, composition, and content filtering have systematic impact
on downstream performance, and that these effects persist even after fine-tuning steps. \cite{zhang2024persistentpretrainingpoisoningllms} demonstrate that poisoning as little as 0.1\% (and even 0.001\%) can produce persistent behavioral changes that survive instruction tuning and alignment. In addition, \cite{havrilla2024understandingeffectnoisellm} showed that LLMs are sensitive to global, accumulative errors in chain-of-thought-structured training data, and that it is critical to filter out documents
containing large amounts of dynamic, global noise during
both pretraining and fine-tuning.

% 3. Methodology
\section{Methodology}\label{sec:methodology}

\begin{figure}[t]
  \centering
  \includegraphics[width=0.8\linewidth]{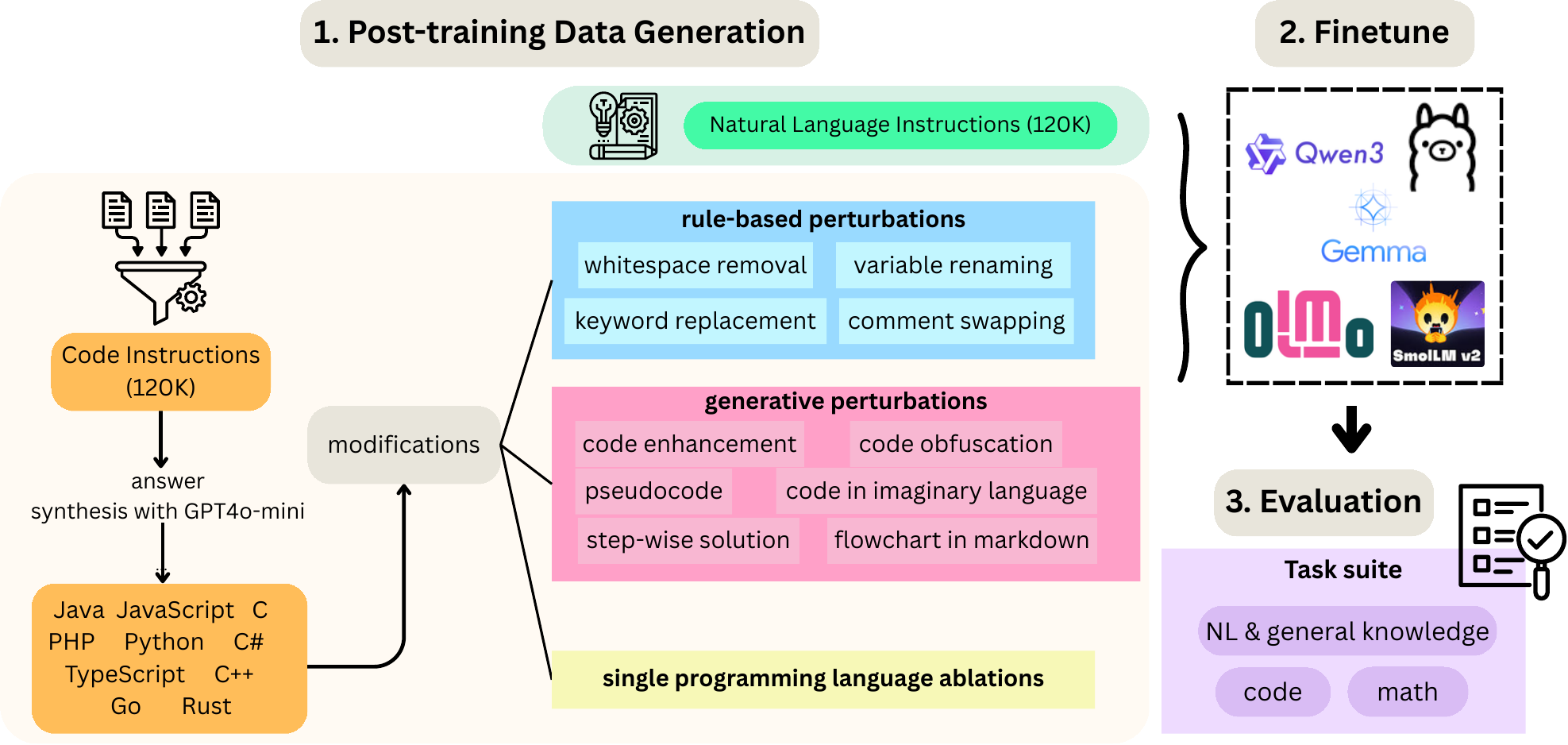}
  \caption{We construct parallel code and natural language instruction datasets, apply targeted modifications (rule-based and generative-based perturbations, single programming language ablations), and fine-tune a separate LLM on each modified dataset. We then evaluate the resulting models across general natural language, code, and math reasoning tasks.}
  \label{fig:framework}
\end{figure}

We design a controlled experimental framework to understand what aspects of code improve reasoning in language models. Our methodology consists of three stages: constructing parallel natural language and code instruction datasets (Section \ref{data_generation}); applying systematic modifications to code instruction data (Section \ref{sec:perturbation_design}); and fine-tuning various language models on each dataset variant and then conducting evaluation (Section \ref{sec:model_training}). An overview of this framework is shown in Figure~\ref{fig:framework}.

\subsection{Instruction Data Generation} \label{data_generation}

We construct two parallel instruction datasets: one in natural language and the other in code, each containing 120{,}000 instruction-response pairs. We collect instructions from publicly available datasets, carefully process and filter them through deduplication and language-agnostic filtering, and augment the code data in a controlled way. This construction enables a more controlled comparison of natural- and code-based instruction following under a unified training framework.

% and then standardized through deduplication, language-agnostic filtering, and controlled augmentation. 

\paragraph{Code instructions}
We aggregate code instructions from Codeforces-CoT~\citep{penedo2025codeforces}, Code-Instruction-122K~\citep{tokenbender2024codeinstructions}, Evol-Instruct-Code-80k-v1~\citep{nickrosh2024evolinstructcode}, CodeInstruction~\citep{red1xe2023codeinstructions}, Code-Instruct-Sets~\citep{atlasunified2023codeinstructsets}, and Code-Instruct-Alpaca-Vicuna-WizardLM~\citep{rombodawg2024codeinstructalpaca}.
We aim to construct instruction data that is high-quality, diverse, and language-agnostic.
% Since the original datasets typically include both instructions and reference answers, we retain only the instructions and later generate our own standardized responses in multiple formats, as described below.

To ensure generality and eliminate redundancy, we first remove all exact-match duplicates across the datasets. We then filter out instructions that are explicitly programming-language-specific (e.g., ``Translate this code from Python to java'') or whose solutions are inherently tied to particular domains, such as web development or databases (e.g., ``webpage'', ``website'', ``SQL'', ``HTML''). 

% Through this process, we obtain a high-quality set of general-purpose instructions that are agnostic to any specific programming language.

For each instruction, we prompt GPT-4o-mini\footnote{Responses are generated with temperature 0.6 and API-default decoding parameters.} to generate answers in ten widely used programming languages: Java, JavaScript, PHP, Python, C\#, TypeScript, C, C++, Go, and Rust. To create these variants, we design 20 language specification templates that explicitly request a solution in a given programming language (Table~\ref{tab:code_instruction_suffix}). For every instruction, we randomly select a template, instantiate it with one of the target languages, and combine it with the general generation instructions to form a complete prompt (Figure~\ref{fig:code_gen_prompt}). From these generations, we sample 120K instruction–response pairs with valid outputs, evenly distributed across all ten languages.

\paragraph{Natural language instructions}
We sample 120K examples from the OpenHermes 2.5 corpus~\citep{teknium2023openhermes}.
We exclude instruction-response pairs associated with categories unrelated to general-purpose instruction following, such as ``agent'' and ``summarization'', as well as those labeled ``coding'' to ensure the dataset is entirely natural language.
To maintain linguistic consistency, we further filter out non-English examples.
This filtered natural language subset complements our code instruction data, enabling a fair comparison between code and natural language instructions.

\subsection{Systematic Perturbation Design} \label{sec:perturbation_design}
% requires: \usepackage{multirow,booktabs,adjustbox}
\begin{table*}[t]
\centering
\small
\renewcommand{\arraystretch}{1.15}
\caption{An example of perturbations (Section~\ref{sec:perturbation_design}) applied to the same original snippet.}
\label{tab:petya_all_perturbations}
\begin{adjustbox}{width=\textwidth}
\begin{tabular}{p{6.0cm} p{1.5cm} p{3.5cm} p{3.8cm} p{5.5cm}}
\toprule
\textbf{Full Original Snippet} & \textbf{Type} & \textbf{Strategy} & \textbf{Original Excerpt} & \textbf{Perturbed Excerpt} \\
\midrule

\multirow{14}{=}{\parbox[t]{6.0cm}{\ttfamily
def process\_string(input\_string): \\
\hspace*{1em}vowels = "aoyeuiAOYEUI" \\
\hspace*{1em}result = [] \\
\\
\hspace*{1em}for char in input\_string: \\
\hspace*{2em}if char not in vowels: \\
\hspace*{3em}result.append('.' + char.lower()) \\
\\
\hspace*{1em}return ''.join(result) \\
\\
\# Read input \\
input\_string = input().strip() \\
\# Process and print the result \\
print(process\_string(input\_string))
}}

& \multirow[c]{7}{*}{Rule-based}
& Whitespace Removal
& \ttfamily result.append('.' + char.lower())
& \ttfamily result.append('.'+char.lower()) \\
\cmidrule(lr){3-5}

&
& Variable Renaming
& \ttfamily for char in input\_string: \ldots
& \ttfamily for var\_4 in var\_1: if var\_4 not in var\_2: \ldots \\
\cmidrule(lr){3-5}

&
& Keyword Replacement (Nonsense)
& \ttfamily if char not in vowels:
& \ttfamily garply i not in baz \\
\cmidrule(lr){3-5}

&
& Keyword Replacement (Non-English)
& \ttfamily for char in input\_string:
& \ttfamily para ch en entrada \\
\cmidrule(lr){3-5}

&
& Comment Swapping (Local)
& \ttfamily \# Read input
& \ttfamily \# Walking \\
\cmidrule(lr){3-5}

&
& Comment Swapping (Global)
& \ttfamily \# Process and print the result
& \ttfamily // Queue for processing nodes \\
\cmidrule(lr){3-5}

&
& Comment Removal
& \ttfamily \# Read input
& \ttfamily /* all comments removed */ \\
\cmidrule(lr){2-5}

& \multirow[c]{6}{*}{Generative}
& Pseudocode
& \ttfamily for char in input\_string: if char not in vowels
& \textit{FOR EACH character IF not vowel THEN append '.'+lowercase} \\
\cmidrule(lr){3-5}

&
& Step-by-Step
& \ttfamily result.append('.' + char.lower())
& \textit{Append '.' before consonants and convert to lowercase} \\
\cmidrule(lr){3-5}

&
& Flowchart
& \ttfamily if char not in vowels:
& \texttt{[Read char] → \{Vowel?\} → [Append '.'+lower]} \\
\cmidrule(lr){3-5}

&
& Code in Imaginary Language
& \ttfamily result.append('.' + char.lower())
& \textit{glorf add '.' $\oplus$ lower(chr)} \\
\cmidrule(lr){3-5}

&
& Comment Enhancement
& \ttfamily \# Process and print the result
& \textit{\# Removes vowels and prefixes consonants with '.'} \\
\cmidrule(lr){3-5}

&
& Comment Obfuscation
& \ttfamily \# Read input
& \textit{\# WARNING: Code may summon aliens; \# TODO: handle quantum vowels} \\

\bottomrule
\end{tabular}
\end{adjustbox}
\end{table*}

To understand which specific structural and semantic properties are responsible for changes in reasoning task performances, we systematically perturb different aspects of the code dataset.
We design the perturbations through two ways: \emph{rule-based} (deterministic transformations) and \emph{generative} (model-generated augmentations).
Notably, our perturbation strategies do not alter the number of examples in the dataset.
We illustrate an examples of these perturbations in Table~\ref{tab:petya_all_perturbations}.
% We also experiment with filtering the dataset to only one programming language.

% rule-based
% whitespace removal
% variable renaming to var\_{num}
% keywords replacement (nonsense english, real non-english)
% whithin snippet and across snippets comment swaps

\subsubsection{Rule-Based Perturbations}
Rule-based perturbations apply deterministic transformations to the code. They are designed to disrupt superficial patterns or semantic signals that may influence model predictions without altering the core logic of the code. We describe five such perturbations below:

\textbf{Whitespace removal}
All whitespace characters are removed from the code. This tests whether models rely on formatting heuristics, such as indentation or visual grouping of blocks, as implicit structural cues, particularly in languages like Python where whitespace is semantically meaningful.

\textbf{Variable renaming}
We replace user-defined variables, function names, and class names with canonical placeholders of the form \texttt{var\_i}, where \( i \in \left[0, n\right) \) and \( n \) is the total number of unique identifiers in the code snippet. This removes semantic cues conveyed by meaningful identifier names (e.g., \texttt{counter}, \texttt{isSorted}).

\textbf{Programming language keyword replacement}
For each of the ten programming languages in our dataset, we identify its reserved keywords (e.g., \texttt{if}, \texttt{return}, \texttt{def} in Python) and substitute all occurrences of them using two strategies. The first replaces keywords with nonsense tokens (e.g., \textit{foo}, \textit{quux}), which have no semantic meaning in any language. In the second strategy, we use non-English but valid words (e.g., \textit{amigo}, \textit{fleur}), which are real words in various languages but semantically unrelated to the programming context. These perturbations aim to challenge models’ reliance on syntactic and semantic cues from familiar language constructs.

\textbf{Comment removal}
We remove all inline and block comments from each code snippet. Code comments often provide useful semantic signals for program comprehension~\citep{buse2009learning,de2005study}. This perturbation tests whether models largely leverage such auxiliary natural-language cues.

\textbf{Comment swapping}
We introduce local and global swapping that misplace code comments to disrupt the semantic alignment between code and documentation. In local swapping, comments within a snippet are randomly reordered, preserving their content but misaligning them with the relevant code segments. In global swapping, we first collect a global pool of comments from the entire dataset. Then, for each comment in a snippet, we replace it with a randomly sampled comment from this pool. This results in documentation that is entirely mismatched to the surrounding code.

\subsubsection{Generative Perturbations}

We create generative perturbations by prompting GPT-4o-mini\footnote{We use temperature of 0.6 and default settings.} to produce alternative versions of code responses generated according to Section~\ref{data_generation}. These rewrites preserve the original intent of the code while introducing more diverse variations beyond what rule-based edits can achieve, allowing us to test model sensitivity and robustness to semantically equivalent inputs expressed in different forms.  The full set of prompts used is available in Appendix~\ref{appsubsec:prompts}.

\textbf{Comment enhancement} We prompt GPT-4o-mini to regenerate the code with high-quality documentation and inline comments (Figure~\ref{fig:good_code_prompt}). The prompt emphasizes two forms of annotation: (1) comprehensive documentation comments for all functions, classes, and key code blocks to describe their purpose, parameters, return values, and assumptions; and (2) informative inline comments that clarify complex or non-obvious logic. These annotations follow the conventions of the target programming language (e.g., Python docstrings, JavaDoc). Unlike the often sparse comments in unperturbed data, the enhanced versions provide consistent, high-quality annotations, which enables us to test the effect of documentation quality on model performance.

\textbf{Comment obfuscation} Here, we generate deliberately misleading, irrelevant, or nonsensical comments, while preserving the code’s functionality (Figure~\ref{fig:bad_code_prompt}). These include (1) inaccurate, off-topic, or absurd documentation (e.g., references to astrology, cooking, or fictitious technologies) and (2) chaotic inline comments that contradict the code's functionality, reference imaginary bugs or features, and use distracting styles such as ALL-CAPS, emojis, and fabricated jargon. This perturbation tests model robustness to extreme noise and deceptive annotations.

\textbf{Pseudocode}
We convert code into high-level pseudocode while preserving its logical structure (Figure~\ref{fig:pseudocode_prompt}). The model is instructed to replace language-specific syntax with pseudocode constructs (e.g., \texttt{IF...THEN...ENDIF}, \texttt{FOR EACH}, etc.), remove low-level implementation details (e.g., type declarations or library calls), and maintain the original control flow and indentation. 
This perturbation evaluates whether models can reason over algorithmic intent without relying on concrete syntax, which offers insight into generalization across abstraction layers in code representation.

\textbf{Flowchart in Markdown} 
We generate a control flow diagram using Mermaid syntax in Markdown for a given code snippet (Figure~\ref{fig:flowchart_prompt}). The diagram captures all major control structures, such as loops, branches, function calls, and return points, using minimal but descriptive labels. This transformation renders executable code as a graphical abstraction, allowing us to understand whether models can reason over symbolic control flow and align it with underlying program semantics.

\textbf{Step-by-step solution}
We rewrite code as a numbered list of natural language steps (Figure~\ref{fig:step_by_step_prompt}). Each step preserves the program’s logic and execution order but uses declarative, language-agnostic phrasing (e.g., ``Define a function named...'', ``Check if the input is valid''). Unlike pseudocode or flowchart formats, this version entirely removes code or symbolic notation and instead emphasizes procedural understanding in purely narrative form.

\textbf{Code in imaginary language}
We translate real code into a fictional language that preserves structure and control flow but replaces all syntax and identifiers with invented tokens (Figure~\ref{fig:imaginary_language_prompt}). The result is semantically consistent yet entirely ungrounded in real languages. This perturbation allows us to examine whether models rely on surface-form familiarity (e.g., recognizing logical patterns.

\subsection{Model Training and Evaluation} \label{sec:model_training}

We train a suite of decoder-only LLMs using supervised fine-tuning (SFT) on our instruction–response datasets detailed in Section~\ref{data_generation}, along with their perturbed variants described in Section~\ref{sec:perturbation_design}. To assess the effect of language-specific patterns, we additionally finetune models on subsets of the code data restricted to a single programming language. This allows us to examine how the syntactical diversity of programming languages influences reasoning performance. 
% , comprising 120K code instructions spanning ten programming languages, along with their systematically perturbed variants, and 120K natural language instructions. 
Each instruction–response pair is treated as a single input–output sequence, and models are trained to autoregressively predict the response tokens conditioned on the instruction and prior context. All models are fine-tuned from the same pre-trained backbone under supervised fine-tuning (SFT) objective to ensure comparability across experimental conditions. Let $x = (x_1, x_2, \dots, x_m)$ be the instruction tokens and $y = (y_1, y_2, \dots, y_n)$ be the response tokens. The SFT objective is defined as: 
\begin{equation}
\mathcal{L}_{\text{SFT}} = - \sum_{t=1}^{n} \log P_\theta(y_t \mid x, y_{<t})
\end{equation}
where $P_\theta$ denotes the model’s conditional probability distribution parameterized by $\theta$, and $y_{<t}$ represents the prefix of the response up to position $t-1$.

\paragraph{Models}
We choose a diverse set of pre- and post-trained language models ranging from 0.6B to 8B parameters. Specifically, we experiment with models from five major families: Qwen3~\citep{yang2025qwen3}, LLaMA-3~\citep{grattafiori2024llama}, Gemma3~\citep{team2025gemma}, OLMo2~\citep{olmo20242}, and SmolLM2~\citep{allal2025smollm2}. For each model family, we select representative sizes (e.g., $<$1B, $\sim$1B, $\sim$3-4B, $\sim$7-8B)\footnote{Due to resource constraint, the larges model we could finetune is 8B.} to evaluate performance across different scales. 
% While we take a pre-trained version as a base model, however, in the case of Qwen3, we also take a post-trained model as a base to understand the effect of code data. 

\paragraph{Training data configurations}
Our base training set consists of 120K instruction–response pairs spanning both code and natural language formats detailed in Section~\ref{data_generation}. From this, we construct several configurations: (1) 100\% code-only, (2) 100\% natural language-only, and (3) mixed data with varying code-to-language ratios. In addition, we train models on each perturbed variant introduced in Section~\ref{sec:perturbation_design}. Finally, we include programming-language-specific subsets, training separate models on data from each of the ten languages ($\sim$12K examples per language) to assess the effect of language specialization. The implementation details are in Section~\ref{sec:implementation_details}. 
% \vspace{-7mm}
\paragraph{Evaluation tasks}
We evaluate model performance across three categories: natural language and general knowledge, math, and code (Table~\ref{tab:eval_tasks}). 

For natural language and general knowledge, we evaluate across commonsense reasoning, science and textbook-style QA, logical reasoning, and instruction-following. All tasks are evaluated using accuracy. For math, we include both elementary and advanced problem-solving datasets (e.g., GSM8K, HRM8K), as well as arithmetic and math-related subsets of MMLU. Open-ended tasks (GSM8K, HRM8K) use exact match, while arithmetic and MMLU (math) are scored with accuracy.

For code, we evaluate both code understanding and generation. Based on preliminary experiments, we adopt the LLM-as-Judge paradigm~\citep{gu2025surveyllmasajudge} instead of execution-based evaluation~\citep{huang2022execution}. Our relatively small, perturbed models often fail to produce fully executable code, making execution-based metrics unreliable. More importantly, our goal is to assess code quality and reasoning under perturbations, not just execution success.

Thus, we prompt \emph{GPT-4o-mini} to first generate an instance-specific rubric on a 1--10 Likert scale given the original instruction, which is expected to capture nuanced quality variation across outputs. The same model is then prompted as a judge to provide a brief reasoning step (``thought’’) and assign a score based on that rubric. Examples of the rubric-generation prompt and judging prompt are shown in Appendix~\ref{appsubsec:prompts} (Figures~\ref{fig:rubric_gen} and \ref{fig:llm_as_judge}).

% \paragraph{Implementation details}
% We train all models under identical hyperparameter settings to ensure a fair comparison across model sizes and data configurations. All experiments are conducted using full finetuning in \textit{BF16} precision with a maximum sequence length of 2048 tokens. We run all experiments on 4$\times$A100 80G node. Models are trained for 2 epochs with a cumulative batch size of 64 for most experiments, except for language-specific settings, where the batch size is reduced to 32. The learning rate is fixed at $1\mathrm{e}{-5}$ and follows a cosine decay schedule with a warmup ratio of 0.1. For memory-efficient parallelism and distributed training, we use \textit{DeepSpeed ZeRO Stage 3}~\citep{ren2021zero}. All models are trained using the LLaMA-Factory framework~\citep{zheng2024llamafactory}. All other parameters and configurations follow the default setting unless otherwise specified.

% 5. Results
% \vspace{-6mm}
\section{Results and Discussion}
% \vspace{-3mm}
\paragraph{RQ1: Does incorporating code in finetuning improve task performance?}\label{sec:results_RQ1}

% \noindent
% \begin{tcolorbox}[
%   colback=yellow!10!white,
%   colframe=black,          % frame color
%   boxrule=0.5pt,           % thin border
%   title=Section Findings,
%   colbacktitle=yellow!30,  % title background
%   coltitle=black,          % title text color
%   fonttitle=\bfseries      % bold title
% ]

% \end{tcolorbox}

\begin{figure}[h]
  \centering
  \includegraphics[width=0.9\columnwidth]{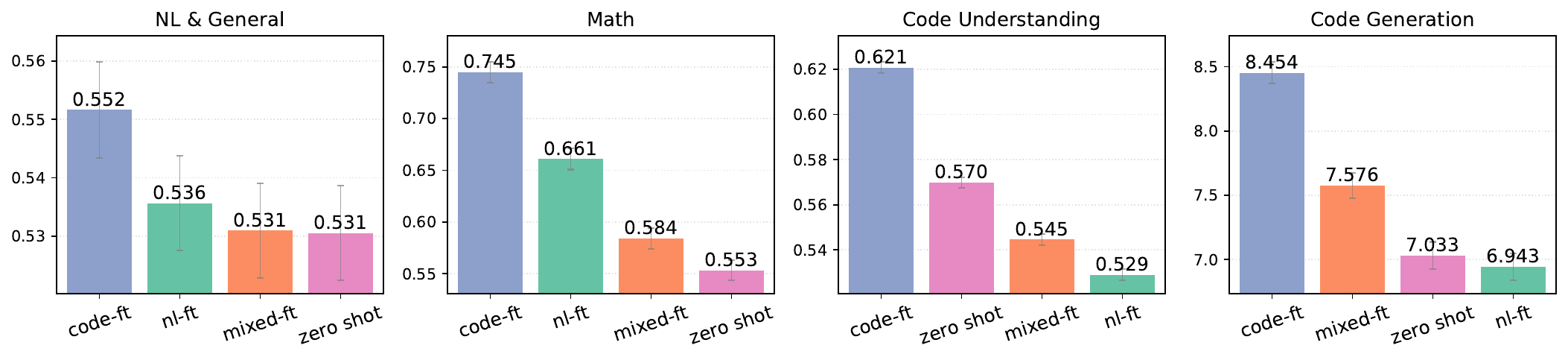}
  \caption{Performance (with stderr bars) of Qwen3-4B-Base across zero-shot, full code finetuning (code-ft), full natural language finetuning (nl-ft), and 50-50 code to NL data ratio finetuning (mixed ft). Incorporating code improves performance across tasks.}
  \label{fig:qwen4bb-rq1}
\end{figure}

First, we validate prior findings that finetuning on code data can enhance downstream reasoning. Following the training setup in Section~\ref{sec:model_training}, we compare performance across four settings: zero-shot, full code finetuning (``code-ft''), full natural language finetuning (``nl-ft''), and mixed data finetuning with equal proportions of code and natural language instructions (``mixed-ft''). Across model families and scales, code-ft and mixed-ft generally achieve leading or competitive performance across tasks (Figure~\ref{fig:qwen4bb-rq1}, and Figures~\ref{fig:rq1_qwen_combined}--\ref{fig:rq1_sml_combined}), with the trend particularly consistent on code generation.  
% In some models like Llama-3.2B-\{1B, 3B\} and Qwen3-8B-Base, the mixed configuration yields top performance on math tasks. 
% \vspace{-3mm}

Overall, across the 14 model bases, either code-ft or mixed-ft achieves the best performance on 64\% of natural language tasks, 86\% of math and code understanding tasks, and all code generation tasks.
% This further motivates us to investigate whether the ratio of code to natural language data in mixed finetuning has an impact on task performances.
Motivated by this, we further examine the effect of varying the proportion of code in mixed finetuning (Figure~\ref{fig:rq4_qwen_combined}). We find that higher fractions of code data generally improve performance across most tasks, with math tasks most sensitive to mixture ratios.
% \vspace{-6mm}

% To this end, we further finetuned Qwen3-\{0.6B, 1.7B\}-Base with code data mixture ratio from \{2.5\%, 5\%, 10\%, 20\%, 40\%, 60\%, 80\%, 90\%, 95\%, 97.5\%\}. We observe that on NL tasks and particular math tasks, mixing ratios of code data is generally worse than code-ft, while the trend is not clear for code tasks. 

\paragraph{RQ2: How do our systematic perturbations affect performance?}\label{sec:results_RQ2}

\noindent
\begin{tcolorbox}[
  colback=yellow!10!white,
  colframe=black,          % frame color
  boxrule=0.5pt,           % thin border
  title=Section Findings,
  colbacktitle=yellow!30,  % title background
  coltitle=black,          % title text color
  fonttitle=\bfseries,      % bold title
  left=10pt, right=2pt, top=2pt, bottom=2pt % tighter margins
]
\begin{itemize}[leftmargin=*, labelsep=0.3em]
    \item Structural perturbations hurt more than semantic ones, especially for math and code.
    \item Appropriate abstractions such as pseudocode and flowcharts can substitute for explicit code structure in reasoning.
    \item Models don’t need verbose code: reduced-token variants perform well as long as core information is preserved.
    \item LLMs can reason effectively from corrupted code by exploiting surface-level regularities.
\end{itemize}
\end{tcolorbox}

% Main figure (takes full width)
% \begin{figure}[t]
%   \centering
%   % trim = left bottom right top (in points). Adjust numbers until it looks right.
%     \includegraphics[width=\columnwidth]{assets/plots_no_hellaswag/RQ2/show_all/Qwen3-4B-Base_rq2.pdf}
%   \caption{Performance of Qwen3-4B-Base under all perturbations across tasks. All perturbations hurt code understanding and math ability.
%   The perturbations seems to make less of a difference for general knowledge and natural language understanding tasks.
%   The only perturbation setting where we see an improvement to benchmark performance is when he enhance the comments in the code, which leads to better code implementation ability. \dei{You should mention what the error bars are. To save vertical space, you should remove the 'code-ft' legend at top, and just explain this in the caption, or else move it to a corner.}}
%   \label{fig:qwen4bb-rq2a}
% \end{figure}

% Secondary figure (floats in text)
% before the paragraph you want it to start wrapping
% Put this near your paragraph start (no blank line before it)
% \begin{wrapfigure}[10]{r}{0.8\columnwidth} % [lines]{r|l}{box width}
\begin{figure}
    \centering
    \includegraphics[width=0.72\linewidth]{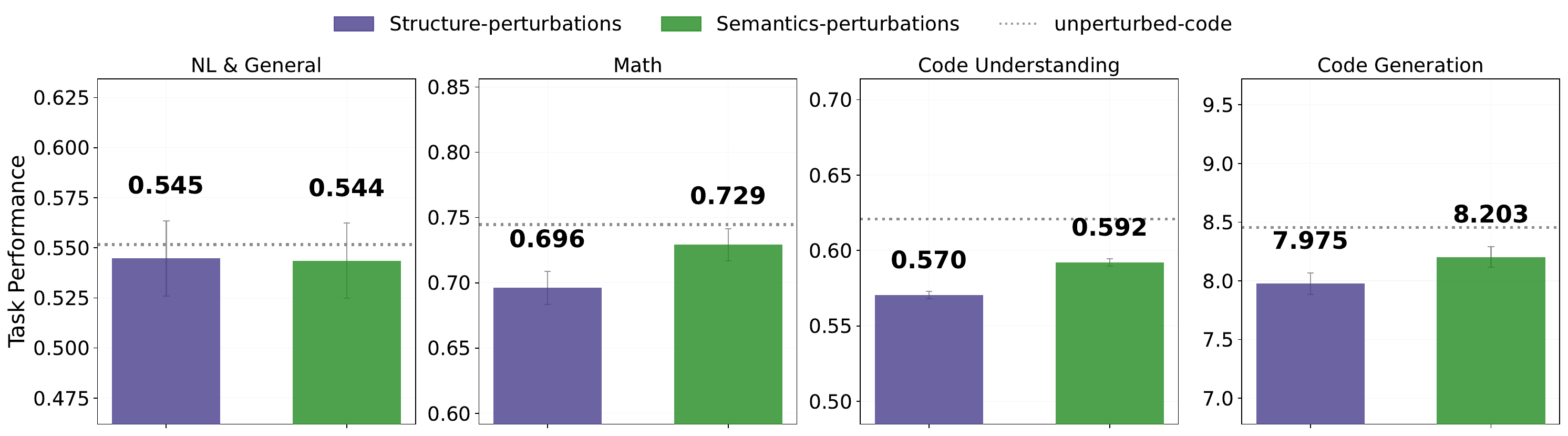}
    \caption{Aggregated performance (with stderr bars) under structural perturbations (e.g. removing whitespace) vs. semantics perturbations (e.g. modifying the comments) of Qwen3-4B-Base. Semantic perturbations tend to be more harmful to performance than semantic ones.}
    \label{fig:qwen4bb-rq2b}
\end{figure}
%   \vspace{-0.6\baselineskip}               % (optional) nudge up
%   \centering
%   \includegraphics[width=\linewidth]{assets/plots_no_hellaswag/RQ2/ss/Qwen3-4B-Base_ss.pdf}
%   \captionsetup{aboveskip=3pt,belowskip=0pt,font=small}
%   \caption{Aggregated performance under structural perturbations (e.g. removing whitespace) vs. semantics perturbations (e.g. modifying the comments) of Qwen3-4B-Base. Semantic perturbations tend to be more harmful to performance than semantic ones.}
%   \label{fig:qwen4bb-rq2b}
%   \vspace{-0.6\baselineskip}               % (optional) trim space below
% \end{wrapfigure}

\begin{figure}
    \centering
    \includegraphics[width=0.75\linewidth]{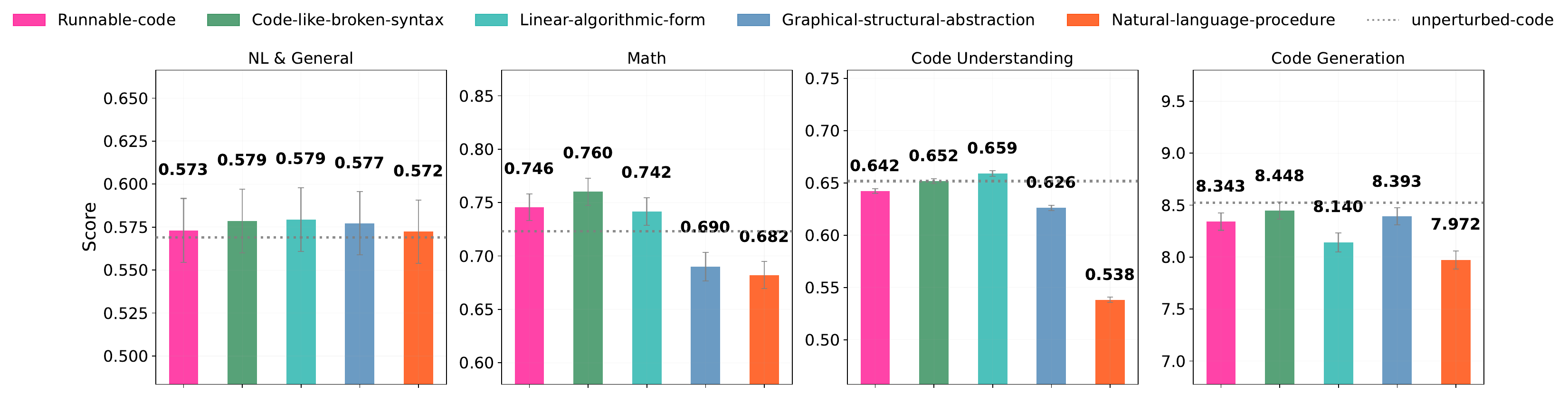}
    \caption{Aggregated performance (with stderr bars) under levels of explicitness of code structure (less explicit going from runnable code to NL procedure) of Qwen3-8B-Base. Certain algorithmic and graphical abstractions benefit reasoning.}
    \label{fig:qwen8b-rq2-ecs}
\end{figure}

\begin{figure}
    \centering
    \includegraphics[width=0.75\linewidth]{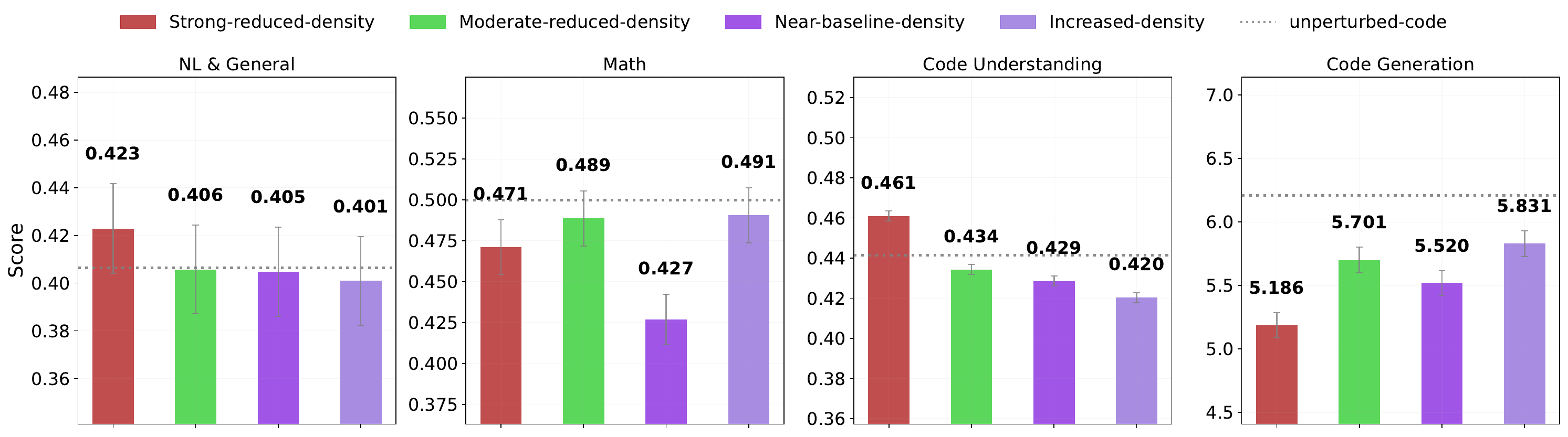}
    \caption{Aggregated performance (with stderr bars) of Qwen3-0.6B-Base with various of token counts wrt to unperturbed code. Reductions can perform comparable or even better than the baseline.}
    \label{fig:qwen0.6bb-rq2-rid}
\end{figure}

\begin{figure}
    \centering
    \includegraphics[width=0.75\linewidth]{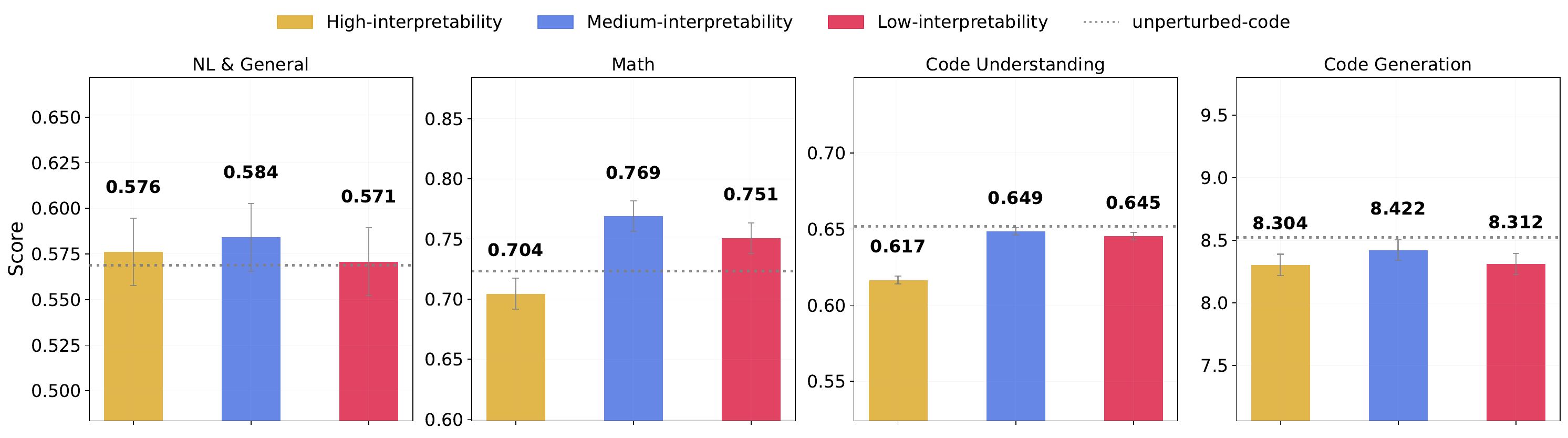}
    \caption{Aggregated performance of Qwen3-8B-Base (with stderr bars), depending on how much the perturbed code data is readable to humans. Low-interpretability with misleading signals can match or perform better than other configurations.}
    \label{fig:qwen8b-rq2-hi}
\end{figure}

Next, we analyze task performance under the perturbations introduced in Section~\ref{sec:perturbation_design}. Based on the properties of each perturbation, we group them into distinct analysis axes that allow us to systematically probe their effects. The grouping details are in Table~\ref{tab:perturbation_axes}. We illustrate performance of individual perturbations in Appendix~\ref{appx:rq2_all}.

\textbf{Structural vs. Semantics Perturbations.} We define structural perturbations as edits that alter the syntactic scaffolding or formatting of code (e.g., whitespace removal, pseudocode, flowcharts), while semantic perturbations modify meaning-bearing tokens such as identifiers, keywords, or comments without disrupting the underlying structure. Across model families and scales (Figures~\ref{fig:rq2_qwen_ss}
% , \ref{fig:rq2_llama_ss}, \ref{fig:rq2_gemma_ss}, \ref{fig:rq2_olmo_ss}, 
--\ref{fig:rq2_sml_ss}), nearly all perturbations reduce performance compared to the unperturbed code-fineturned baseline. More importantly, structural perturbations consistently degrade performance more severely than semantic ones, especially for math and code tasks (e.g., Figure~\ref{fig:qwen4bb-rq2b}). The discrepancy is more evident as models scale up (e.g., Figure~\ref{fig:rq2_qwen_ss}). This resembles prior work that reasoning structure rather than content is more critical to the learning process~\citep{li2025llms}. We hypothesize that tasks such as math and code rely more heavily on formatting and layout cues to shape reasoning. 

\textbf{Explicitness of Code Structure.} Building on the importance of structure, we examine perturbations along a spectrum of how explicitly they preserve code structure: from runnable or code-like forms, through intermediate abstractions such as pseudocode and flowcharts, to natural language step-by-step procedures. For code generation, where executable outputs are required, it is natural that perturbations that preserve explicit code structure, whether runnable or not, lead to the best performance. For other tasks, however, certain abstractions such as pseudocode or flowcharts often match or even surpass unperturbed code, as they highlight algorithmic structure while removing superficial syntax. By contrast, the most implicit form, natural language procedures, provides little advantage and generally performs worst across tasks (e.g. Figure~\ref{fig:qwen8b-rq2-ecs}, Figures~\ref{fig:rq2_qwen_ecs}--\ref{fig:rq2_sml_ecs}).

\textbf{Relative Information Density.} Because our constructed instruction datasets are parallel, the amount of information they convey about the code is comparable across perturbations. We define relative information density as (number of tokens in perturbed dataset) $\div$ (number of tokens in the original code-ft dataset), which reflects how compactly the same content is represented. Perturbations differ in how they adjust density: some produce highly compact forms that strip away most tokens but preserve the algorithmic skeleton (e.g., flowcharts, pseudocode), others moderately reduce density by removing comments or using imaginary languages, while others preserve or even increase density through verbose variable renamings or enriched documentation.
We find that strong or moderate reductions in density often perform close to, and sometimes better than, the baseline (e.g. Figure~\ref{fig:qwen0.6bb-rq2-rid}, Figures~\ref{fig:rq2_qwen_rid}--\ref{fig:rq2_sml_rid}). However, this advantage doesn't extend to code generation, where preserving richer surface detail is important. In addition, smaller models are more sensitive to density differences, whereas larger models remain robust. Overall, this suggests that the benefit of code for reasoning doesn't lie in its verbosity but but in the efficiency with which essential information is preserved.

\textbf{Human Interpretability.} We also examine perturbations through the lens of human readability: high-interpretability (enriched explanations and visual scaffolds), medium (local edits leaving most code intact), and low (obscured readability or misleading signals). Interestingly, low-iterpretability variants, despite adding noise or distortion, often do not degrade performance too much from the unperturbed baseline, and often match or even surpass medium-interpretability ones (e.g. Figure~\ref{fig:qwen8b-rq2-hi}, Figures~\ref{fig:rq2_qwen_hi}--\ref{fig:rq2_sml_hi}). This counterintuitive trend suggests that the models could exploit surface-level regularities and recurring structural cues that persist even in noisy or opaque forms.

\paragraph{RQ3: How does performance vary across programming languages?}\label{sec:results_RQ3}

\noindent
\begin{tcolorbox}[
  colback=yellow!10!white,
  colframe=black,          % frame color
  boxrule=0.5pt,           % thin border
  title=Section Findings,
  colbacktitle=yellow!30,  % title background
  coltitle=black,          % title text color
  fonttitle=\bfseries,      % bold title
  boxsep=2pt,      % inner padding
  left=10pt, right=2pt, top=2pt, bottom=2pt % tighter margins
]
\begin{itemize}[leftmargin=*, labelsep=0.5em]
    \item Lower-level languages benefit math tasks.
    \item Python aligns best with NL tasks, while Java and Rust often rank among the top for math.
\end{itemize}
\end{tcolorbox}

\begin{figure}[t]
  \centering

  \begin{subfigure}{\columnwidth}
    \centering
    \includegraphics[width=0.75\columnwidth]{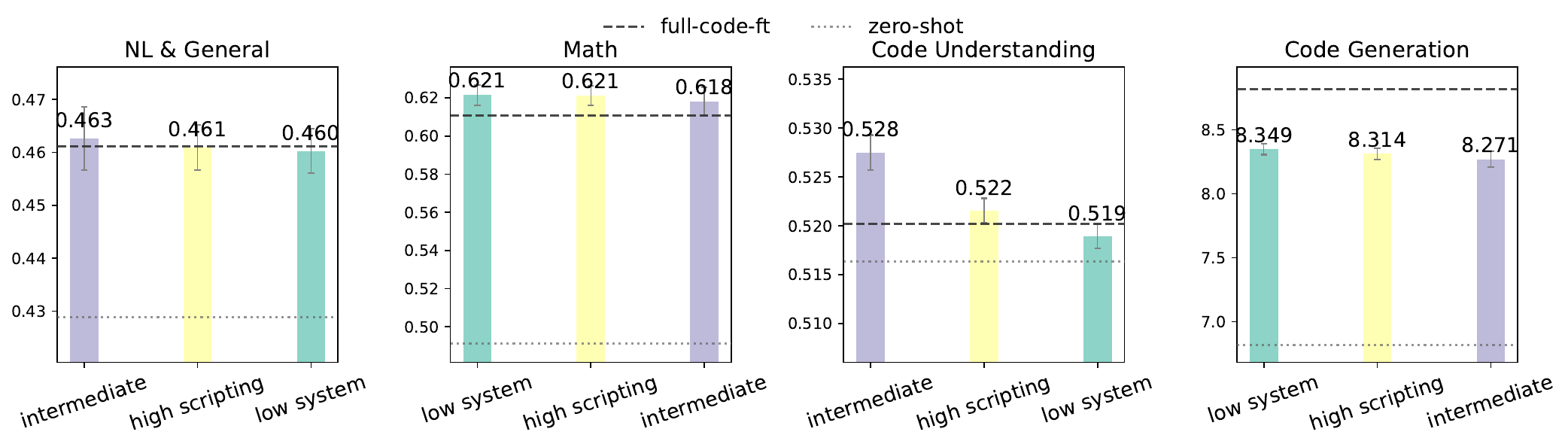}
    % \label{fig:qwen1.7b_rq3_groups}
  \end{subfigure}

  \begin{subfigure}{\columnwidth}
    \centering
    \includegraphics[width=0.75\columnwidth, clip, trim={0 0 0 30}]{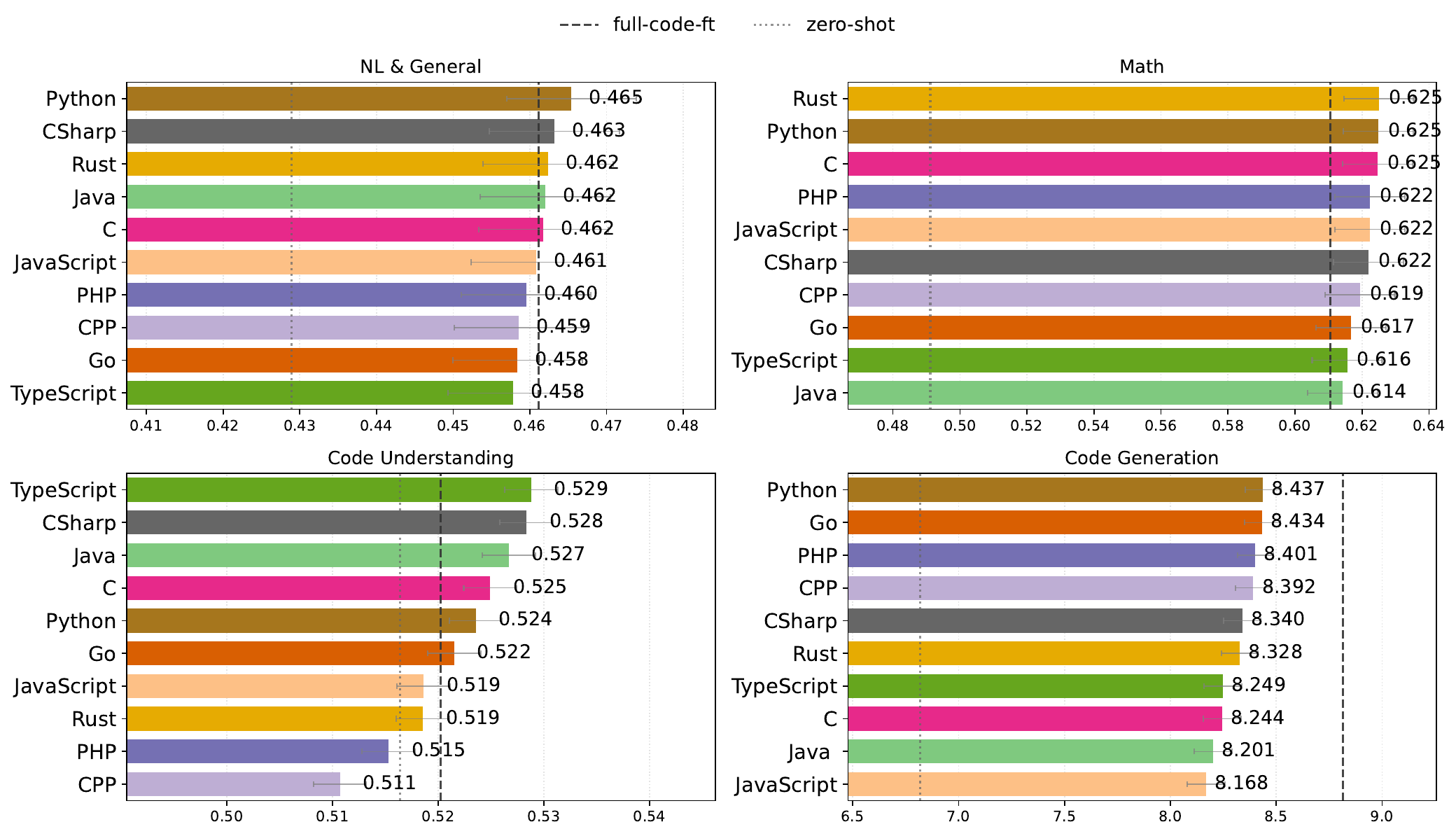}
    % \label{fig:qwen1.7b_rq3_langs}
  \end{subfigure}

  \caption{Performance (with stderr bars) of Qwen3-1.7B. \textit{Top:} grouped by abstraction level (low-system, intermediate, high-scripting). Low-system and intermediate languages outperform on math. 
  \textit{Bottom:} individual programming languages. Python aligns best with NL, Rust leads on math.}
  \label{fig:rq3_qwen3_1.7b}
\end{figure}

The strong impact of structure in RQ2 motivates the question of whether syntactic regularities in programming languages also influence model performance. To explore this, we group the ten programming languages into high-scripting (Python, PHP, JavaScript, TypeScript), intermediate (Java, C\#), and low-system (C, C++, Rust, Go) according to their abstraction level. Overall, differences across groups are small. On NL and code tasks, the impact of language groups is largely model-dependent. However, on math tasks, high-scripting languages consistently underperform relative to intermediate and low-system ones (e.g. top Figure~\ref{fig:rq3_qwen3_1.7b}, Figures~\ref{fig:qwen_rq3_groups_combined}--\ref{fig:sml_rq3_groups}). We hypothesize that richer structural detail in lower-level languages provides beneficial signals for mathematical reasoning. For code generation, finetuning on any single language improves over zeroshot but lags behind full code finetuning, which suggests the benefit of multi-language diversity for code generation.
% Moreover, in model families like OLMo2 (Figure~\ref{fig:olmo_rq3_groups}) and Qwen3 (Figure~\ref{fig:qwen_rq3_groups_combined}), these language-specific configurations can even surpass the full code configuration (``code-ft''), which suggests that certain languages may be more suitable for particular tasks. This motivates us to take a closer look at the role of individual programming languages.

At the individual language level (e.g. bottom Figure~\ref{fig:rq3_qwen3_1.7b}, Figures~\ref{fig:qwen_rq3_langs_combined}--\ref{fig:sml_rq3_langs}), across models, Python often leads on NL tasks, probably due to its surface form being closer to natural language. Aligning with the group-level results, lower-level languages such as Java and Rust often rank among the top for math. For code tasks that span multiple languages, results are more mixed, with no clear leaders, and performance gaps remain relatively small.

% 6. Conclusion
\section{Conclusion}\label{sec:conclusion}
In this work, we aim to understand what aspects of code enhance reasoning in LLMs and which aspects matter most. Through 3,331 finetuning experiments spanning five model families, eight scales, ten programming languages, and a suite of systematic perturbations, we arrive at four central conclusions. First, structural properties of code are critical: disrupting them leads to consistent performance drops, especially on math and code tasks. Second, appropriate abstractions and efficient encodings can be just as effective as raw code. Moreover, models remain surprisingly robust even to corrupted or low-interpretability code, exploiting statistical regularities that persist despite surface distortions. Finally, lower-level programming languages provide more benefits for math tasks. Together, we want to provide a more precise account of how code supports reasoning and point toward practical design principles for constructing effective training data beyond executable programs.

% 7. Limitations
\section{Limitations}\label{sec:limitations}
Our study focuses on small- to mid-scale base models due to resource constraints. Future work could extend our framework to larger models. Our perturbations, although diverse, may still not cover enough and leave out other factors like code complexity and data diversity. Finally, although we evaluate across a broad suite of reasoning tasks, our benchmarks still capture only part of the reasoning spectrum, and future work could extend the analysis to additional domains.

\section{Reproducibility Statement}
We provide extensive details throughout the paper and supplementary materials. Section~\ref{data_generation} describes the construction and processing of both the code and natural language datasets. Section~\ref{sec:implementation_details} outlines model training and implementation details. Appendix~\ref{appsubsec:prompts} includes all prompts used for data generation, perturbations, and LLM-as-Judge evaluation.

\bibliography{iclr2026_conference, custom}
\bibliographystyle{iclr2026_conference}

\appendix
\section{Appendix}
% Appendix 
\subsection{Evaluation suite details}
See Table~\ref{tab:eval_tasks}.
% requires: \usepackage{multirow,booktabs,adjustbox}
\begin{table}[htbp]
\centering
% \small
\caption{Evaluation suite spanning natural language and general knowledge, math, and code tasks.}
\label{tab:eval_tasks}
\begin{adjustbox}{width=\linewidth}
\begin{tabular}{p{1in} l p{0.5\linewidth} p{1in}}
\toprule
\textbf{Task Type} & \textbf{Topic} & \textbf{Benchmarks} & \textbf{Metric} \\
\midrule

% ---------- NL & General ----------
\multirow{7}{=}{Natural Language \& General Knowledge}
  & Commonsense 
  & PIQA~\citep{bisk2019piqareasoningphysicalcommonsense} 
  & \multirow{6}{*}{Accuracy} \\
  \cmidrule(l){2-3}
  & \multirow{4}{*}{Science / Textbook} 
  & ARC-Easy~\citep{clark2018thinksolvedquestionanswering} & \\
  &  
  & ARC-Challenge~\citep{clark2018thinksolvedquestionanswering} & \\
  &  
  & OpenBookQA~\citep{mihaylov2018suitarmorconductelectricity} & \\
  &  
  & MMLU (non-math)~\citep{hendrycks2021measuringmassivemultitasklanguage} & \\
  \cmidrule(l){2-3}
  & Logic-Heavy 
  & LogiQA~\citep{liu2020logiqachallengedatasetmachine} & \\
  \cmidrule(lr){2-3}\cmidrule(lr){4-4}
  & Instruction Following
  & IFEval~\citep{zhou2023instructionfollowingevaluationlargelanguage} 
  & Prompt-level Accuracy \\
\midrule

% ---------- Math ----------
\multirow{4}{=}{Math}
  & \multirow{2}{*}{--} 
  & GSM8K~\citep{cobbe2021trainingverifierssolvemath} 
  & \multirow{2}{*}{Exact Match} \\
  &
  & HRM8K~\citep{ko2025understandsolvetranslatebridging} & \\
  \cmidrule(lr){2-3}\cmidrule(lr){4-4}
  & \multirow{2}{*}{--}
  & Arithmetic~\citep{brown2020language} 
  & \multirow{2}{*}{Accuracy} \\
  &
  & MMLU (math)~\citep{hendrycks2021measuringmassivemultitasklanguage} & \\
\midrule

% ---------- Code ----------
\multirow{2}{=}{Code}
  & Code Understanding
  & CodeMMLU~\citep{dung2024codemmlu} 
  & Accuracy \\
  \cmidrule(l){2-3}
  & Code Generation
  & HumanEvalX~\citep{zheng2023codegeex} 
  & LLM-as-Judge \\
\bottomrule
\end{tabular}
\end{adjustbox}
% \caption{Evaluation suite spanning natural language and general knowledge, math, and code tasks.}
% \label{tab:eval_tasks}
\end{table}

\subsection{Categorization of perturbations for RQ2 analysis}
See Table~\ref{tab:perturbation_axes}.
% requires: \usepackage{multirow,booktabs}
\begin{table*}[t]
\centering
\small
\renewcommand{\arraystretch}{1.15}
\caption{Categorization of perturbations across four analysis axes: structural vs.\ semantic (S/S) perturbations, explicitness of code structure (ECS), relative information density (RID), and human interpretability (HI).}
\label{tab:perturbation_axes}
\begin{tabular}{l l l l l}
\toprule
\textbf{Perturbation} & \textbf{S/S Perturbations} & \textbf{ECS} & \textbf{RID} & \textbf{HI} \\
\midrule

Whitespace removal      & \multirow{5}{*}{Structural} & Broken syntax  & Moderate-reduced & Medium \\
Pseudocode              &                             & Algorithmic    & Strong-reduced   & High \\
Imaginary               &                             & Broken syntax  & Moderate-reduced & Low \\
Step-by-step            &                             & NL procedure   & Moderate-reduced & High \\
Flowchart               &                             & Graphical      & Strong-reduced   & High \\
\midrule
Comment removal          & \multirow{8}{*}{Semantic}   & Runnable       & Moderate-reduced & Medium \\
Variable renaming        &                             & Runnable       & Increased        & Medium \\
Keyword repl. (nonsense) &                             & Broken syntax  & Increased        & Low \\
Keyword repl. (non-Eng.) &                             & Broken syntax  & Increased        & Low \\
Comment swap (global)    &                             & Runnable       & Near-baseline    & Low \\
Comment swap (local)     &                             & Runnable       & Near-baseline    & Low \\
Comment enhancement      &                             & Runnable       & Increased        & High \\
Comment obfuscation      &                             & Runnable       & Increased        & Low \\
\bottomrule
\end{tabular}
% \caption{Categorization of perturbations across four analysis axes: structural vs.\ semantic (S/S) perturbations, explicitness of code structure (ECS), relative information density (RID), and human interpretability (HI).}
% \label{tab:perturbation_axes}
\end{table*}

\subsection{Implementation details} \label{sec:implementation_details}
We train all models under identical hyperparameter settings to ensure a fair comparison across model sizes and data configurations. All experiments are conducted using full finetuning in \textit{BF16} precision with a maximum sequence length of 2048 tokens. We run all experiments on 4$\times$A100 80G node. Models are trained for 2 epochs with a cumulative batch size of 64 for most experiments, except for language-specific settings, where the batch size is reduced to 32. The learning rate is fixed at $1\mathrm{e}{-5}$ and follows a cosine decay schedule with a warmup ratio of 0.1. For memory-efficient parallelism and distributed training, we use \textit{DeepSpeed ZeRO Stage 3}~\citep{ren2021zero}. All models are trained using the LLaMA-Factory framework~\citep{zheng2024llamafactory}. All other parameters and configurations follow the default setting unless otherwise specified.

% \newpage
\subsection{Prompts}\label{appsubsec:prompts}
\begin{table}[t]
\centering
{\fontsize{8}{11}\selectfont  % Slightly larger than \scriptsize
\caption{Language specification templates with placeholders that can be instantiated with different programming languages.}
\label{tab:code_instruction_suffix}
\begin{tabularx}{\linewidth}{@{}p{0.31\linewidth}@{\hskip 0.2cm}p{0.31\linewidth}@{\hskip 0.2cm}p{0.31\linewidth}@{}}
\toprule
% \multicolumn{3}{c}{\textbf{Instruction suffixes (with \{language\} placeholder)}} \\
% \midrule
Generate the code in \{language\}. & Provide code in \{language\}. & Write the code in \{language\}. \\
Build the code using \{language\}. & Create the code using \{language\}. & Draft the code in \{language\}. \\
Produce a code snippet in \{language\}. & Develop the code using \{language\}. & Generate a solution in \{language\}. \\
Create a script in \{language\}. & Implement the code in \{language\}. & Design the code in \{language\}. \\
Construct the code using \{language\}. & Format the code in \{language\}. & Write a program in \{language\}. \\
Prepare a code snippet in \{language\}. & Write a function in \{language\}. & Deliver the code in \{language\}. \\
\bottomrule
\end{tabularx}
}
% \caption{Language specification templates with placeholders that can be instantiated with different programming languages.}
% \label{tab:code_instruction_suffix}
\end{table}

\paragraph{Standard generation prompt}  
We provide the standard prompt to generate code for a given instruction in a specific language in Figure~\ref{fig:code_gen_prompt}.  
\begin{figure*}[h]
    \centering
    \tcbset{
  colframe=black!80,   % lighter border
  colback=gray!9,      % lighter background
  arc=4mm,             % slightly softer corners
  boxrule=0.8pt,       % thinner border line
  fonttitle=\bfseries, % bold if you use titles
  fontupper=\large,     % larger font size for content
  title={Code Instruction Data Generation Prompt}
}
    \begin{tcolorbox}
    \tiny
    \textbf{You are tasked with generating code based on a specified programming language and instruction. Your goal is to generate code that follows the syntax and semantics of the specified language. If the instruction is invalid (e.g., contradicts the language's rules or references functions or constructs from a different language), you must strictly respond with \textbf{"invalid."}}

    \textbf{Guidelines:}  
    - \textbf{Valid Code:}  
       - The generated code must be syntactically and semantically correct according to the specified language.  
       - The code should follow standard conventions and best practices for the given language.  
       - Do \textbf{not} provide any explanation for valid code — only output the code itself.  

    - \textbf{Invalid Instruction:}  
       - If the instruction references constructs, functions, or syntax not supported by the specified language, respond with `"invalid"`.  
       - Do \textbf{not} attempt to correct the invalid instruction — just respond with `"invalid"`.  
       - Do \textbf{not} provide a reason or explanation for why the instruction is invalid.

    \textbf{Examples:}  

    Example 1: \\
    Instruction: "Write a function to convert a list to a set."  \\
    Language: Python  \\
    Response: 
    \begin{verbatim}
    def list_to_set(input_list):
        return set(input_list)
    \end{verbatim}

    Example 2:\\
    Instruction: "Create a class with a method that prints 'Hello' using console.log()." \\
    Language: Python \\
    Response: invalid

    Example 3: \\
    Instruction: "List all files, including hidden ones, in the current directory." \\
    Language: Shell \\
    Response: ls -a

    Example 4: \\
    Instruction: "Define a function using 'def' that returns the length of a string." \\
    Language: JavaScript \\
    Response: invalid \\

    Instruction: \\
    If the instruction is valid, output the code directly (no explanations). \\ 
    If the instruction is invalid, respond with "invalid" (no explanation). \\

    \textbf{Input:}  
    Instruction: \{instruction\}  \\
    Language: \{language\} \\

    \textbf{Output:}  \\
    \{\{response\}\}
    \end{tcolorbox}
    \caption{Code instruction data generation prompt. The task is to generate valid code or respond with ``invalid'' for unsupported instructions.}
    \label{fig:code_gen_prompt}
\end{figure*}, where the \textit{instruction} can be instantiated using one of the templates in Table~\ref{tab:code_instruction_suffix}.

\paragraph{Comment enhancement prompt}  
The prompt to enhance the quality and readability of a given code snippet by adding detailed documentation is shown in Figure~\ref{fig:good_code_prompt}.  
\begin{figure*}[h!]
    \centering
        \tcbset{
  colframe=black!80,   % lighter border
  colback=gray!9,      % lighter background
  arc=4mm,             % slightly softer corners
  boxrule=0.8pt,       % thinner border line
  fonttitle=\bfseries, % bold if you use titles
  fontupper=\large,     % larger font size for content
  title={Comment Enhancement Prompt}
}
    \begin{tcolorbox}
    \tiny
    \textbf{You are tasked with enhancing the response to the given code instruction by adding meaningful comments and documentation.} The goal is to improve the code's readability, maintainability, and clarity across any programming language, without altering its original logic or structure.

    \textbf{Your modifications must include:}

    1. \textbf{Documentation Comments}:
       - Add clear, technically accurate, and concise documentation for every function, method, class, and major code block.
       - Describe the purpose, all parameters (with correct types and usage), return values, and any assumptions or notes relevant to correct usage.
       - Use the standard documentation format appropriate for the programming language (e.g., Python docstrings, JavaDoc for Java, Doxygen for C/C++).

    2. \textbf{Inline Comments}:
       - Insert informative and contextually helpful inline comments near complex, unintuitive, or important operations.
       - Focus on explaining logic, control flow, edge-case handling, design decisions, or dependencies.
       - Avoid redundant, obvious, or overly literal comments (e.g., avoid "i = 0  // set i to 0").

    \textbf{Guidelines:}

    - \textbf{Do not change the logic, structure, or behavior} of the original code.
    - Do not introduce new functionality, abstractions, or formatting changes.
    - Keep comments strictly technical, relevant, and useful—avoid verbosity or informal tone.
    - Do not include any meta-comments or explanatory notes about what was changed (e.g., no "This version adds comments" or similar).
    - Apply these modifications \textbf{only} to the code portion of the original response. Leave any non-code parts completely unmodified.
    - Ensure all comments follow the style conventions of the language in use.

    This task is language-agnostic and applies to any of the following: Python, Java, C++, JavaScript, PHP, TypeScript, Go, Rust, C\#, Ruby, Swift, Kotlin, Scala, and Shell.

    \textbf{Input:}  
    Instruction: \{instruction\}  
    Original Response: \{response\}

    \textbf{Output:}  
    \{\{updated response\}\}
    \end{tcolorbox}
    \caption{Comment enhancement prompt. The task is to improve code clarity through meaningful comments while preserving original functionality.}
    \label{fig:good_code_prompt}
\end{figure*}

\paragraph{Comment obfuscation prompt}  
The prompt used to generate obfuscated versions of code from a given instruction is presented in Figure~\ref{fig:bad_code_prompt}.  
\begin{figure*}[h]
    \centering
        \tcbset{
  colframe=black!80,   % lighter border
  colback=gray!9,      % lighter background
  arc=4mm,             % slightly softer corners
  boxrule=0.8pt,       % thinner border line
  fonttitle=\bfseries, % bold if you use titles
  fontupper=\large,     % larger font size for content
  title={Comment Obfuscation Prompt}
}
    \begin{tcolorbox}
    \tiny
    \textbf{You are tasked with modifying the response to the given code instruction in a way that \textbf{significantly degrades its quality and clarity}.} Your goal is to make the code as \textbf{confusing, unhelpful, and misleading} as possible through the use of deliberately bad, distracting, and nonsensical comments—without altering the actual logic or functionality of the code.

    \textbf{Your modifications must include:}

    1. \textbf{Severely Misleading or Irrelevant Documentation Comments}:
       - Add documentation to every function, class, and major code block that includes wildly inaccurate descriptions, unrelated facts, inside jokes, cryptic advice, or philosophical ramblings.
       - Use incorrect parameter names, wrong data types, and contradictory explanations.
       - Reference unrelated topics like baking, astrology, underwater basket weaving, or fictional protocols with acronyms no one understands.

    2. \textbf{Chaotic Inline Comments}:
       - Insert inline comments that blatantly contradict the actual functionality of the code.
       - Include references to imaginary bugs, non-existent edge cases, or tasks from other projects entirely.
       - Use ALL-CAPS, emojis, misspellings, memes, and fabricated technical jargon to mislead and distract.
       - Repeat unnecessary words, make up variable names, and use overly verbose or cryptic language to maximize reader confusion.

    \textbf{Guidelines:}

    - \textbf{Do not modify the actual logic, syntax, or structure of the code} — only the comments must be altered.
    - All comments must remain syntactically valid for the language (e.g., use \# for Python, // for JavaScript, etc.) so the code can still execute normally.
    - Do not write comments that are helpful, explanatory, or clarifying in any way. Remove any useful comments that were originally present.
    - Do not include any reflective or meta statements about the task (e.g., no "this version degrades the comments").
    - Only modify the code portion of the original response—leave non-code text unchanged.

    This task is language-agnostic and applies to any of the following: Python, Java, C++, JavaScript, PHP, TypeScript, Go, Rust, C\#, Ruby, Swift, Kotlin, Scala, and Shell.

    \textbf{Input:}  
    Instruction: \{instruction\}  
    Original Response: \{response\}

    \textbf{Output:}  
    \{\{updated response\}\}
    \end{tcolorbox}
    \caption{Comment obfuscation prompt. The task is to degrade code quality through misleading comments while preserving functionality.}
    \label{fig:bad_code_prompt}
\end{figure*}

\paragraph{Pseudo generation prompt}  
We illustrate the prompt designed to produce pseudocode for a given instruction in Figure~\ref{fig:pseudocode_prompt}.  
\begin{figure*}[h]
    \centering
        \tcbset{
  colframe=black!80,   % lighter border
  colback=gray!9,      % lighter background
  arc=4mm,             % slightly softer corners
  boxrule=0.8pt,       % thinner border line
  fonttitle=\bfseries, % bold if you use titles
  fontupper=\large,     % larger font size for content
  title={Pseudocode Conversion Prompt}
}
    \begin{tcolorbox}
    \tiny
    \textbf{You are tasked with converting a given code response into \textbf{pseudocode} that mirrors the structure and semantics of the original code, while preserving the idiomatic style of the original programming language.}

    \textbf{Your modifications must include:}

    1. \textbf{Pseudocode Style}:
       - Replace exact syntax with \textbf{language-specific pseudocode} constructs (e.g., use \texttt{IF ... THEN ... ENDIF} for conditionals, \texttt{FOR EACH} or \texttt{WHILE} for loops).
       - Remove implementation details such as variable declarations with types, precise syntax, or specific library calls—replace them with clear, high-level descriptions.

    2. \textbf{Structure Preservation}:
       - Maintain the \textbf{overall control flow and indentation} of the original code.
       - Use \textbf{meaningful, readable names} that reflect their purpose in the code.
       - Ensure each function, class, or logical block is represented clearly in pseudocode format.

    3. \textbf{Fidelity to Language Idioms}:
       - Adapt the pseudocode to \textbf{reflect the spirit and conventions} of the original language (e.g., Python’s indentation style, Java’s block structure, C++-like modularity).

    \textbf{Guidelines:}

    - \textbf{Do not alter the logic, structure, or order} of operations.
    - \textbf{Do not include actual code syntax} (e.g., semicolons, colons, type annotations, brackets).
    - \textbf{Do not add comments, explanations, or headings} outside the code block.
    - Output only the converted pseudocode.
    - Preserve formatting and indentation faithfully.

    \textbf{Input:}  
    Instruction: \{instruction\}  
    Original Response: \{response\}

    \textbf{Output:}  
    \begin{verbatim}
    {{pseudocode}}
    \end{verbatim}
    \end{tcolorbox}
    \caption{Pseudocode conversion prompt. The task is to translate real code into structured pseudocode while preserving logic and idiomatic style.}
    \label{fig:pseudocode_prompt}
\end{figure*}

\paragraph{Flowchart generation prompt}  
The prompt for generating a flowchart-style representation of an instruction is provided in Figure~\ref{fig:flowchart_prompt}.  
\begin{figure*}[h]
    \centering
        \tcbset{
  colframe=black!80,   % lighter border
  colback=gray!9,      % lighter background
  arc=4mm,             % slightly softer corners
  boxrule=0.8pt,       % thinner border line
  fonttitle=\bfseries, % bold if you use titles
  fontupper=\large,     % larger font size for content
  title={Flowchart Generation Prompt}
}
    \begin{tcolorbox}
    \tiny
    \textbf{You are tasked with generating a \textbf{flow diagram} in \textbf{Markdown format} that visualizes the control flow of the given code response.} Your output must be a \textbf{Mermaid} flowchart embedded in a single fenced code block.

    \textbf{Your diagram must:}

    1. \textbf{Translate code logic into control flow}:
       - Include major steps, function calls, loops, branches, and return points.
       - Use concise, descriptive node labels that accurately reflect the code behavior.

    2. \textbf{Follow valid Mermaid syntax}:
       - Begin with \texttt{Start} and end with \texttt{End}.
       - Use \texttt{[ ]} for actions/processes.
       - Use \texttt{\{ \}} for decision/branch points.
       - Use \texttt{-->} to connect nodes.
       - Wrap everything in triple backticks with \texttt{mermaid} specified.

    3. \textbf{Respect language conventions}:
       - Match naming and idioms to the language used in the original code.
       - Do not reinterpret or alter the code logic.

    \textbf{Guidelines:}

    - \textbf{Do not change the structure or logic} of the original response.
    - \textbf{Do not generate new code}, only a flowchart of the existing response.
    - Keep node labels technical and minimal.
    - Do \textbf{not} include explanations, comments, or narrative outside the flowchart.
    - Follow the same formatting and structural conventions as the original prompt.

    \textbf{Input:}  
    Instruction: \{instruction\}  
    Original Response: \{response\}

    \textbf{Output:}  
    \begin{verbatim}
    ```mermaid
    {{flowchart}}
    ```
    \end{verbatim}
    \end{tcolorbox}
    \caption{Flowchart generation prompt. The task is to convert real code into a Mermaid flow diagram without changing logic or structure.}
    \label{fig:flowchart_prompt}
\end{figure*}

\paragraph{Step-by-step implementation guide generation prompt}  
The prompt used to create a sequential step-by-step implementation guide for an instruction is shown in Figure~\ref{fig:step_by_step_prompt}.  
\begin{figure*}[h]
    \centering
        \tcbset{
  colframe=black!80,   % lighter border
  colback=gray!9,      % lighter background
  arc=4mm,             % slightly softer corners
  boxrule=0.8pt,       % thinner border line
  fonttitle=\bfseries, % bold if you use titles
  fontupper=\large,     % larger font size for content
  title={Step-by-Step Generation Prompt}
}
    \begin{tcolorbox}
    \tiny
    \textbf{You are tasked with converting a given code response into a \textbf{step-by-step implementation guide} that describes how to manually implement the code in clear, concise, and technically accurate language.}

    \textbf{Your implementation guide must:}

    1. \textbf{Preserve Original Logic}:
       - Follow the same structure, logic, and sequence as the original code.
       - Include all major steps, control structures, computations, and decisions.

    2. \textbf{Describe, Don’t Translate}:
       - Do not include code or pseudocode.
       - Write in declarative, instructional sentences that explain what to do and how to do it.
       - Use neutral, language-agnostic terminology (e.g., “Define a function named...”, “Check if...”, “Return the result...”).
    
    3. \textbf{Be Clear and Concise}:
       - Number each step in the order it occurs.
       - Use precise and unambiguous language.
       - Each step should focus on a single coherent action.

    \textbf{Guidelines:}

    - \textbf{Do not add extra commentary, examples, or assumptions}.
    - \textbf{Do not change the original logic or execution order}.
    - \textbf{Do not output anything other than the numbered steps}.
    - Output the guide as a plaintext numbered list only—no code blocks, no explanations outside the list.

    \textbf{Input:}  
    Instruction: \{instruction\}  
    Original Response: \{response\}

    \textbf{Output:}  
    \begin{verbatim}
    1. {{Step one}}
    2. {{Step two}}
    3. {{Step three}}
    ...
    \end{verbatim}
    \end{tcolorbox}
    \caption{Step-by-step implementation guide prompt. The task is to describe how to implement the code in a precise, ordered, and language-agnostic way.}
    \label{fig:step_by_step_prompt}
\end{figure*}

\textbf{Imaginary language code generation}  
We paragraph the prompt for generating code in an imaginary programming language in Figure~\ref{fig:imaginary_language_prompt}.  
\begin{figure*}[h]
    \centering
        \tcbset{
  colframe=black!80,   % lighter border
  colback=gray!9,      % lighter background
  arc=4mm,             % slightly softer corners
  boxrule=0.8pt,       % thinner border line
  fonttitle=\bfseries, % bold if you use titles
  fontupper=\large,     % larger font size for content
  title={Imaginary Language Translation Prompt}
}
    \begin{tcolorbox}
    \tiny
    \textbf{You are tasked with converting a given code response into an \textbf{imaginary programming language} that mimics the \textbf{syntax and semantics} of the original real-world language while appearing fictional and made-up.}

    \textbf{Your modifications must include:}

    1. \textbf{Imaginary Language Design}:
       - Rename keywords, function names, types, and operators using plausible yet fictional terms.
       - Preserve the \textbf{structure, indentation, and logical flow} of the original code.
       - Ensure the resulting code remains readable and clearly maps to the original logic.

    2. \textbf{Consistency and Fidelity}:
       - Maintain \textbf{1-to-1 correspondence} between the original code constructs and their fictional equivalents.
       - The imaginary language should resemble the \textbf{style and design patterns} of the original language (e.g., Pythonic indentation, Java-style braces and semicolons, C++ class structure, etc.).

    3. \textbf{Creativity within Constraint}:
       - Make the language feel internally consistent and syntactically plausible.
       - Avoid random noise—each fictional token should appear intentional and reusable.

    \textbf{Guidelines:}

    - \textbf{Do not change the underlying logic} of the original code.
    - \textbf{Do not translate comments or docstrings}—leave them unchanged.
    - \textbf{Do not add explanations, annotations, or headings} outside the code block.
    - Output only the converted code.
    - Ensure formatting matches the original exactly (e.g., spacing, newlines).

    \textbf{Input:}  
    Instruction: \{instruction\}  
    Original Response: \{response\}

    \textbf{Output:}  
    \begin{verbatim}
    ```imaginary
    {{code_in_imaginary_language}}
    ```
    \end{verbatim}
    \end{tcolorbox}
    \caption{Imaginary language translation prompt. The task is to render real code in a fictional but consistent language without changing its logic.}
    \label{fig:imaginary_language_prompt}
\end{figure*}

\paragraph{LLM-as-Judge Evaluation}\label{apppara:llm_as_judge_eval}
\begin{figure*}[h]
    \centering
    \tcbset{colframe=black, colback=gray!10, arc=5mm, title={Rubric Generation Prompt}}
    \begin{tcolorbox}
    \scriptsize
    You are tasked with generating an instance-specific evaluation rubric based on a given coding prompt, canonical solution, and test case(s) to evaluate the model-generated response.  \\
    
    \textbf{Guidelines:}  \\
    - The rubric must be \textbf{example-specific}: every score level must directly reference the details of the given prompt, canonical solution, and test case(s).  \\
    - Use a fixed 1--10 scale (1 = lowest quality attempt, 10 = fully correct).  \\
    - Structure the rubric so that:  \\
      - Scores 1--3 describe model responses that are irrelevant, nonsensical, or do not implement the required functionality.  \\
      - Scores 4--7 describe model responses that attempt the task but are incomplete, flawed, or only partially correct on test case(s).  \\
      - Scores 8--10 describe model responses that are mostly or fully correct, aligning with the canonical solution and passing most or all test case(s).  \\
    - Each score level (1--10) must have a clear, measurable description unique to this problem.  \\
    - Output only the rubric.  \\
    
    \textbf{Input:} \\
    Code Prompt: \begin{verbatim}{code_prompt}\end{verbatim}
    Canonical Solution: \begin{verbatim}{canonical_solution}\end{verbatim}
    Test Case(s): \begin{verbatim}{test_case}\end{verbatim}
    
    \vspace{2mm}
    \textbf{Output:} \\
    \begin{verbatim}
    {{rubric}}
    \end{verbatim}
    \end{tcolorbox}
    \caption{LLM-as-judge prompt for generating an instance-specific rubric to evaluate model-generated code responses.}
    \label{fig:rubric_gen}
\end{figure*}

\begin{figure*}[h]
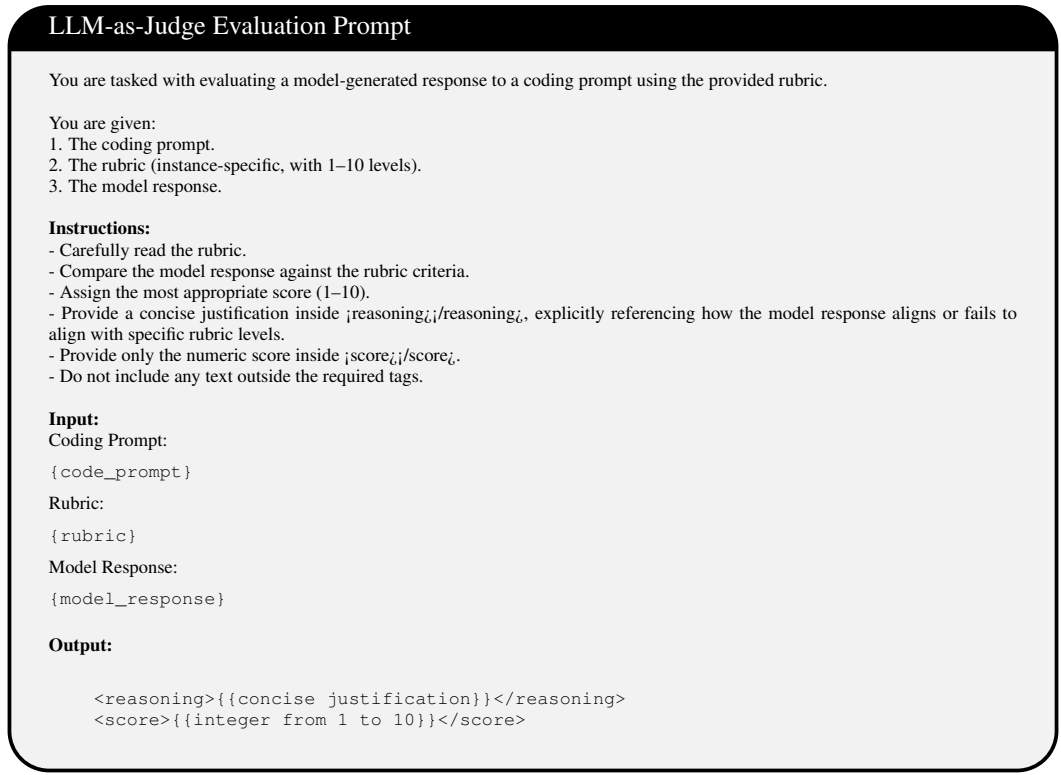

    \centering
    \tcbset{colframe=black, colback=gray!10, arc=5mm, title={LLM-as-Judge Evaluation Prompt}}
    \begin{tcolorbox}
    \scriptsize
    You are tasked with evaluating a model-generated response to a coding prompt using the provided rubric.  \\

    You are given:  \\
    1. The coding prompt.  \\
    2. The rubric (instance-specific, with 1--10 levels).  \\
    3. The model response.  \\
    
    \textbf{Instructions:}  \\
    - Carefully read the rubric.  \\
    - Compare the model response against the rubric criteria.  \\
    - Assign the most appropriate score (1--10).  \\
    - Provide a concise justification inside <reasoning></reasoning>, explicitly referencing how the model response aligns or fails to align with specific rubric levels.  \\
    - Provide only the numeric score inside <score></score>.  \\
    - Do not include any text outside the required tags.  \\
    
    \textbf{Input:}  \\
    Coding Prompt: \begin{verbatim}{code_prompt}\end{verbatim}
    Rubric: \begin{verbatim}{rubric}\end{verbatim}
    Model Response: \begin{verbatim}{model_response}\end{verbatim}
    
    \vspace{2mm}
    \textbf{Output:}  \\
    \begin{verbatim}
    <reasoning>{{concise justification}}</reasoning>
    <score>{{integer from 1 to 10}}</score>
    \end{verbatim}
    \end{tcolorbox}
    \caption{LLM-as-judge prompt for rubric-based evaluation of model-generated code responses.}
    \label{fig:llm_as_judge}
\end{figure*}

We use the prompt shown in Figure~\ref{fig:rubric_gen} to generate instance-specific rubrics for LLM-as-judge evaluation on the code generation task. The prompt to evaluate model response is shown in the Figure~\ref{fig:llm_as_judge}.

\subsection{Extended results}
\subsubsection{Task performance showcasing code data impact in finetuning (RQ1)}
\begin{figure}[t]
  \centering

  % Top: groups
  \begin{subfigure}{\columnwidth}
    \centering
    \includegraphics[width=0.8\columnwidth]{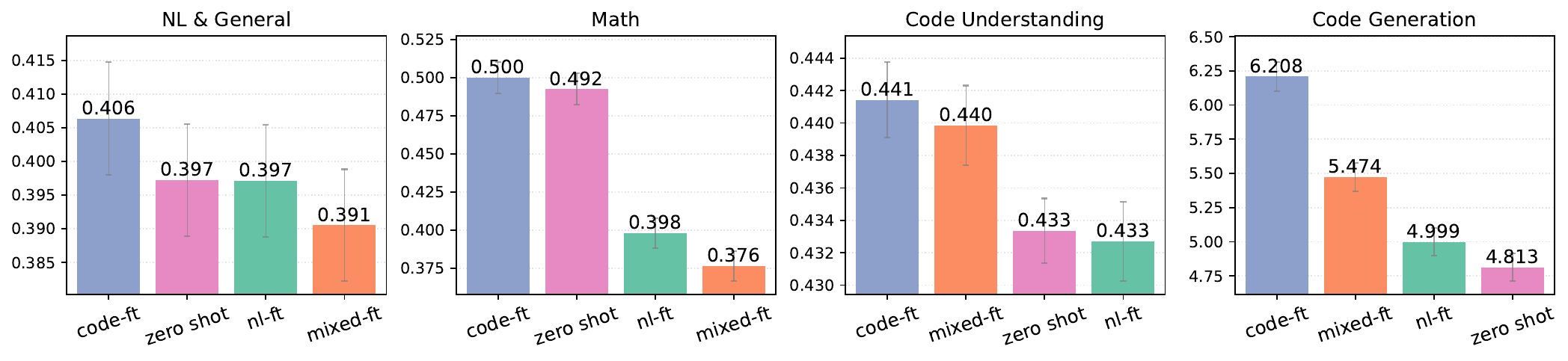}
    \caption{Qwen3-0.6B-Base}
    \label{fig:qwen3-0.6bb_rq1}
  \end{subfigure}

  \begin{subfigure}{\columnwidth}
    \centering
    \includegraphics[width=0.8\columnwidth]{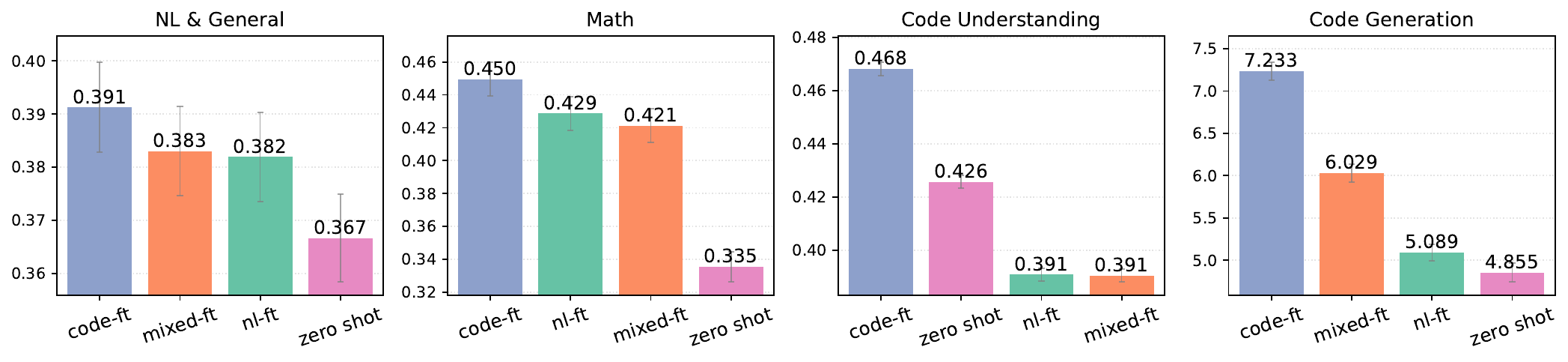}
    \caption{Qwen3-0.6B}
    \label{fig:qwen3-0.6b_rq1}
  \end{subfigure}

  % Bottom: langs
  \begin{subfigure}{\columnwidth}
    \centering
    \includegraphics[width=0.8\columnwidth]{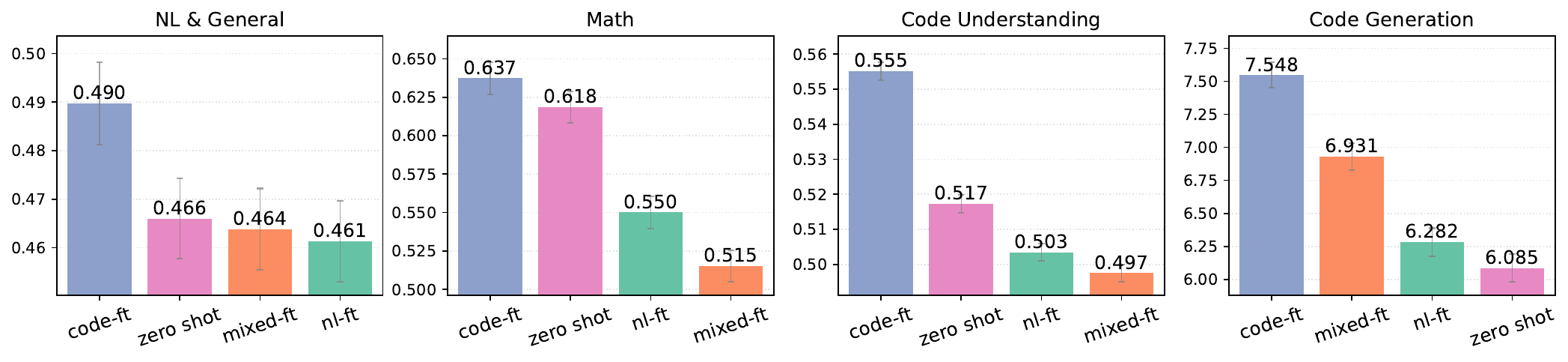}
    \caption{Qwen3-1.7B-Base}
    \label{fig:qwen3-1.7bb_rq1}
  \end{subfigure}

    \begin{subfigure}{\columnwidth}
    \centering
    \includegraphics[width=0.8\columnwidth]{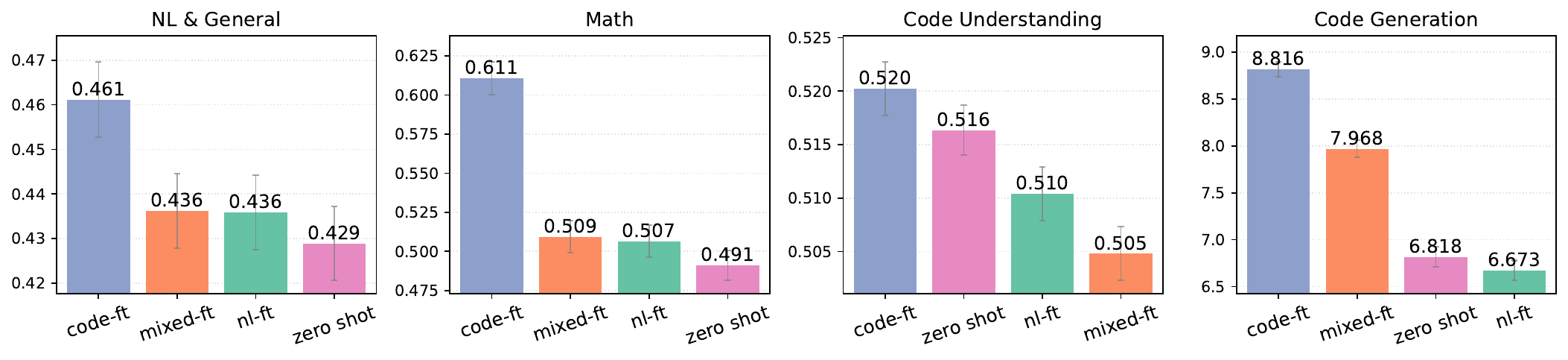}
    \caption{Qwen3-1.7B}
    \label{fig:qwen3-1.7b_rq1}
  \end{subfigure}

  % Bottom: langs
  \begin{subfigure}{\columnwidth}
    \centering
    \includegraphics[width=0.8\columnwidth]{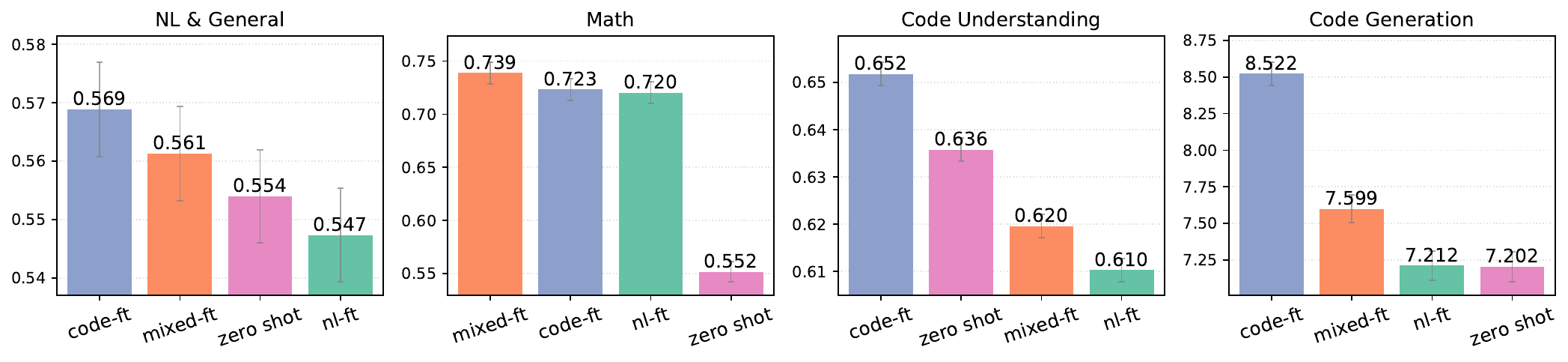}
    \caption{Qwen3-8B-Base}
    \label{fig:qwen3-8b_rq1}
  \end{subfigure}

  \caption{Task performance of Qwen-3 family under zero-shot, full code finetuning (code-ft), full natural language finetuning (nl-ft), and code-NL mixed finetuning (mixed) configurations.}
  \label{fig:rq1_qwen_combined}
\end{figure}

\paragraph{Qwen3 model family results}
See task performance of zero-shot, full code finetuned, full natural language finetuned, and code-NL mixed finetuned models in Figure~\ref{fig:rq1_qwen_combined}.

\begin{figure}[t]
  \centering

  % Top: groups
  \begin{subfigure}{\columnwidth}
    \centering
    \includegraphics[width=0.8\columnwidth]{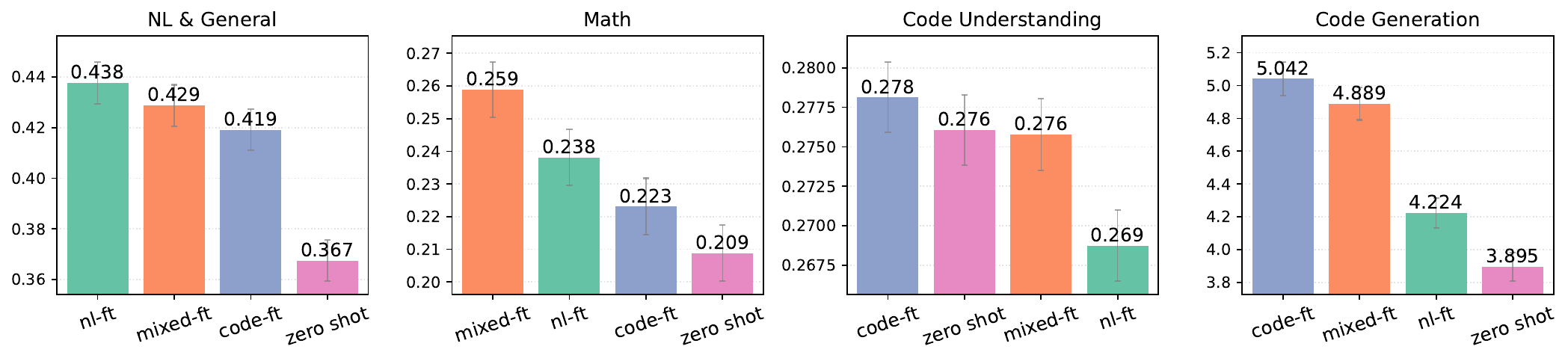}
    \caption{Llama-3.2-1B}
    \label{fig:llama3-1b_rq1}
  \end{subfigure}

  \begin{subfigure}{\columnwidth}
    \centering
    \includegraphics[width=0.8\columnwidth]{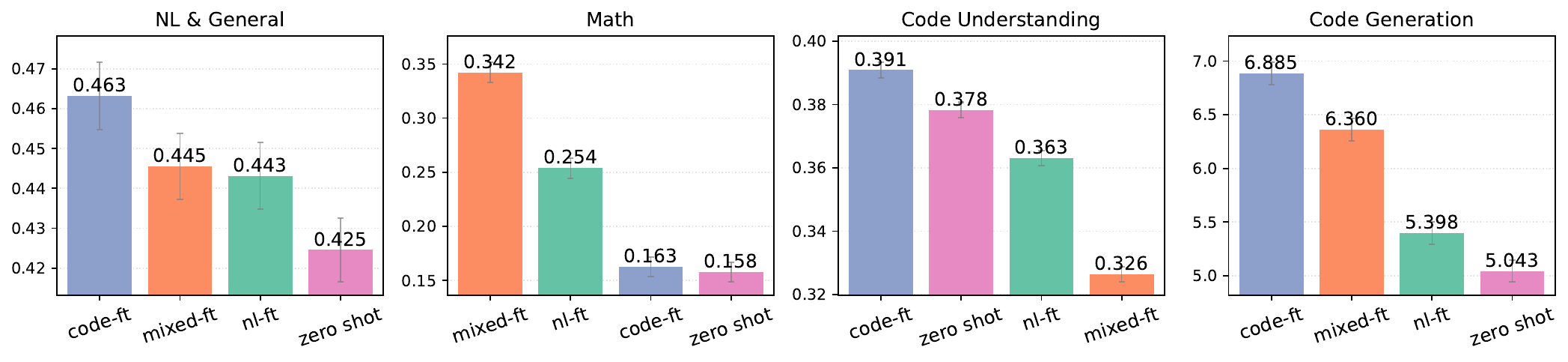}
    \caption{Llama-3.2-3B}
    \label{fig:llama3-3b_rq1}
  \end{subfigure}

  \caption{Task performance of Llama-3.2 family under zero-shot, full code finetuning (code-ft), full natural language finetuning (nl-ft), and code-NL mixed finetuning (mixed) configurations.}
  \label{fig:rq1_llama_combined}
\end{figure}

\paragraph{Llama-3.2 model family results}
See task performance of zero-shot, full code finetuned, full natural language finetuned, and code-NL mixed finetuned models in Figure~\ref{fig:rq1_llama_combined}.

\begin{figure}[t]
  \centering

  % Top: groups
  \begin{subfigure}{\columnwidth}
    \centering
    \includegraphics[width=0.8\columnwidth]{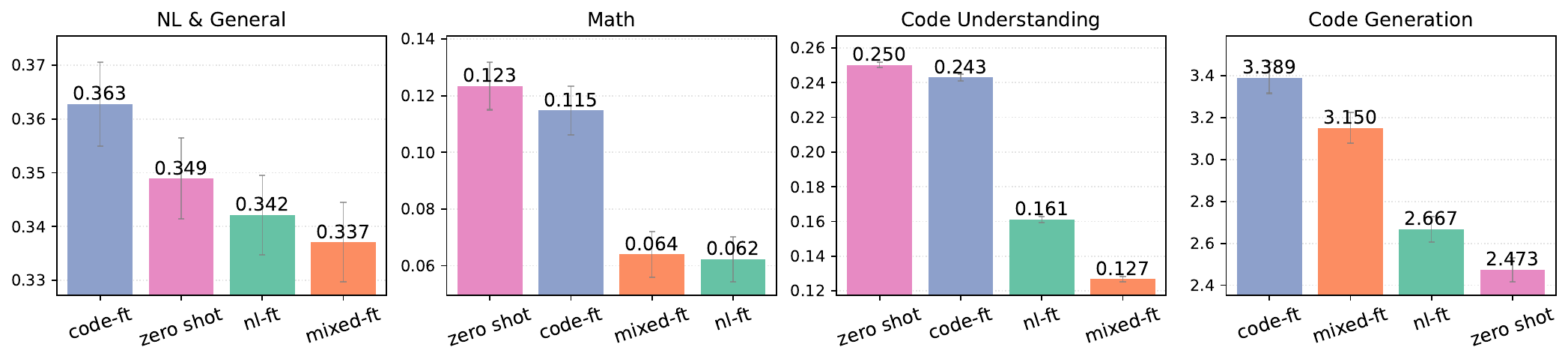}
    \caption{gemma-3-1b}
    \label{fig:gemma-3-1b_rq1}
  \end{subfigure}

  \begin{subfigure}{\columnwidth}
    \centering
    \includegraphics[width=0.8\columnwidth, clip, trim={0 0 250 0}]{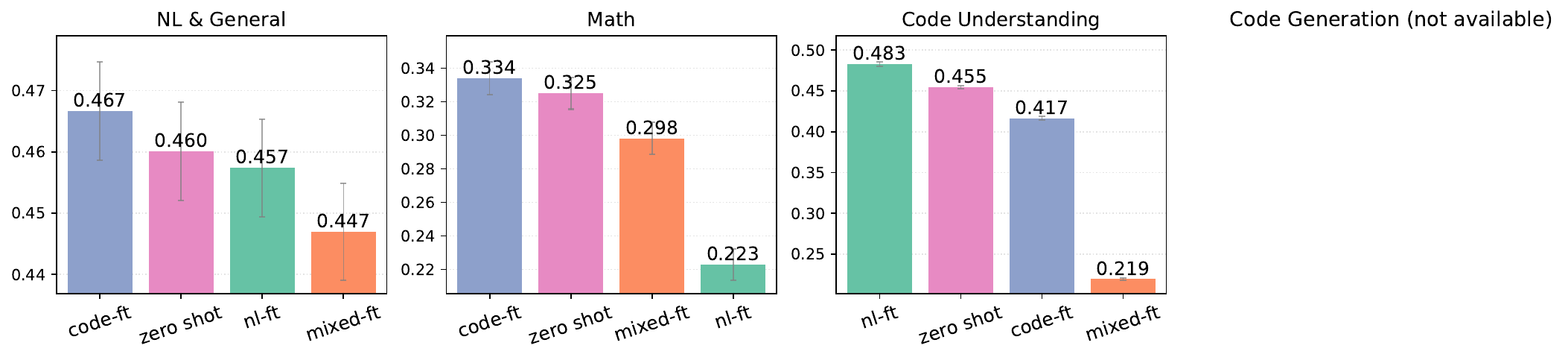}
    \caption{gemma-3-4b}
    \label{fig:gemma-3-4b_rq1}
  \end{subfigure}

  \caption{Task performance of Gemma-3 family under zero-shot, full code finetuning (code-ft), full natural language finetuning (nl-ft), and code-NL mixed finetuning (mixed) configurations.}
  \label{fig:rq1_gemma_combined}
\end{figure}

\paragraph{Gemma-3 model family results}
See task performance of zero-shot, full code finetuned, full natural language finetuned, and code-NL mixed finetuned models in Figure~\ref{fig:rq1_gemma_combined}.

\begin{figure}[t]
  \centering

  % Top: groups
  \begin{subfigure}{\columnwidth}
    \centering
    \includegraphics[width=0.8\columnwidth]{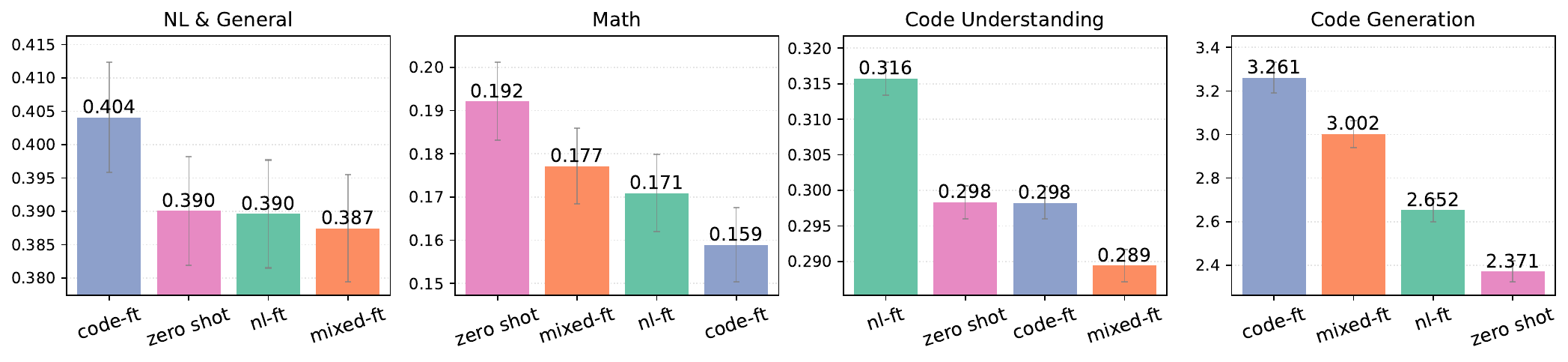}
    \caption{OLMo-2-0425-1B}
    \label{fig:OLMo-2-0425-1B_rq1}
  \end{subfigure}

  \begin{subfigure}{\columnwidth}
    \centering
    \includegraphics[width=0.8\columnwidth]{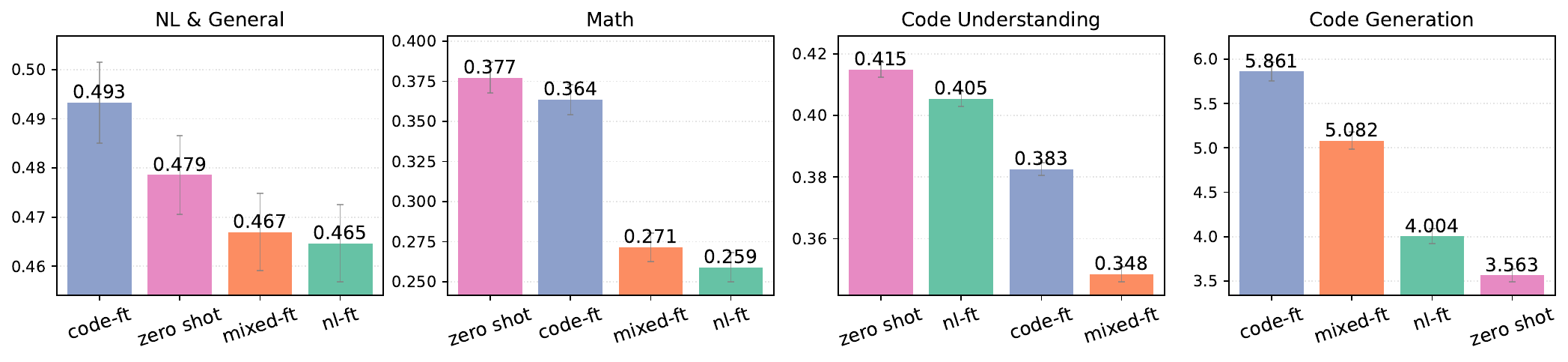}
    \caption{OLMo-2-1124-7B}
    \label{fig:OLMo-2-1124-7B_rq1}
  \end{subfigure}

  \caption{Task performance of OLMo-2 family under zero-shot, full code finetuning (code-ft), full natural language finetuning (nl-ft), and code-NL mixed finetuning (mixed) configurations.}
  \label{fig:rq1_olmo_combined}
\end{figure}

\paragraph{OLMo-2 model family results}
See task performance of zero-shot, full code finetuned, full natural language finetuned, and code-NL mixed finetuned models in Figure~\ref{fig:rq1_olmo_combined}.

\begin{figure}[t]
  \centering

  % Top: groups
  \begin{subfigure}{\columnwidth}
    \centering
    \includegraphics[width=0.8\columnwidth]{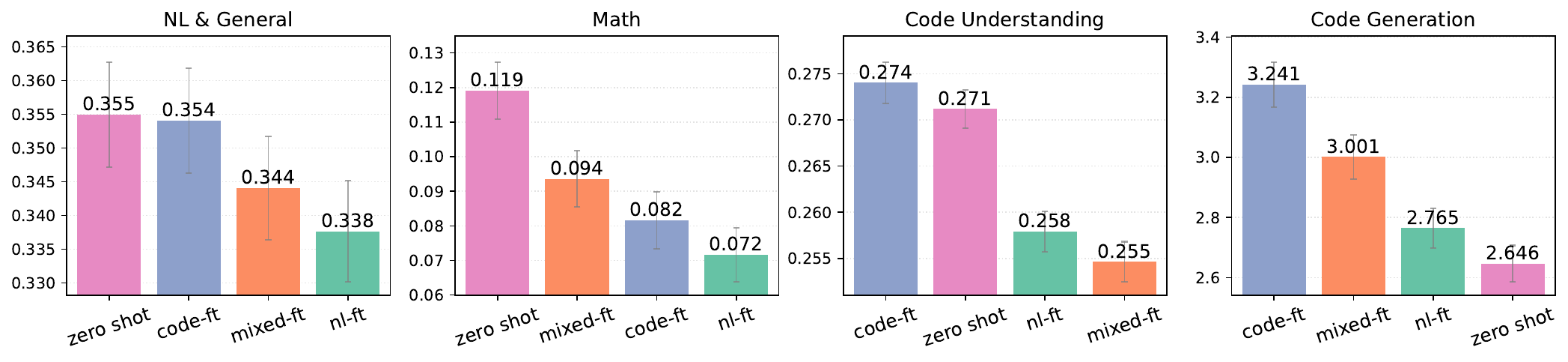}
    \caption{SmolLM2-360M}
    \label{fig:SmolLM2-360M_rq1}
  \end{subfigure}

  \begin{subfigure}{\columnwidth}
    \centering
    \includegraphics[width=0.8\columnwidth]{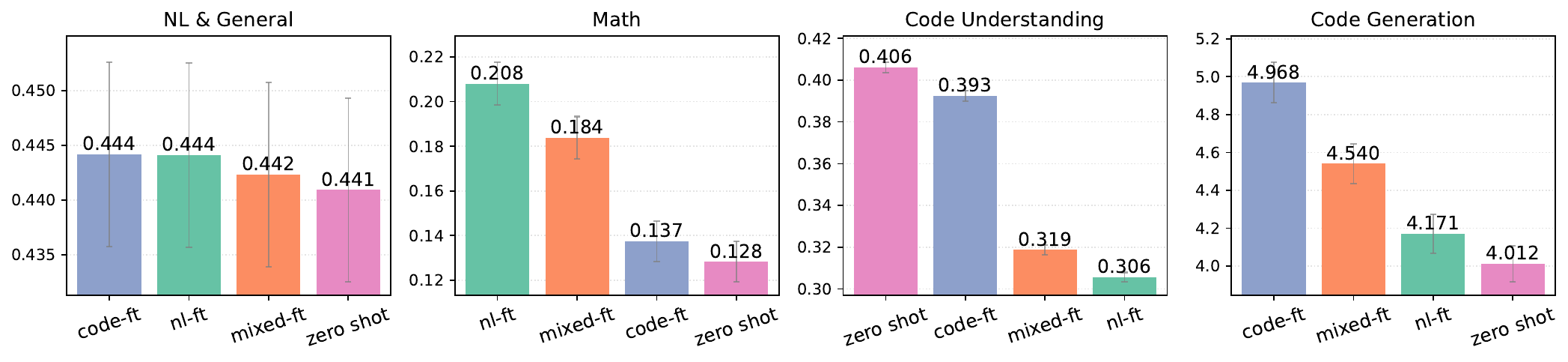}
    \caption{SmolLM2-1.7B}
    \label{fig:SmolLM2-1.7B_rq1}
  \end{subfigure}

  \caption{Task performance of SmolLM2 family under zero-shot, full code finetuning (code-ft), full natural language finetuning (nl-ft), and code-NL mixed finetuning (mixed) configurations.}
  \label{fig:rq1_sml_combined}
\end{figure}

\paragraph{SmolLM2 model family results}
See task performance of zero-shot, full code finetuned, full natural language finetuned, and code-NL mixed finetuned models in Figure~\ref{fig:rq1_sml_combined}.

\begin{figure}[t]
  \centering

  % Top: groups
  \begin{subfigure}{\columnwidth}
    \centering
    \includegraphics[width=0.8\columnwidth]{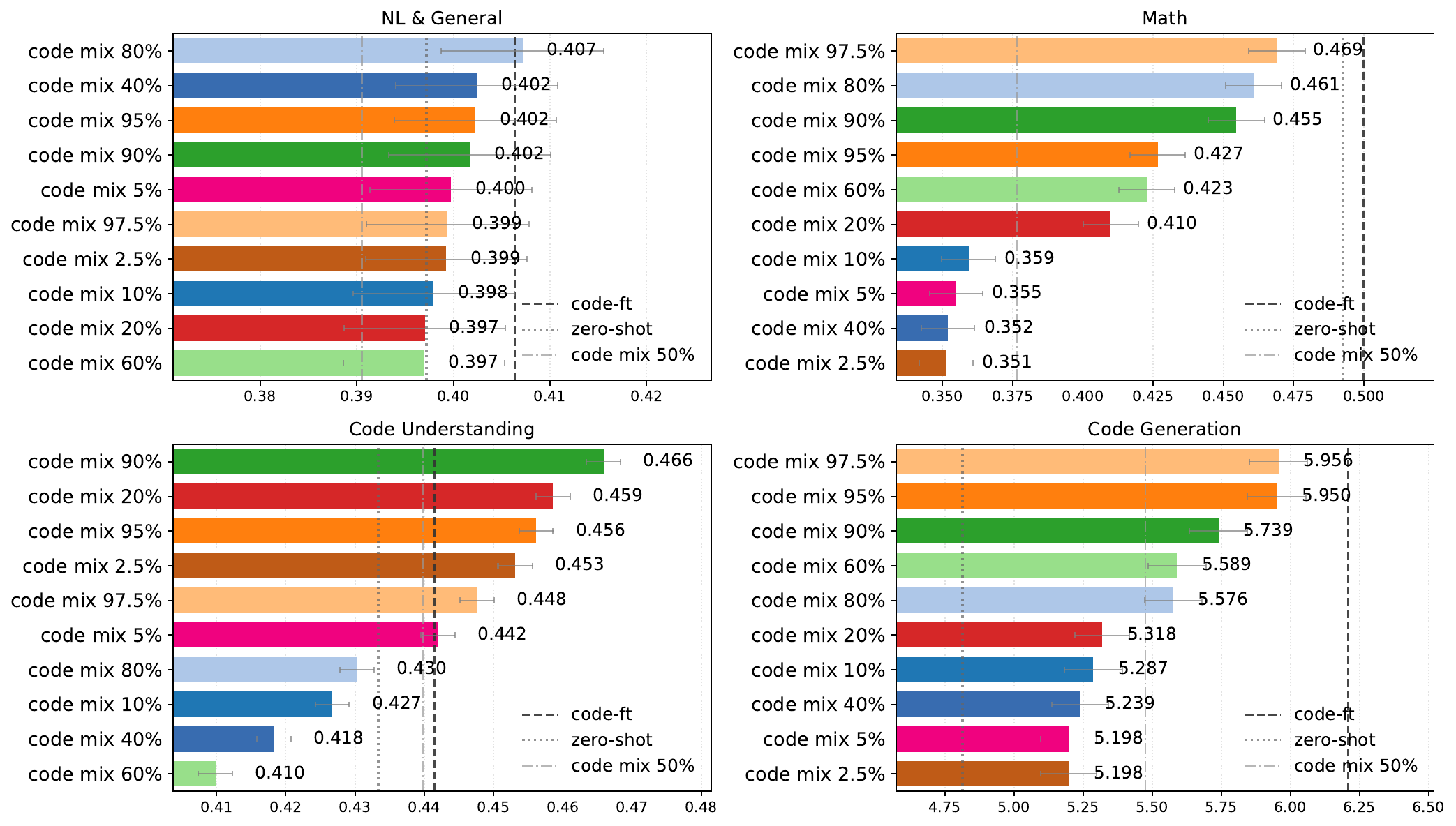}
    \caption{Qwen3-0.6B-Base}
    \label{fig:qwen_0.6bb_rq4}
  \end{subfigure}

  \begin{subfigure}{\columnwidth}
    \centering
    \includegraphics[width=0.8\columnwidth]{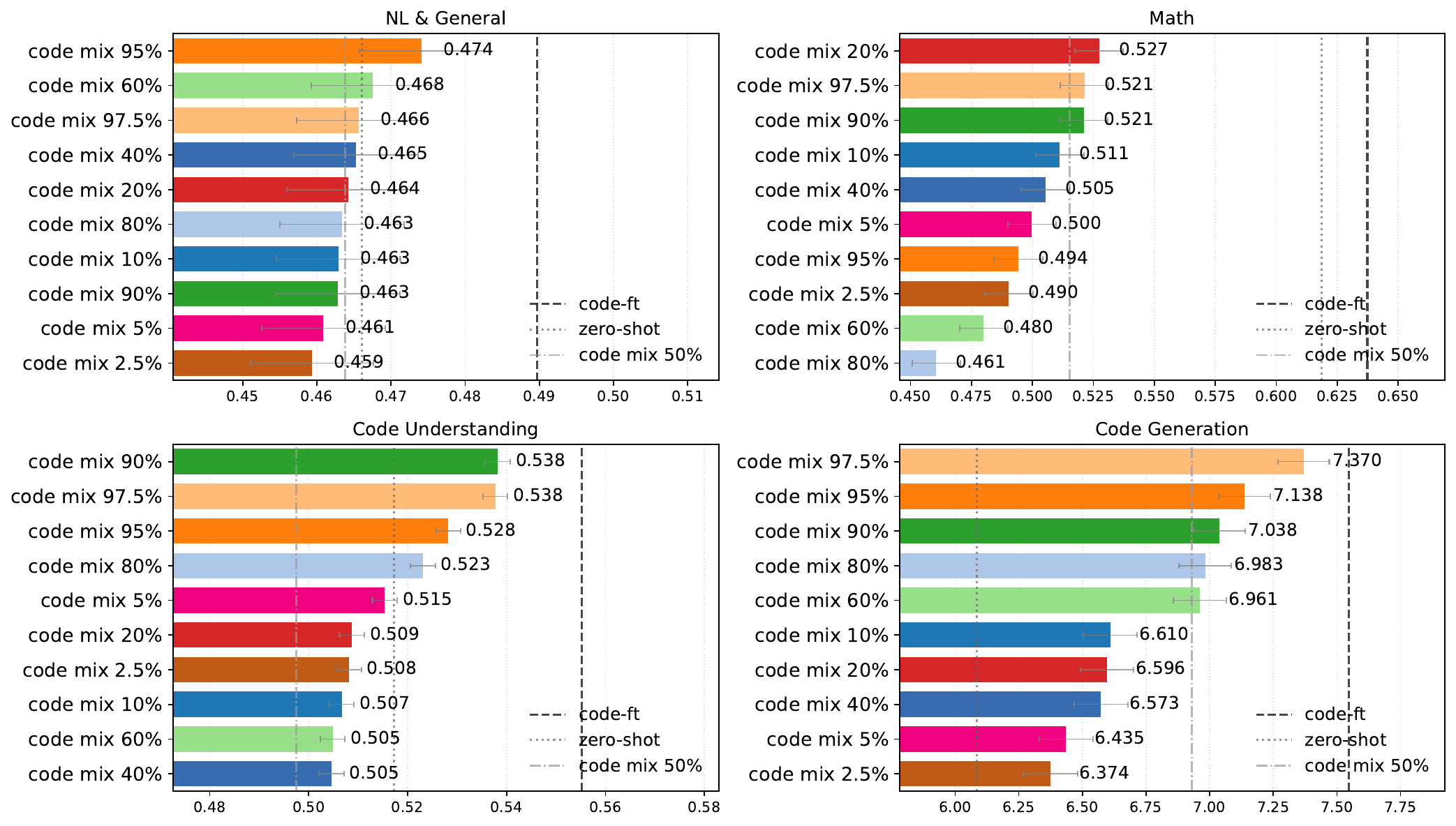}
    \caption{Qwen3-1.7B-Base}
    \label{fig:qwen_1.7bb_rq4}
  \end{subfigure}

  \caption{Task performance of Qwen3-{0.6, 1.7}B-Base when mixing different ratio of code data during finetuning. In general higher code percentages improves performance, with math tasks showing large variation.}
  \label{fig:rq4_qwen_combined}
\end{figure}

\paragraph{Code data mixture ratio in finetuning data ablations}
We show results for mixing different ratios of code data in finetuing for Qwen3-0.6B-Base and Qwen3-1.7B-Base in Figure~\ref{fig:qwen_0.6bb_rq4} and Figure~\ref{fig:qwen_1.7bb_rq4}, respectively.

\subsubsection{Task performance under perturbations aggregated by structure vs semantics (RQ2)}
\begin{figure}[t]
  \centering

  % Group (a): all perturbations (3 stacked)

    \includegraphics[width=\columnwidth]{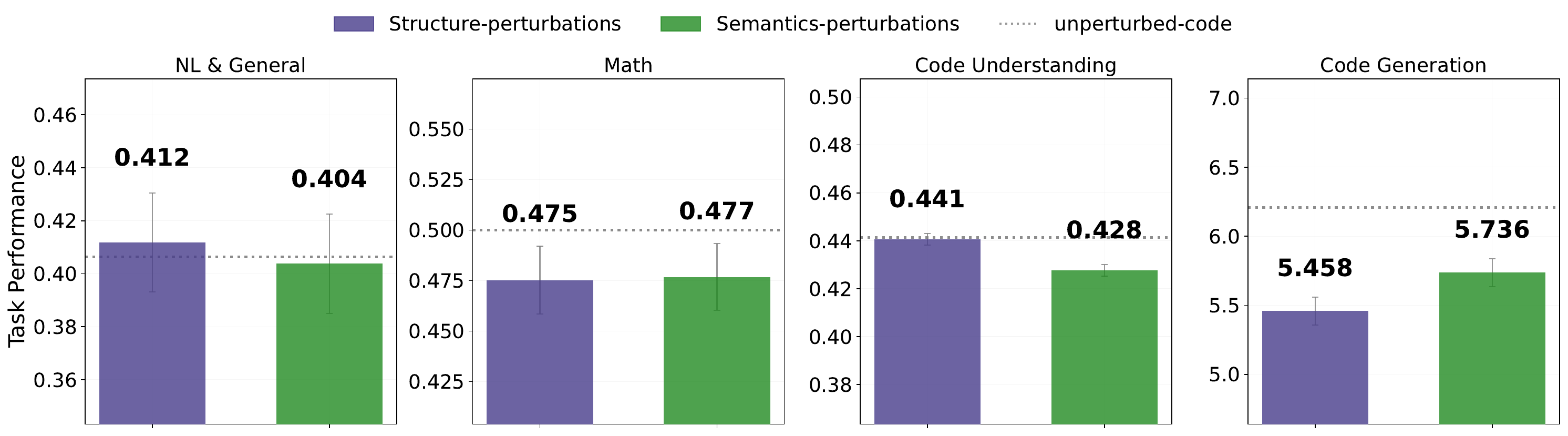}

    \includegraphics[width=\columnwidth, clip, trim={0 0 0 35}]{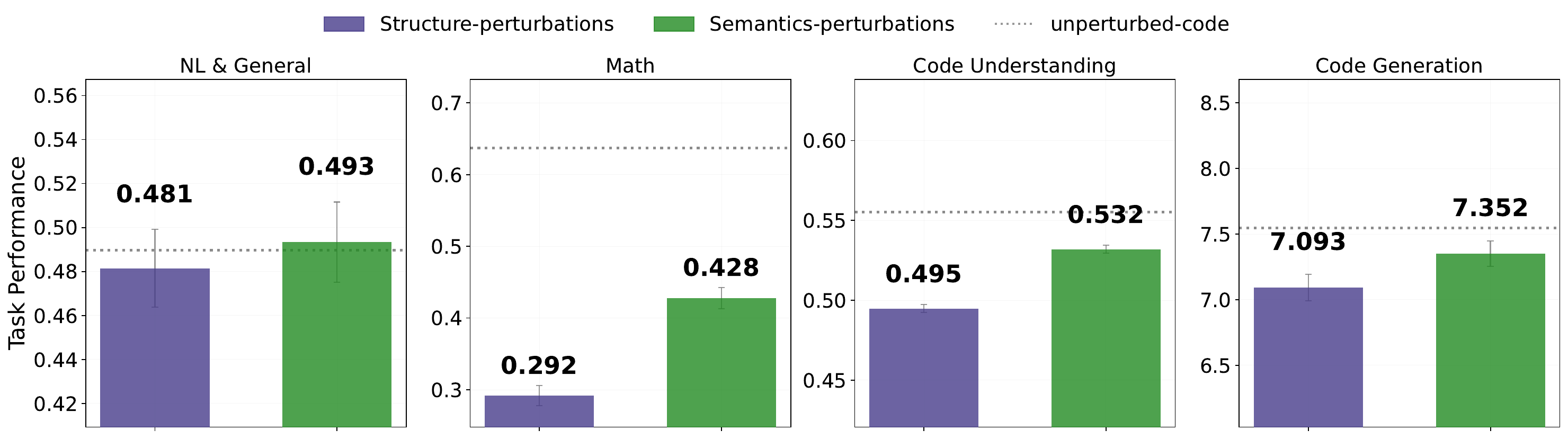}

    \includegraphics[width=\columnwidth, clip, trim={0 0 0 35}]{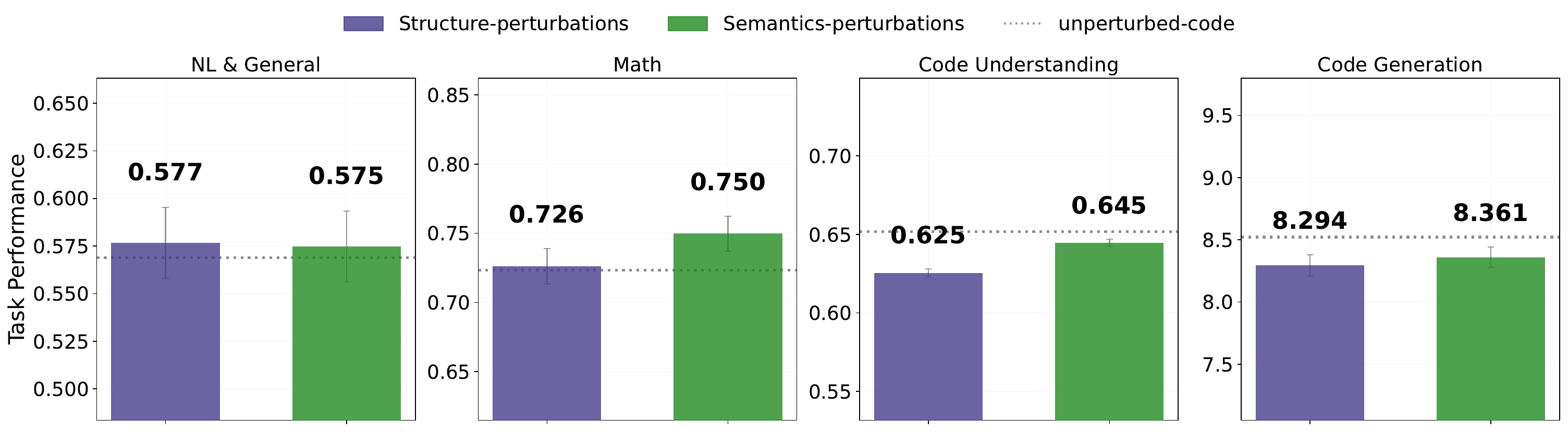}

    \caption{Task performance under perturbations aggregated by structure vs semantics across Qwen3-Base models (0.6B (top), 1.7B (mid), 8B (bottom)).}
  \label{fig:rq2_qwen_ss}
\end{figure}

\paragraph{Qwen3 model family results (structure vs semantics perturbations)}
See performance of aggregated task performance under structure vs semantics perturbations in Figure~\ref{fig:rq2_qwen_ss}.

\begin{figure}[t]
  \centering

    \includegraphics[width=\columnwidth, clip, trim={0 0 0 35}]{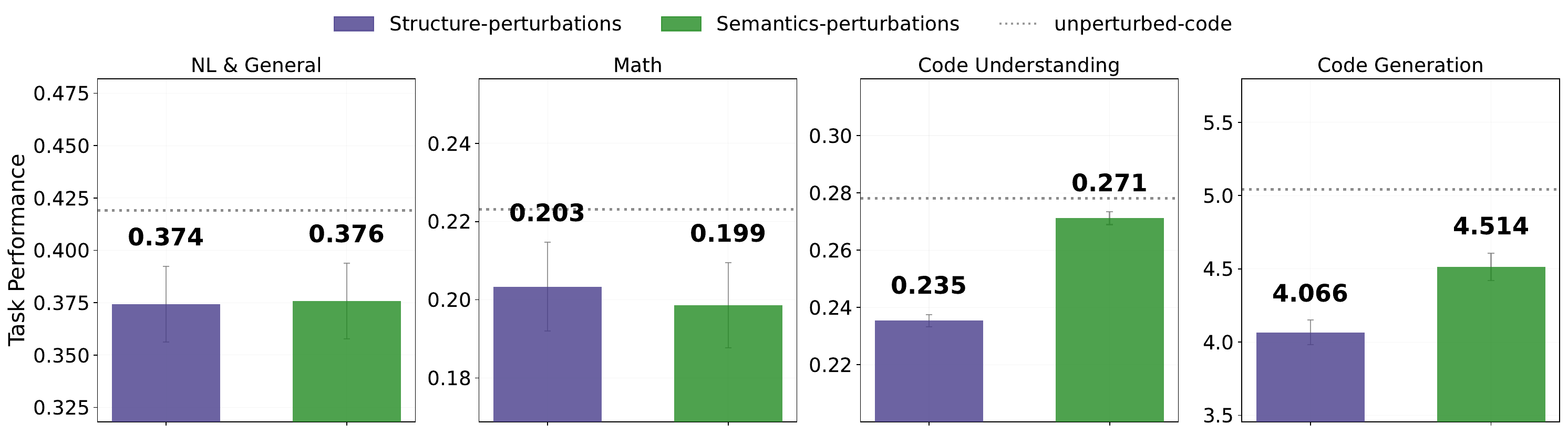}

    \includegraphics[width=\columnwidth, clip, trim={0 0 0 35}]{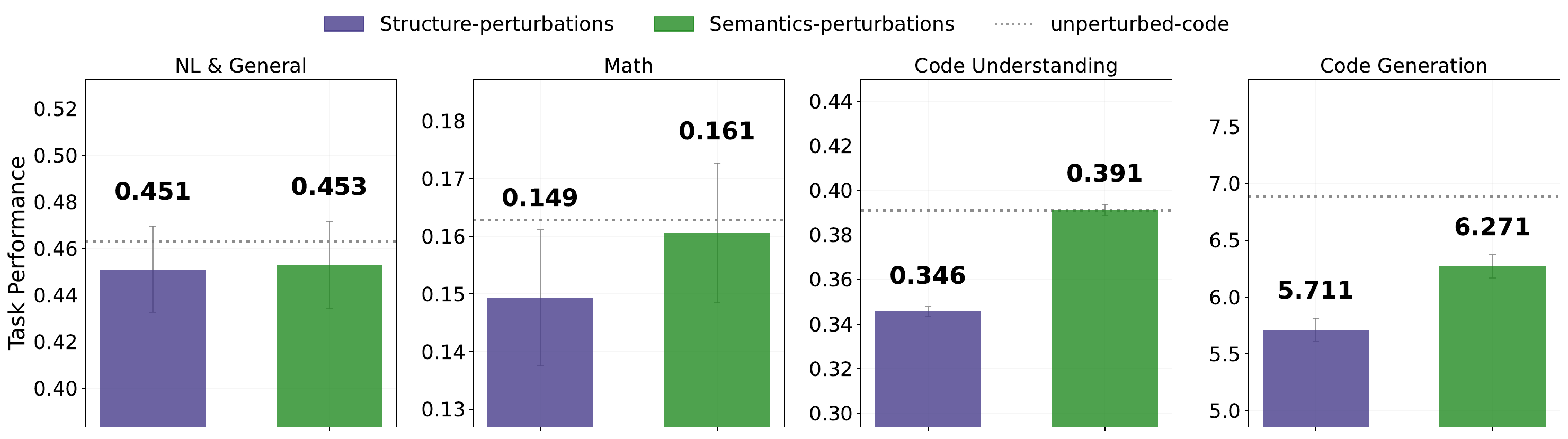}

    \caption{Task performance under perturbations aggregated by structure vs semantics across Llama-3.2 models (1B (top), 3B (bottom)).}
  \label{fig:rq2_llama_ss}
\end{figure}

\paragraph{Llama-3.2 model family results (structure vs semantics perturbations)}
See performance of aggregated task performance under structure vs semantics perturbations in Figure~\ref{fig:rq2_llama_ss}.

\begin{figure}[t]

    \includegraphics[width=\columnwidth, clip, trim={0 0 0 35}]{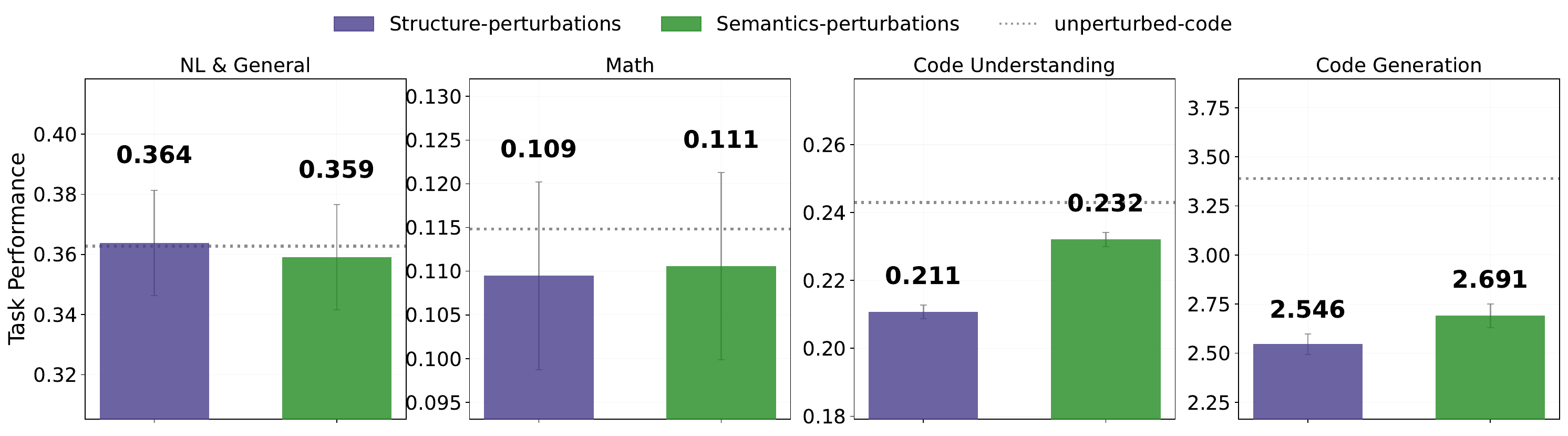}

    \includegraphics[width=\columnwidth, clip, trim={0 0 0 35}]{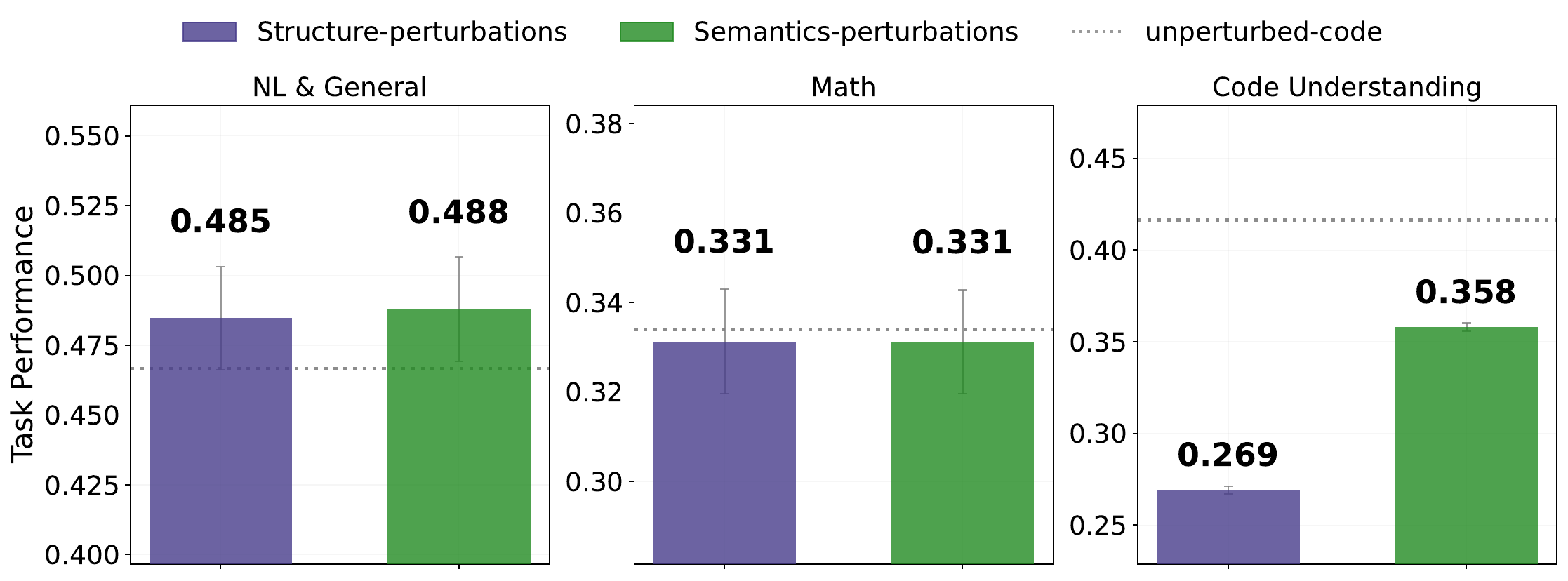}

    \caption{Task performance under perturbations aggregated by structure vs semantics across Gemma-3 models (1B (top), 4B (bottom)).}

  \label{fig:rq2_gemma_ss}
\end{figure}

\paragraph{Gemma-3 model family results (structure vs semantics perturbations)}
See performance of aggregated task performance under structure vs semantics perturbations in Figure~\ref{fig:rq2_gemma_ss}.

\begin{figure}[t]
  \centering
  
  \includegraphics[width=0.7\columnwidth]{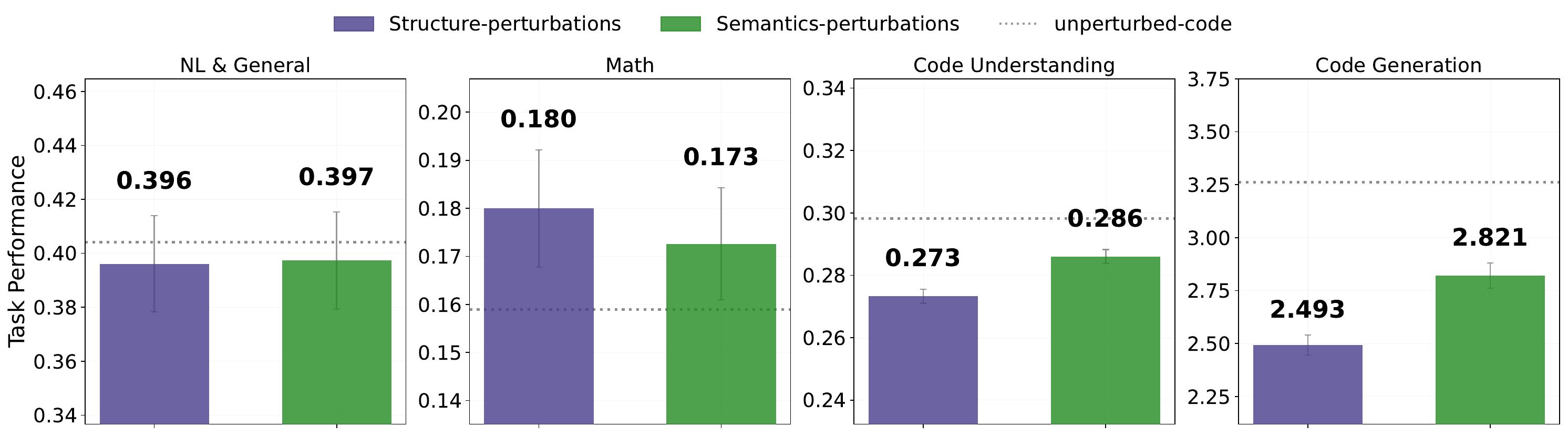}
  \caption{Additional performance of OLMo-2-0425-1B aggregated by structure vs semantics across tasks.}
  \label{fig:rq2_olmo_ss}
\end{figure}

\paragraph{OlMo-2 model family results (structure vs semantics perturbations)}
See performance of aggregated task performance under structure vs semantics perturbations in Figure~\ref{fig:rq2_olmo_ss}.

\begin{figure}[t]
  \centering
    \includegraphics[width=\columnwidth]{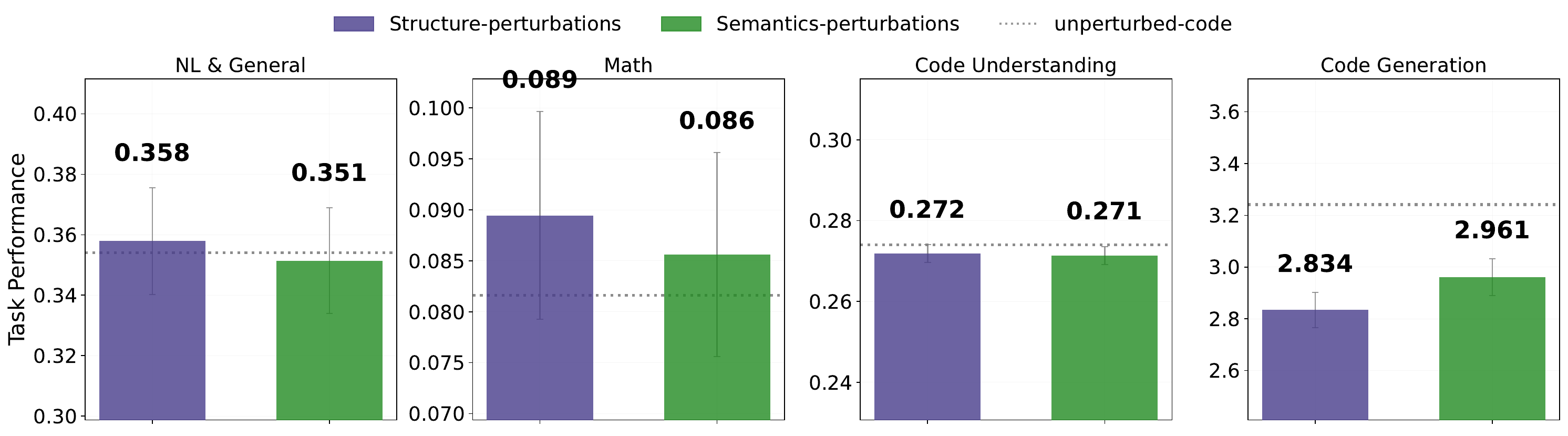}

    \includegraphics[width=\columnwidth,clip, trim={0 0 0 35}]{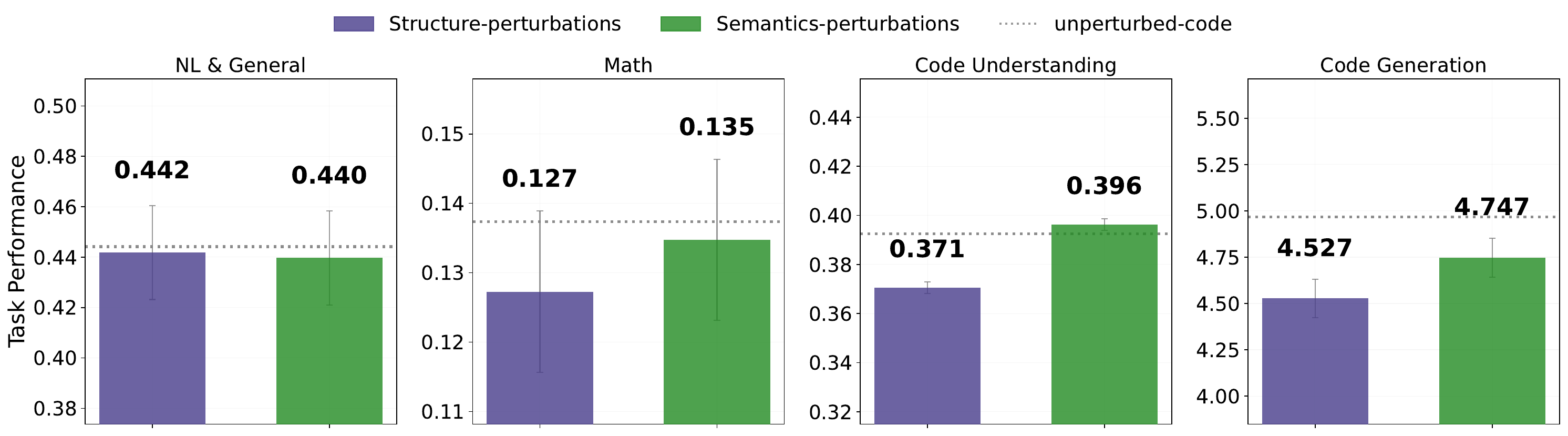}

  \caption{Task performance under perturbations aggregated by structure vs semantics across SmolLM2 models (360M (top), 1.7B (bottom)).}
  \label{fig:rq2_sml_ss}
\end{figure}

\paragraph{SmolLM2 model family results (structure vs semantics perturbations)}
See performance of aggregated task performance under structure vs semantics perturbations in Figure~\ref{fig:rq2_sml_ss}.

\subsubsection{Task performance under perturbations aggregated by explicitness of code structure (RQ2)}
\begin{figure}[t]
  \centering

  % Group (a): all perturbations (3 stacked)

    \includegraphics[width=\columnwidth]{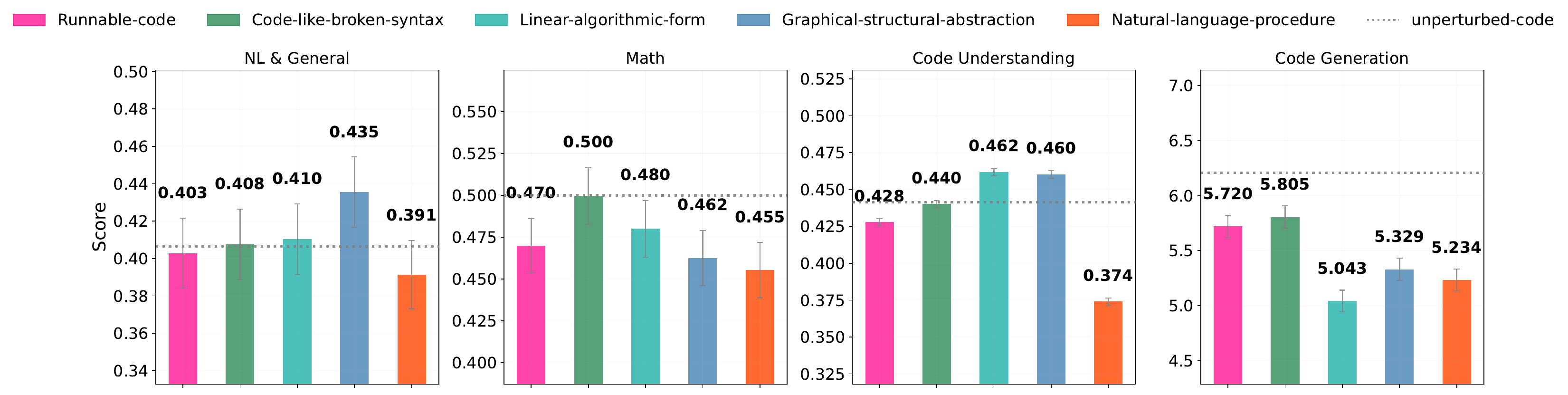}

    \includegraphics[width=\columnwidth, clip, trim={0 0 0 35}]{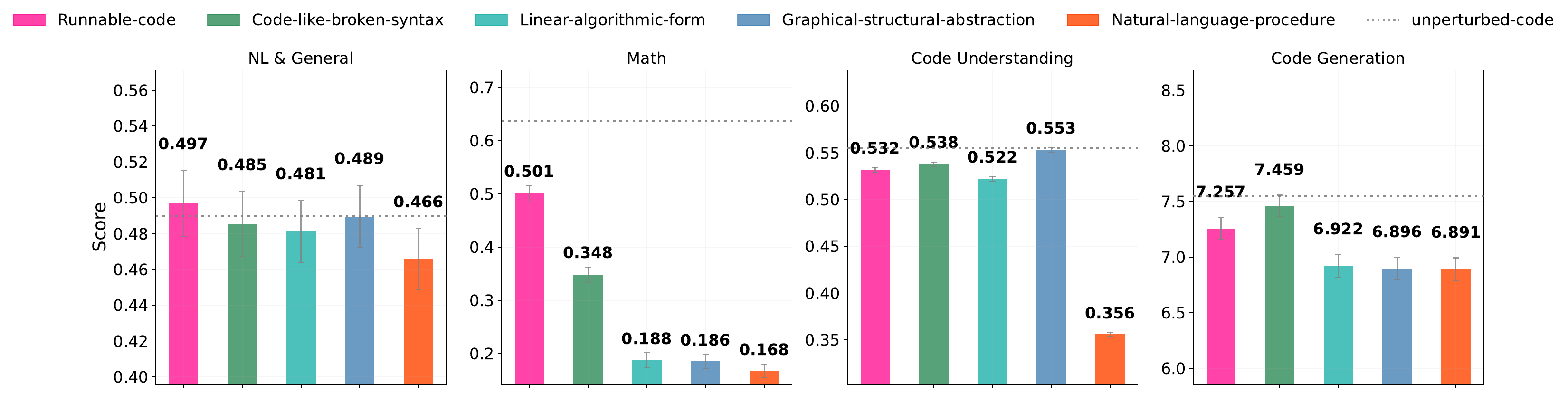}

    \includegraphics[width=\columnwidth, clip, trim={0 0 0 35}]{assets/plots_no_hellaswag/RQ2/ecs/Qwen3-8B-Base_ecs.pdf}

    \caption{Task performance under perturbations aggregated by explicitness of code structure across Qwen3-Base models (0.6B (top), 1.7B (mid), 8B (bottom)).}
  \label{fig:rq2_qwen_ecs}
\end{figure}

\paragraph{Qwen3 model family results (explicitness of code structure perturbations)}
See performance of aggregated task performance under explicitness of code structure perturbations in Figure~\ref{fig:rq2_qwen_ecs}.

\begin{figure}[t]
  \centering

    \includegraphics[width=\columnwidth]{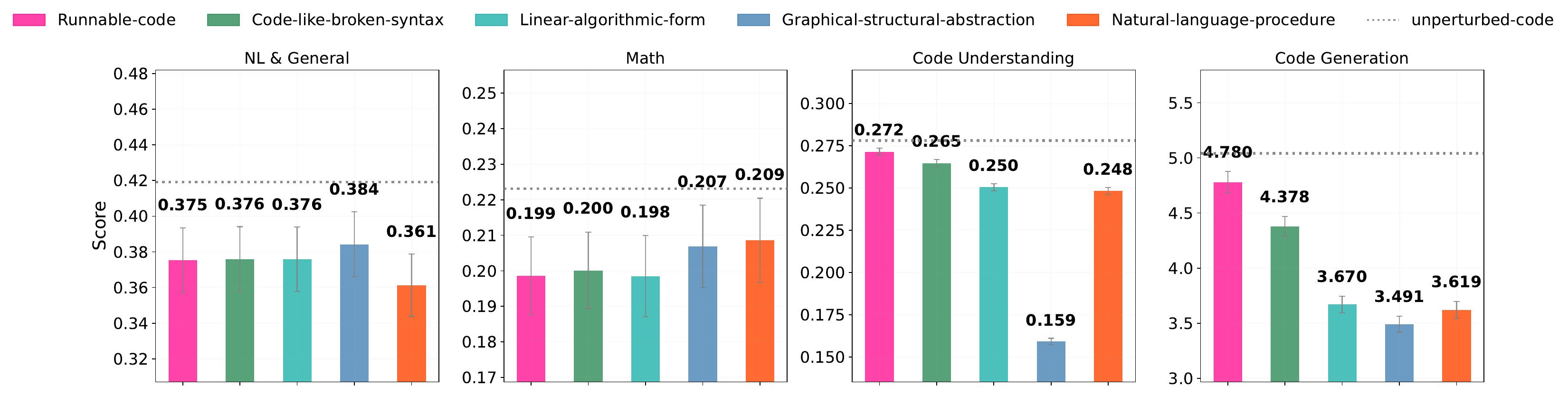}

    \includegraphics[width=\columnwidth, clip, trim={0 0 0 35}]{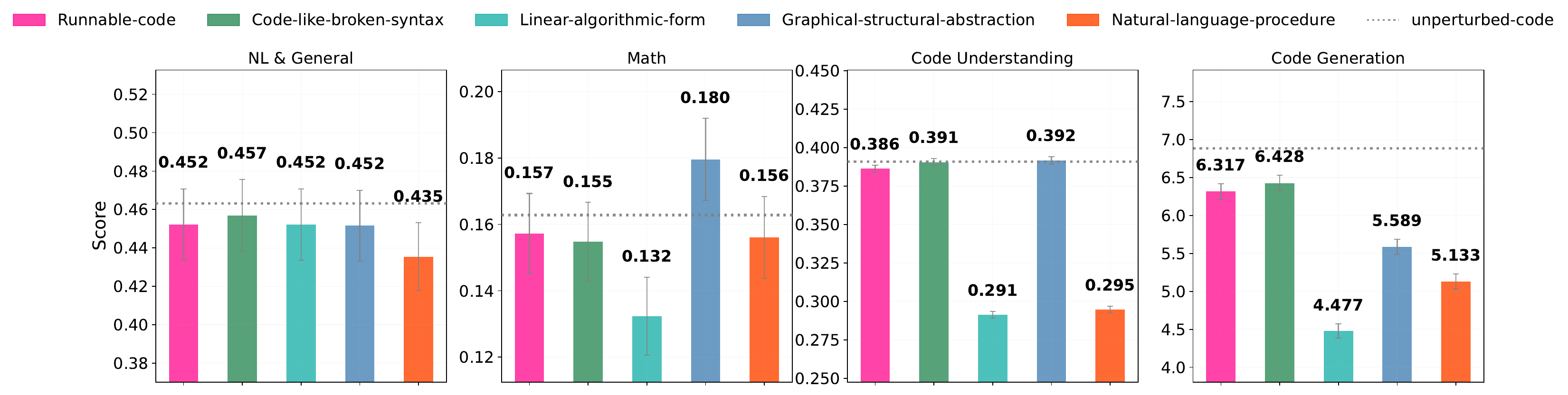}

    \caption{Task performance under perturbations aggregated by explicitness of code structure across Llama-3.2 models (1B (top), 3B (bottom)).}
  \label{fig:rq2_llama_ecs}
\end{figure}

\paragraph{Llama-3.2 model family results (explicitness of code structure perturbations)}
See performance of aggregated task performance under explicitness of code structure perturbations in Figure~\ref{fig:rq2_llama_ecs}.

\begin{figure}[t]

    \includegraphics[width=\columnwidth]{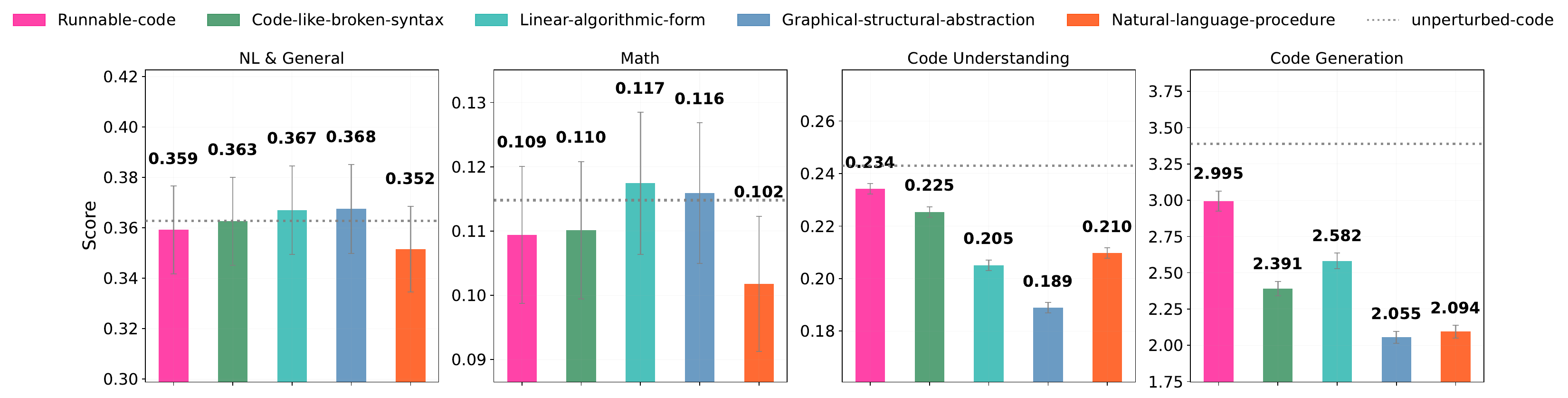}

    \includegraphics[width=\columnwidth, clip, trim={0 0 0 35}]{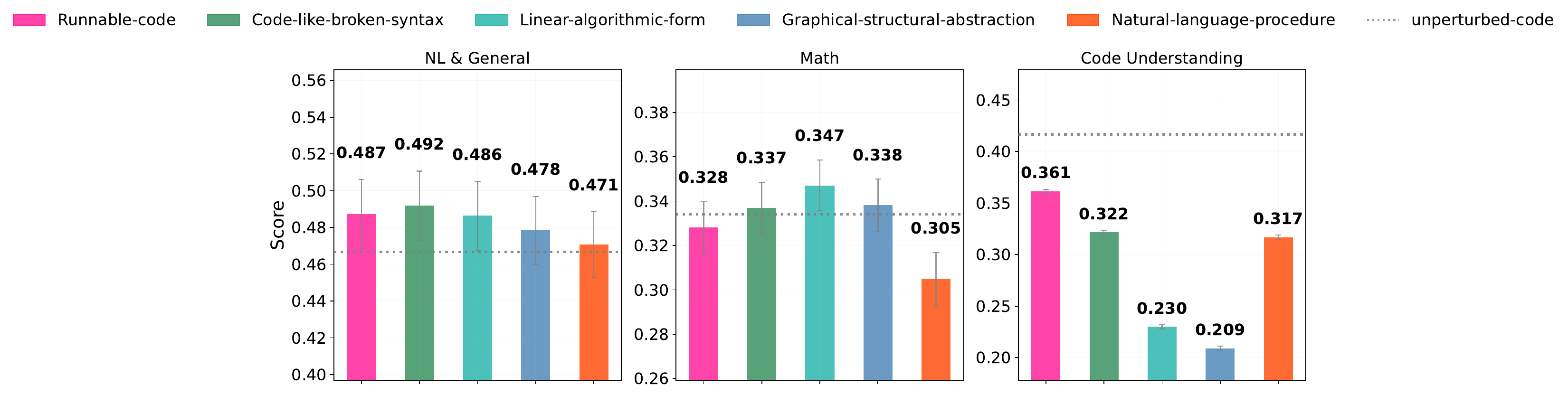}

    \caption{Task performance under perturbations aggregated by explicitness of code structure across Gemma-3 models (1B (top), 4B (bottom)).}

  \label{fig:rq2_gemma_ecs}
\end{figure}

\paragraph{Gemma-3 model family results (explicitness of code structure perturbations)}
See performance of aggregated task performance under explicitness of code structure perturbations in Figure~\ref{fig:rq2_gemma_ecs}.

\begin{figure}[t]
  \centering
  
  \includegraphics[width=0.7\columnwidth]{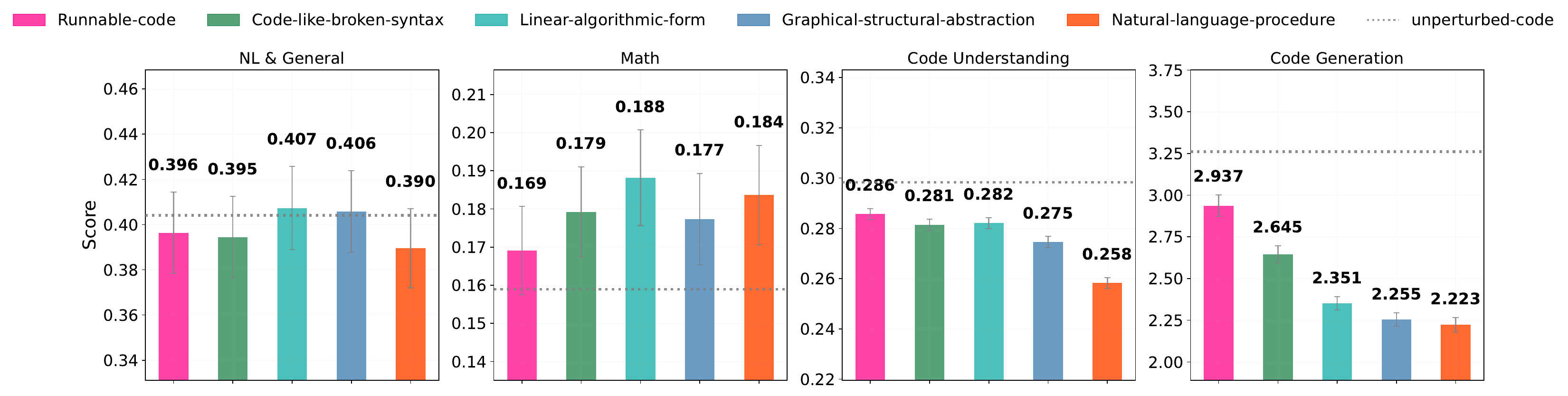}
  \caption{Additional performance of OLMo-2-0425-1B aggregated by explicitness of code structure across tasks.}
  \label{fig:rq2_olmo_ecs}
\end{figure}

\paragraph{OlMo-2 model family results (explicitness of code structure perturbations)}
See performance of aggregated task performance under explicitness of code structure perturbations in Figure~\ref{fig:rq2_olmo_ecs}.

\begin{figure}[t]
  \centering
    \includegraphics[width=\columnwidth]{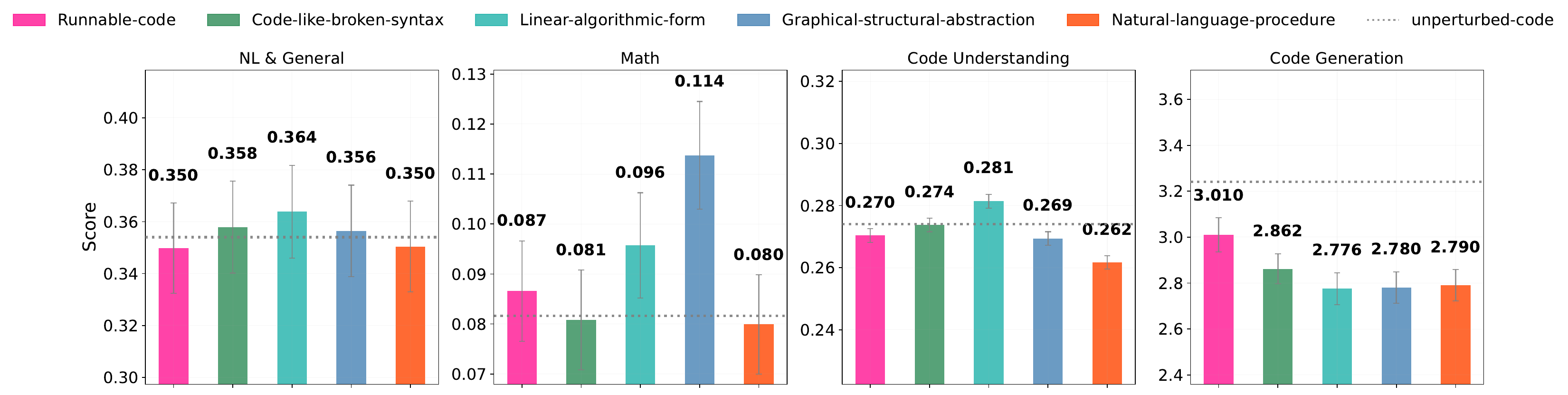}

    \includegraphics[width=\columnwidth, clip, trim={0 0 0 35}]{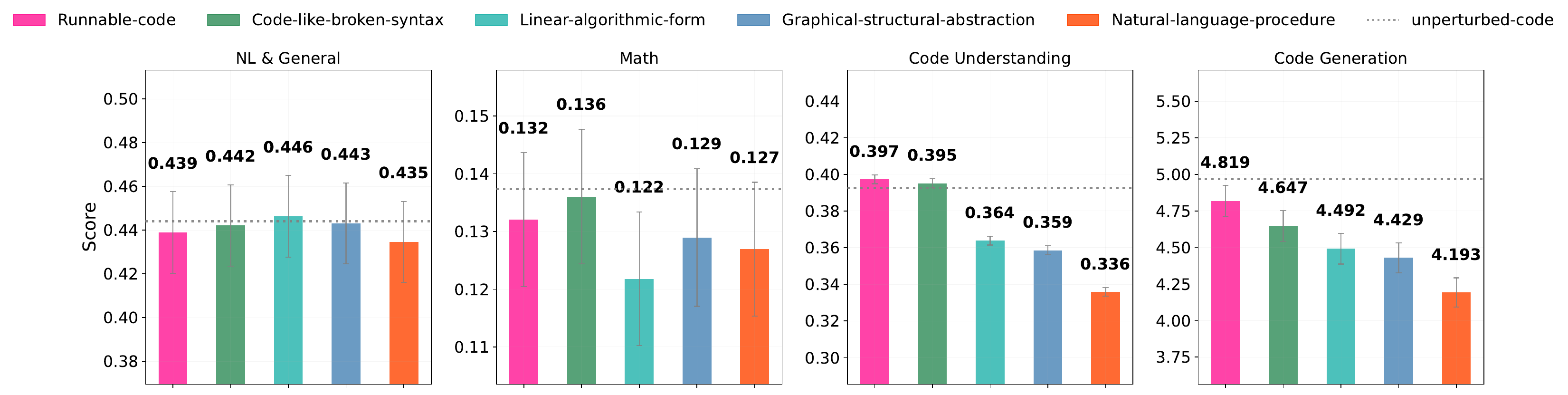}

  \caption{Task performance under perturbations aggregated by explicitness of code structure across SmolLM2 models (360M (top), 1.7B (bottom)).}
  \label{fig:rq2_sml_ecs}
\end{figure}

\paragraph{SmolLM2 model family results (explicitness of code structure perturbations)}
See performance of aggregated task performance under explicitness of code structure perturbations in Figure~\ref{fig:rq2_sml_ecs}.

\subsubsection{Task performance under perturbations aggregated by relative information density (RQ2)}
\begin{figure}[t]
  \centering

  % Group (a): all perturbations (3 stacked)

    \includegraphics[width=\columnwidth]{assets/plots_no_hellaswag/RQ2/rid/Qwen3-0.6B-Base_rid.pdf}

    \includegraphics[width=\columnwidth, clip, trim={0 0 0 35}]{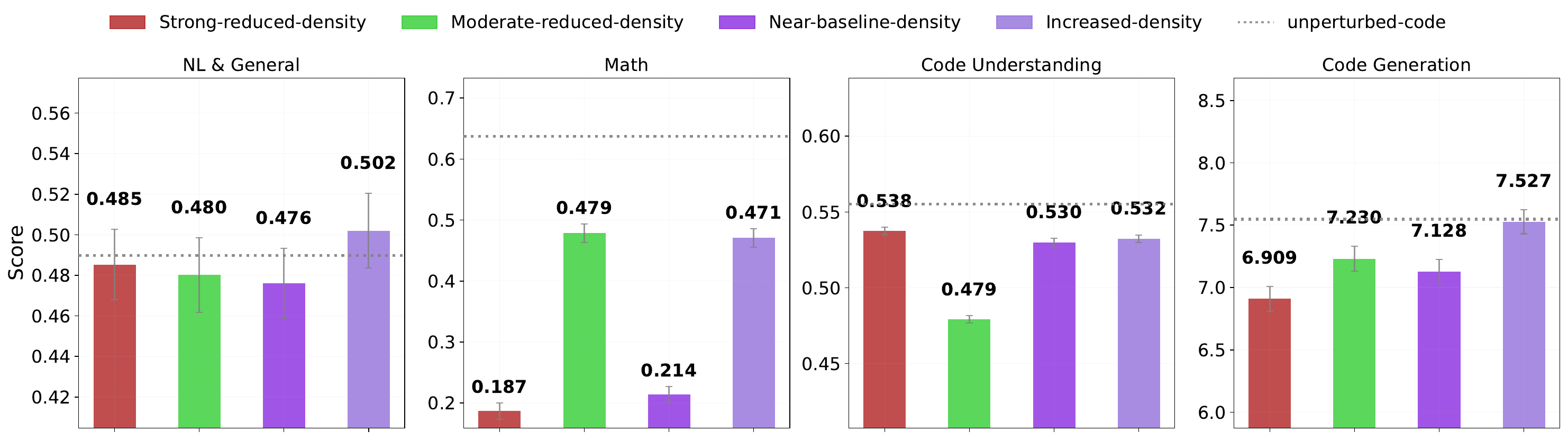}

    \includegraphics[width=\columnwidth, clip, trim={0 0 0 35}]{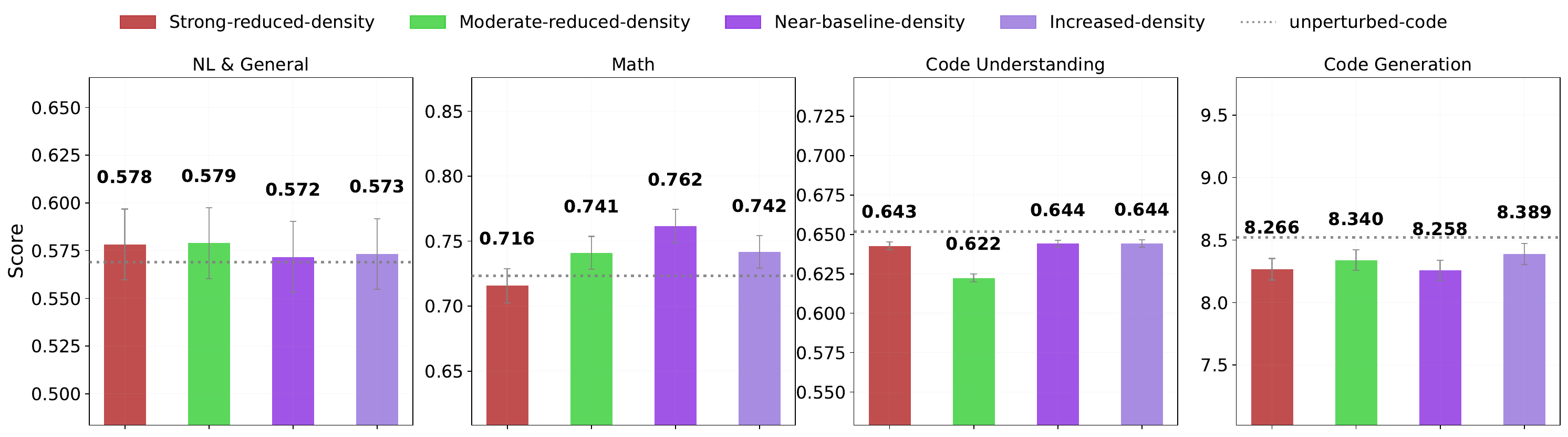}

    \caption{Task performance under perturbations aggregated by relative information density across Qwen3-Base models (0.6B (top), 1.7B (mid), 8B (bottom)).}
  \label{fig:rq2_qwen_rid}
\end{figure}

\paragraph{Qwen3 model family results (relative information density perturbations)}
See performance of aggregated task performance under relative information density perturbations in Figure~\ref{fig:rq2_qwen_rid}.

\begin{figure}[t]
  \centering

    \includegraphics[width=\columnwidth]{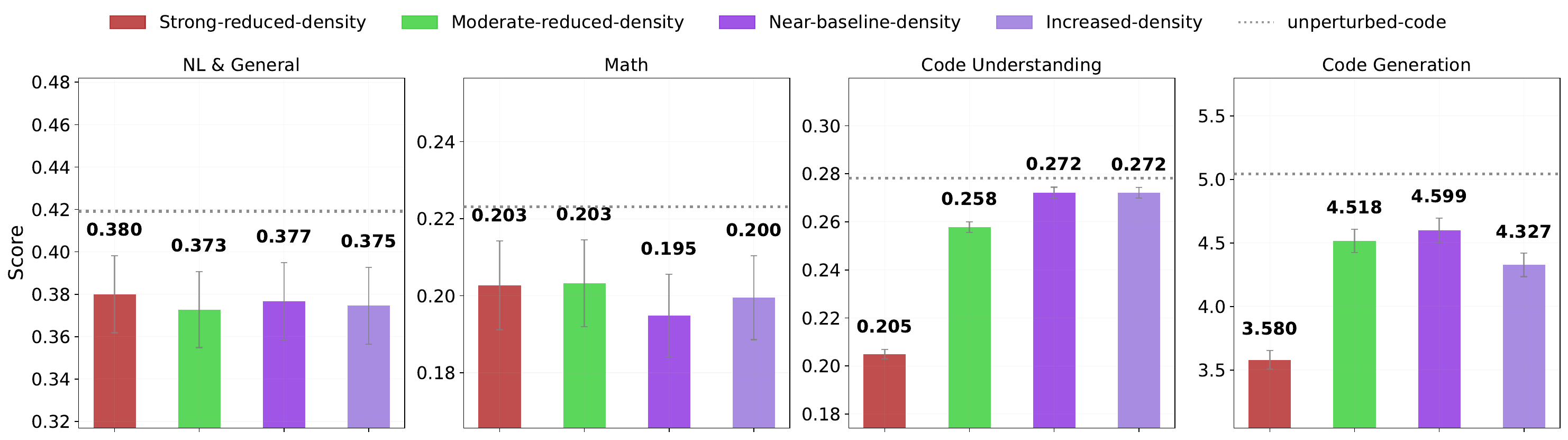}

    \includegraphics[width=\columnwidth, clip, trim={0 0 0 35}]{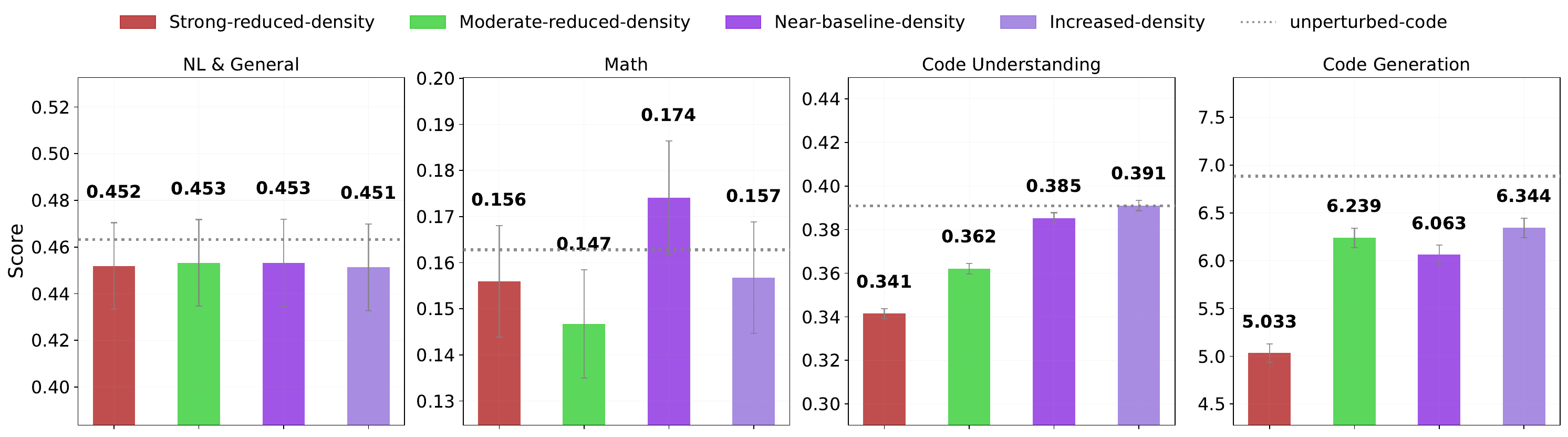}

    \caption{Task performance under perturbations aggregated by relative information density across Llama-3.2 models (1B (top), 3B (bottom)).}
  \label{fig:rq2_llama_rid}
\end{figure}

\paragraph{Llama-3.2 model family results (relative information density perturbations)}
See performance of aggregated task performance under relative information density perturbations in Figure~\ref{fig:rq2_llama_rid}.

\begin{figure}[t]

    \includegraphics[width=\columnwidth]{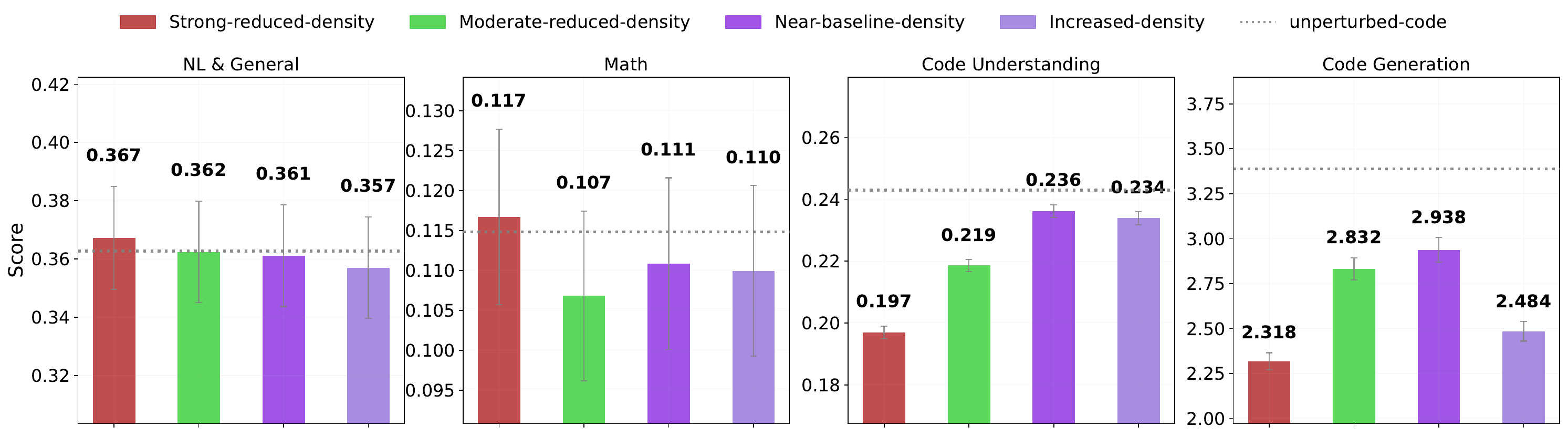}

    \includegraphics[width=\columnwidth, clip, trim={0 0 0 35}]{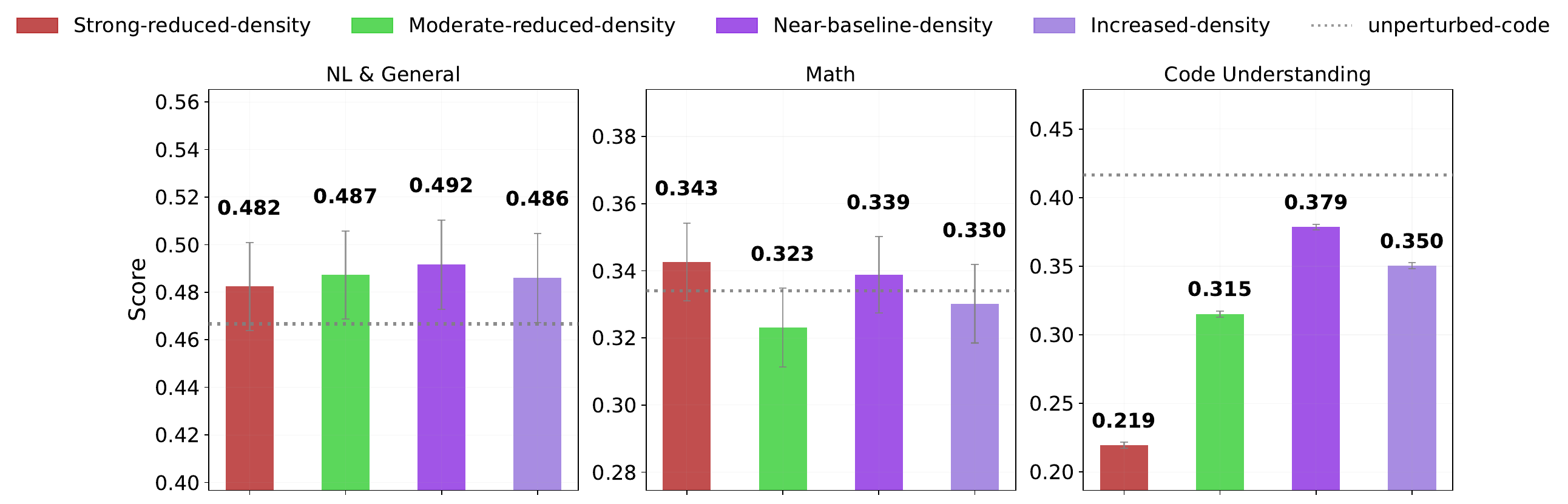}

    \caption{Task performance under perturbations aggregated by relative information density across Gemma-3 models (1B (top), 4B (bottom)).}

  \label{fig:rq2_gemma_rid}
\end{figure}

\paragraph{Gemma-3 model family results (relative information density perturbations)}
See performance of aggregated task performance under relative information density perturbations in Figure~\ref{fig:rq2_gemma_rid}.

\begin{figure}[t]
  \centering
  
  \includegraphics[width=0.7\columnwidth]{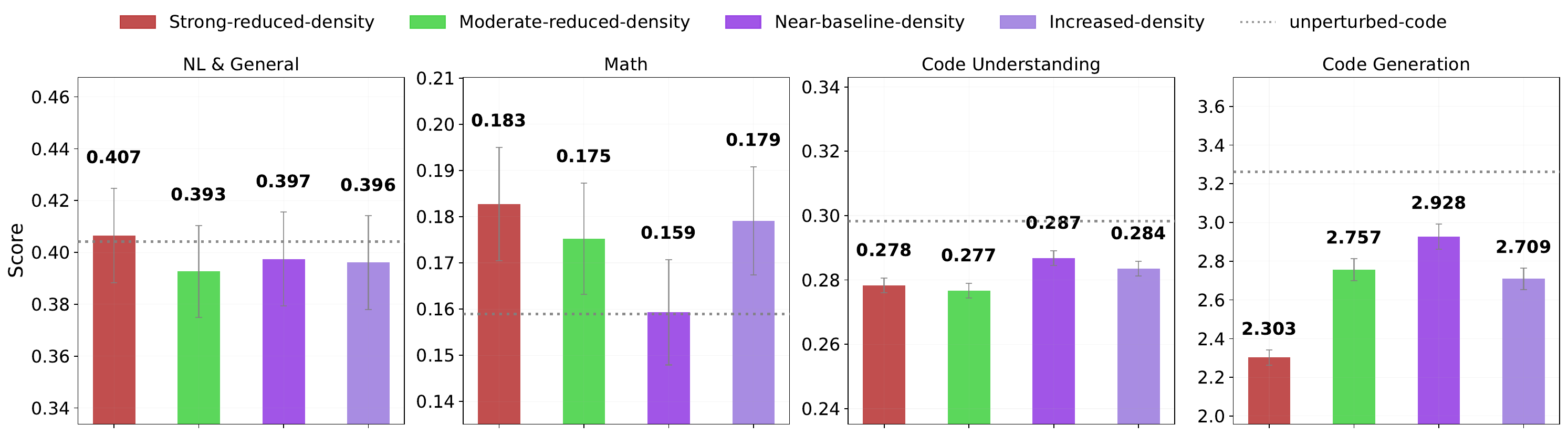}
  \caption{Additional performance of OLMo-2-0425-1B aggregated by relative information density across tasks.}
  \label{fig:rq2_olmo_rid}
\end{figure}

\paragraph{OlMo-2 model family results (relative information density perturbations)}
See performance of aggregated task performance under relative information density perturbations in Figure~\ref{fig:rq2_olmo_rid}.

\begin{figure}[t]
  \centering
    \includegraphics[width=\columnwidth]{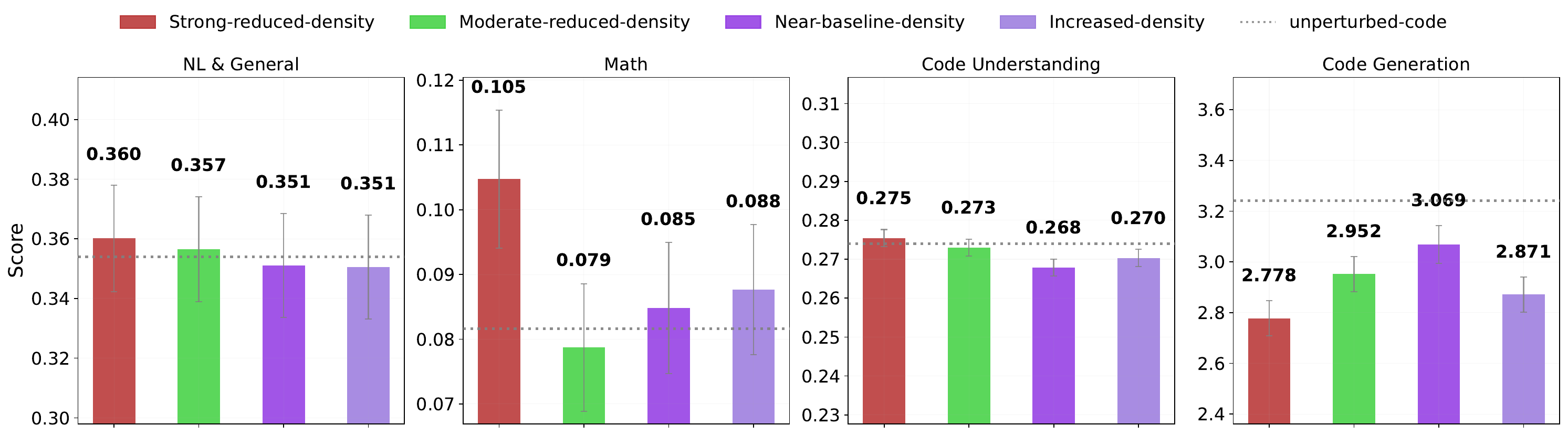}

    \includegraphics[width=\columnwidth, clip, trim={0 0 0 35}]{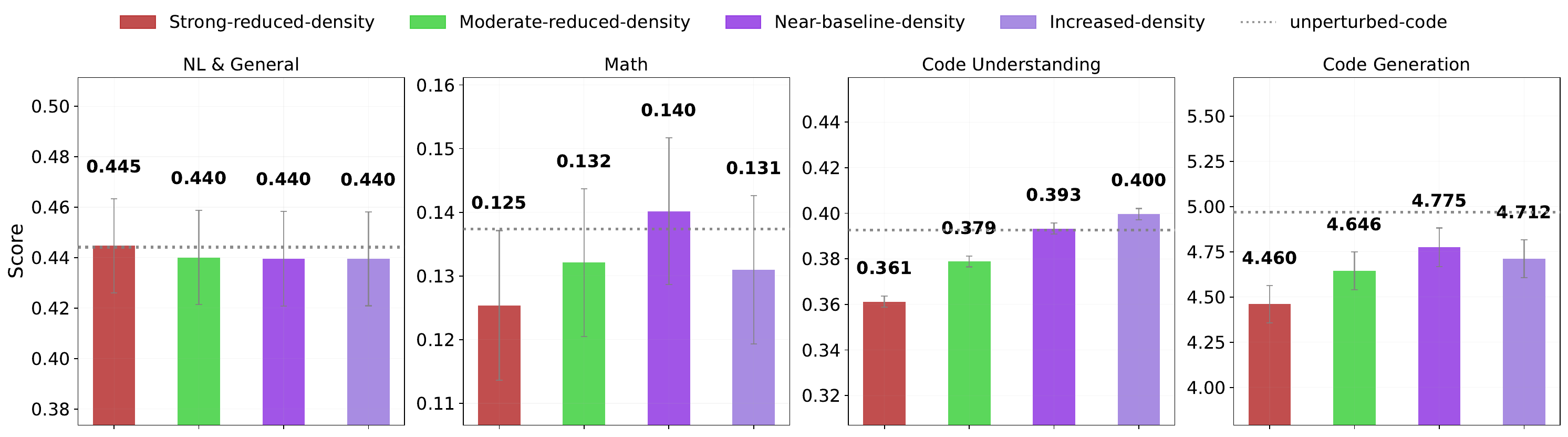}

  \caption{Task performance under perturbations aggregated by relative information density across SmolLM2 models (360M (top), 1.7B (bottom)).}
  \label{fig:rq2_sml_rid}
\end{figure}

\paragraph{SmolLM2 model family results (relative information density perturbations)}
See performance of aggregated task performance under relative information density perturbations in Figure~\ref{fig:rq2_sml_rid}.

\subsubsection{Task performance under perturbations aggregated by human interpretability (RQ2)}
\begin{figure}[t]
  \centering

  % Group (a): all perturbations (3 stacked)

    \includegraphics[width=\columnwidth]{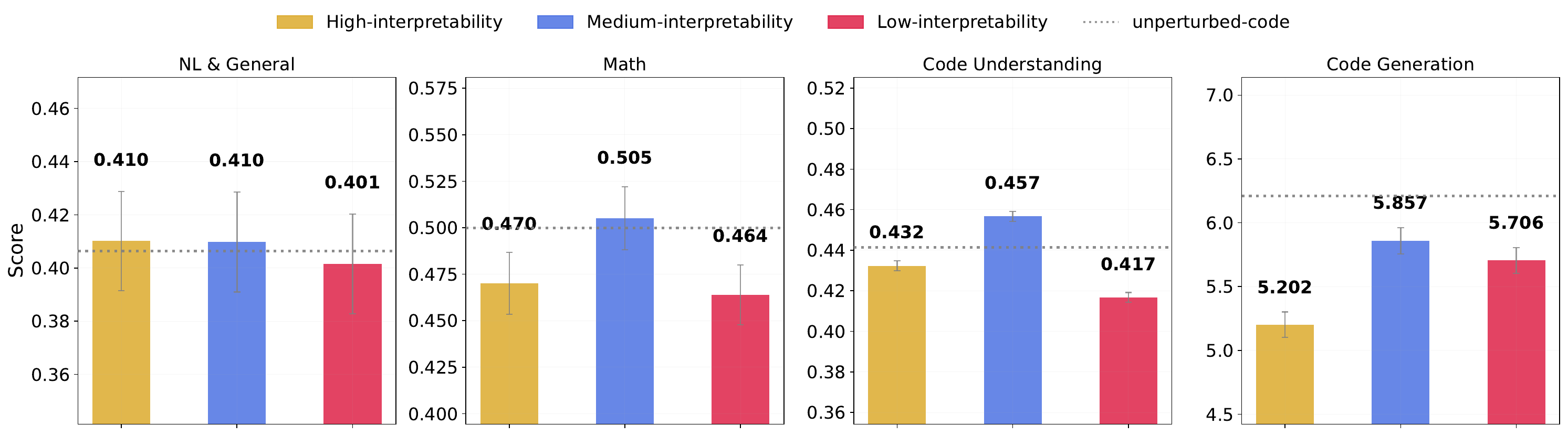}

    \includegraphics[width=\columnwidth]{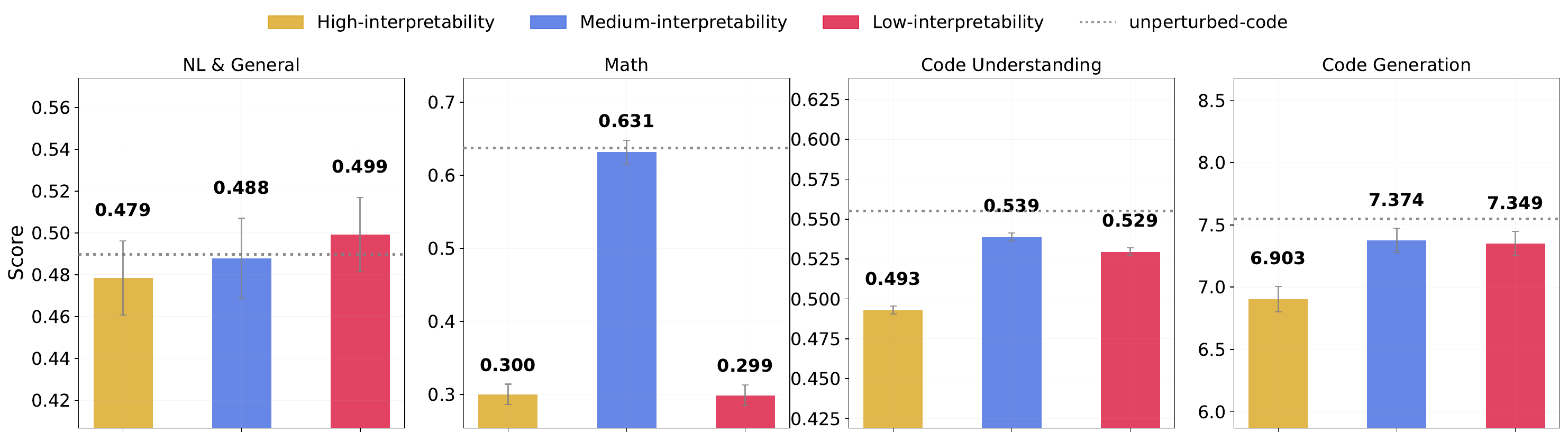}

    \includegraphics[width=\columnwidth, clip, trim={0 0 0 35}]{assets/plots_no_hellaswag/RQ2/hi/Qwen3-8B-Base_hi.pdf}

    \caption{Task performance under perturbations aggregated by human interpretability across Qwen3-Base models (0.6B (top), 1.7B (mid), 8B (bottom)).}
  \label{fig:rq2_qwen_hi}
\end{figure}

\paragraph{Qwen3 model family results (human interpretability perturbations)}
See performance of aggregated task performance under human interpretability perturbations in Figure~\ref{fig:rq2_qwen_hi}.

\begin{figure}[t]
  \centering

    \includegraphics[width=\columnwidth]{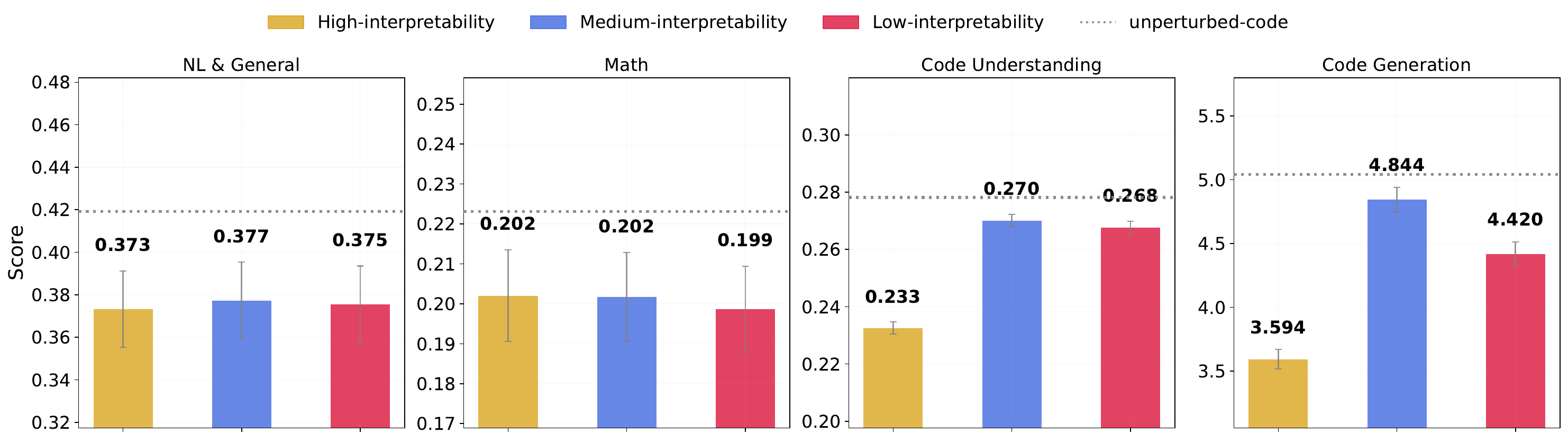}

    \includegraphics[width=\columnwidth, clip, trim={0 0 0 35}]{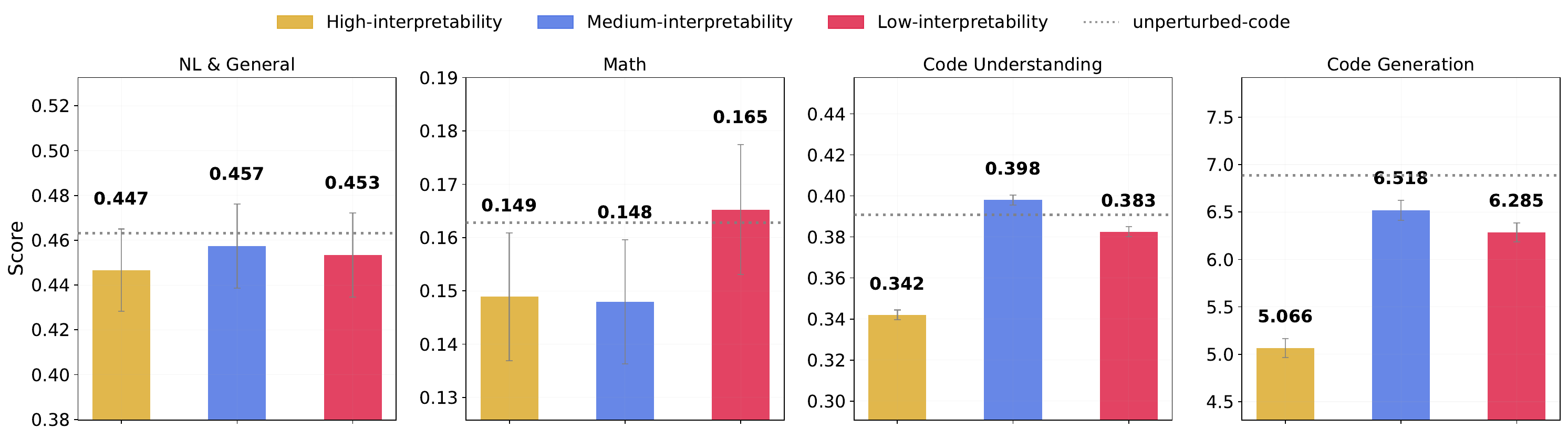}

    \caption{Task performance under perturbations aggregated by human interpretability across Llama-3.2 models (1B (top), 3B (bottom)).}
  \label{fig:rq2_llama_hi}
\end{figure}

\paragraph{Llama-3.2 model family results (human interpretability perturbations)}
See performance of aggregated task performance under human interpretability perturbations in Figure~\ref{fig:rq2_llama_hi}.

\begin{figure}[t]

    \includegraphics[width=\columnwidth]{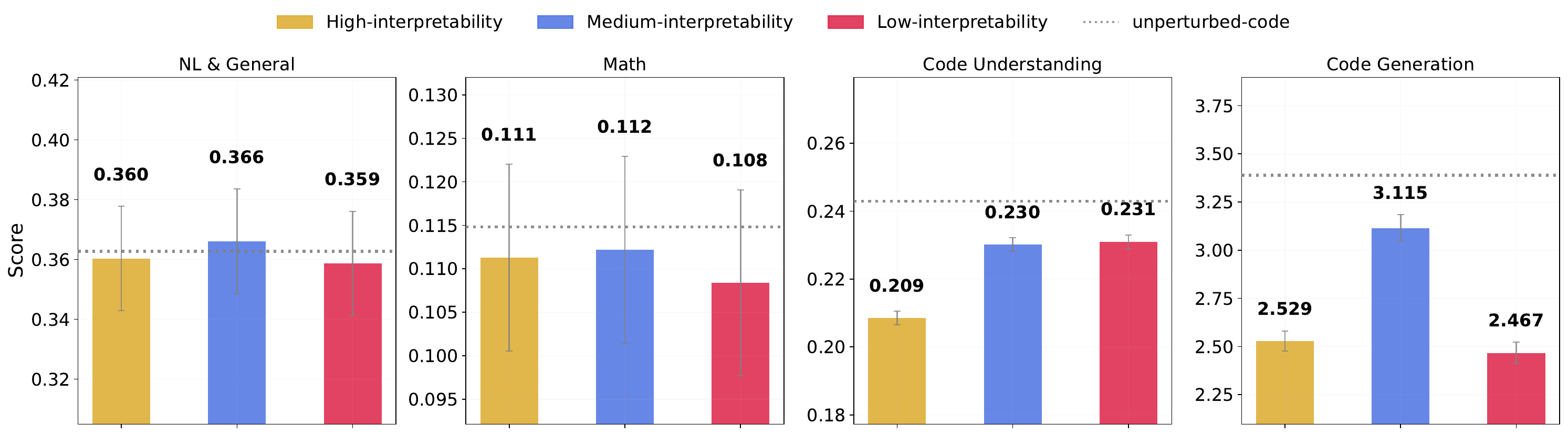}

    \includegraphics[width=\columnwidth, clip, trim={0 0 0 35}]{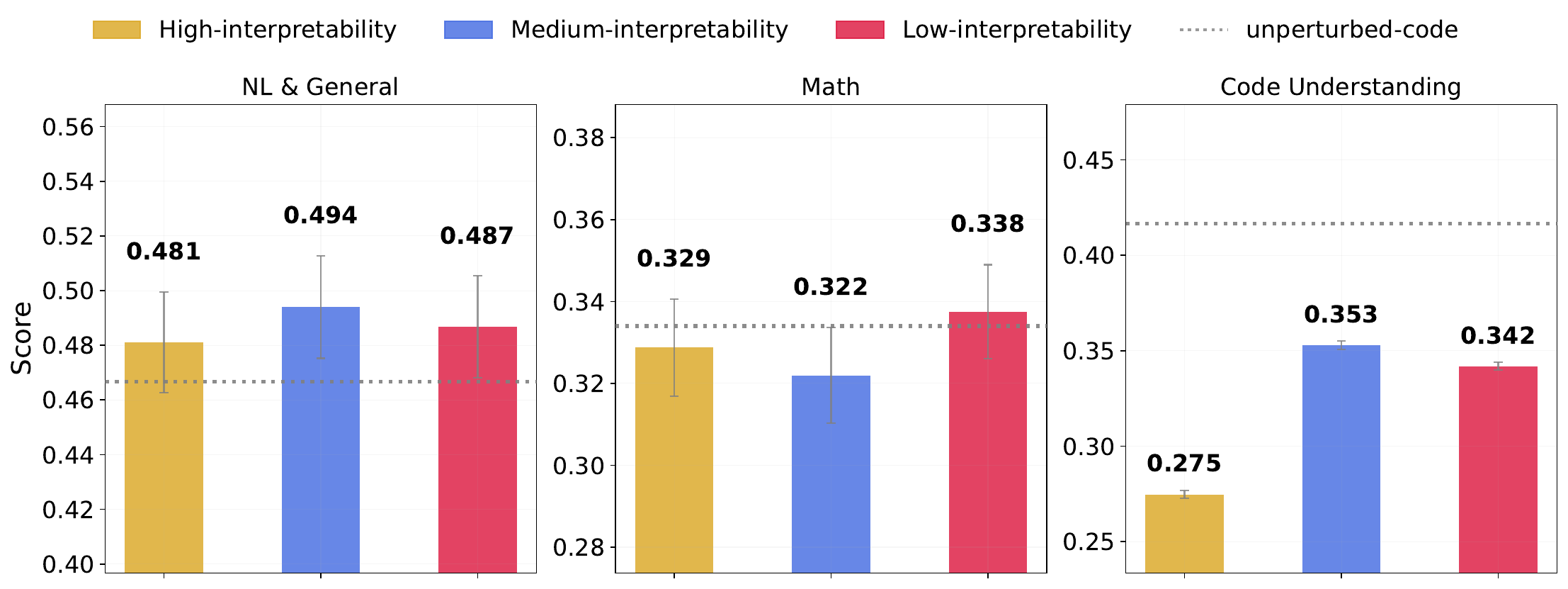}

    \caption{Task performance under perturbations aggregated by human interpretability across Gemma-3 models (1B (top), 4B (bottom)).}

  \label{fig:rq2_gemma_hi}
\end{figure}

\paragraph{Gemma-3 model family results (human interpretability perturbations)}
See performance of aggregated task performance under human interpretability perturbations in Figure~\ref{fig:rq2_gemma_hi}.

\begin{figure}[t]
  \centering
  
  \includegraphics[width=0.7\columnwidth]{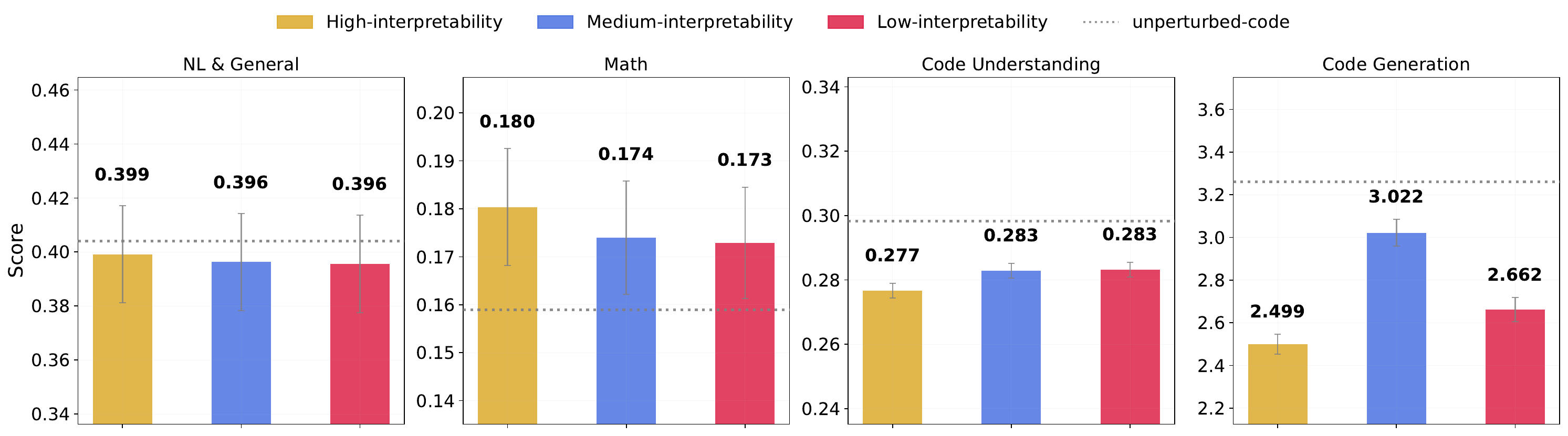}
  \caption{Additional performance of OLMo-2-0425-1B aggregated by human interpretability across tasks.}
  \label{fig:rq2_olmo_hi}
\end{figure}

\paragraph{OlMo-2 model family results (human interpretability perturbations)}
See performance of aggregated task performance under human interpretability perturbations in Figure~\ref{fig:rq2_olmo_hi}.

\begin{figure}[t]
  \centering
    \includegraphics[width=\columnwidth]{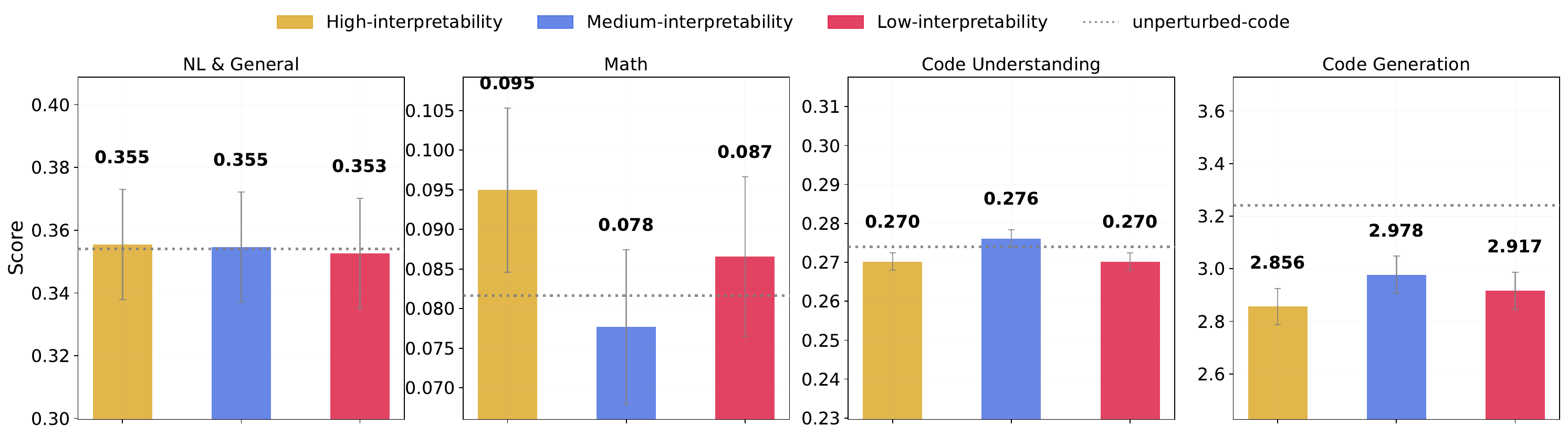}

    \includegraphics[width=\columnwidth, clip, trim={0 0 0 35}]{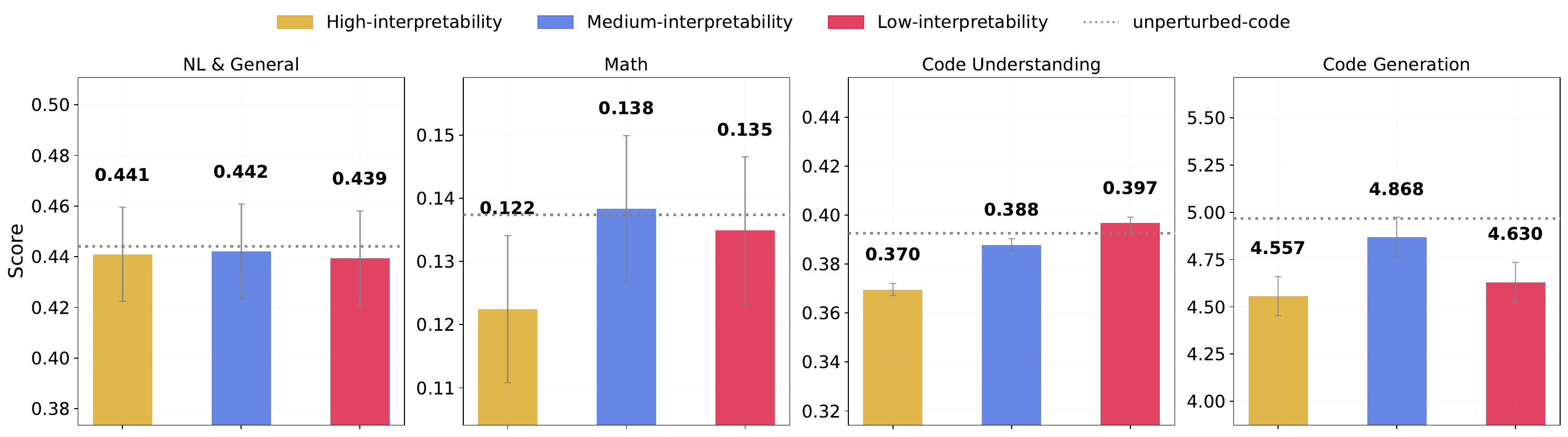}

  \caption{Task performance under perturbations aggregated by human interpretability across SmolLM2 models (360M (top), 1.7B (bottom)).}
  \label{fig:rq2_sml_hi}
\end{figure}

\paragraph{SmolLM2 model family results (human interpretability perturbations)}
See performance of aggregated task performance under human interpretability perturbations in Figure~\ref{fig:rq2_sml_hi}.

\subsubsection{Task performance for all individual perturbations (RQ2)} \label{appx:rq2_all}
\begin{figure}[t]
  \centering

    \includegraphics[width=0.9\columnwidth]{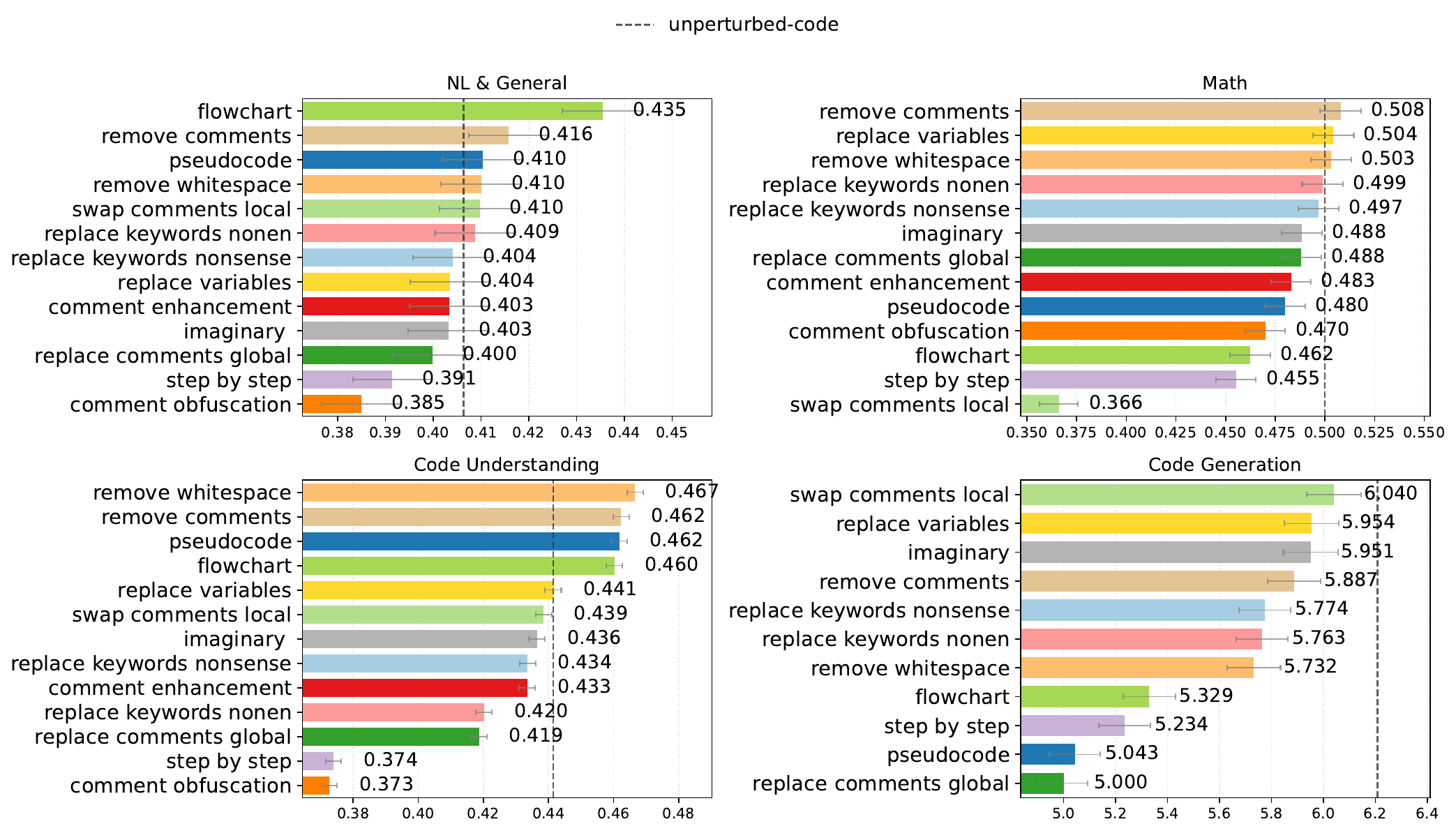}

    \includegraphics[width=0.9\columnwidth]{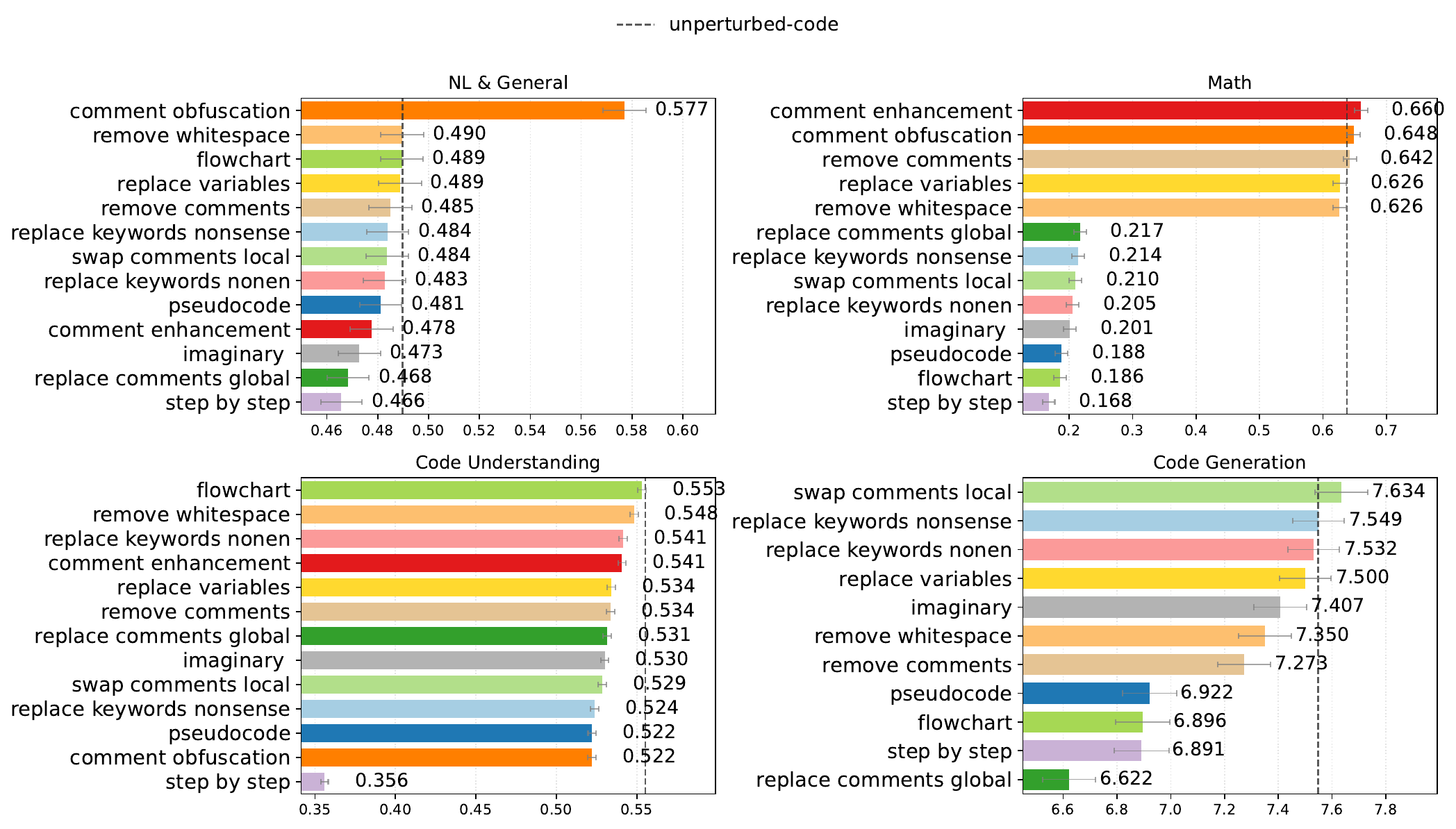}

    \includegraphics[width=0.9\columnwidth]{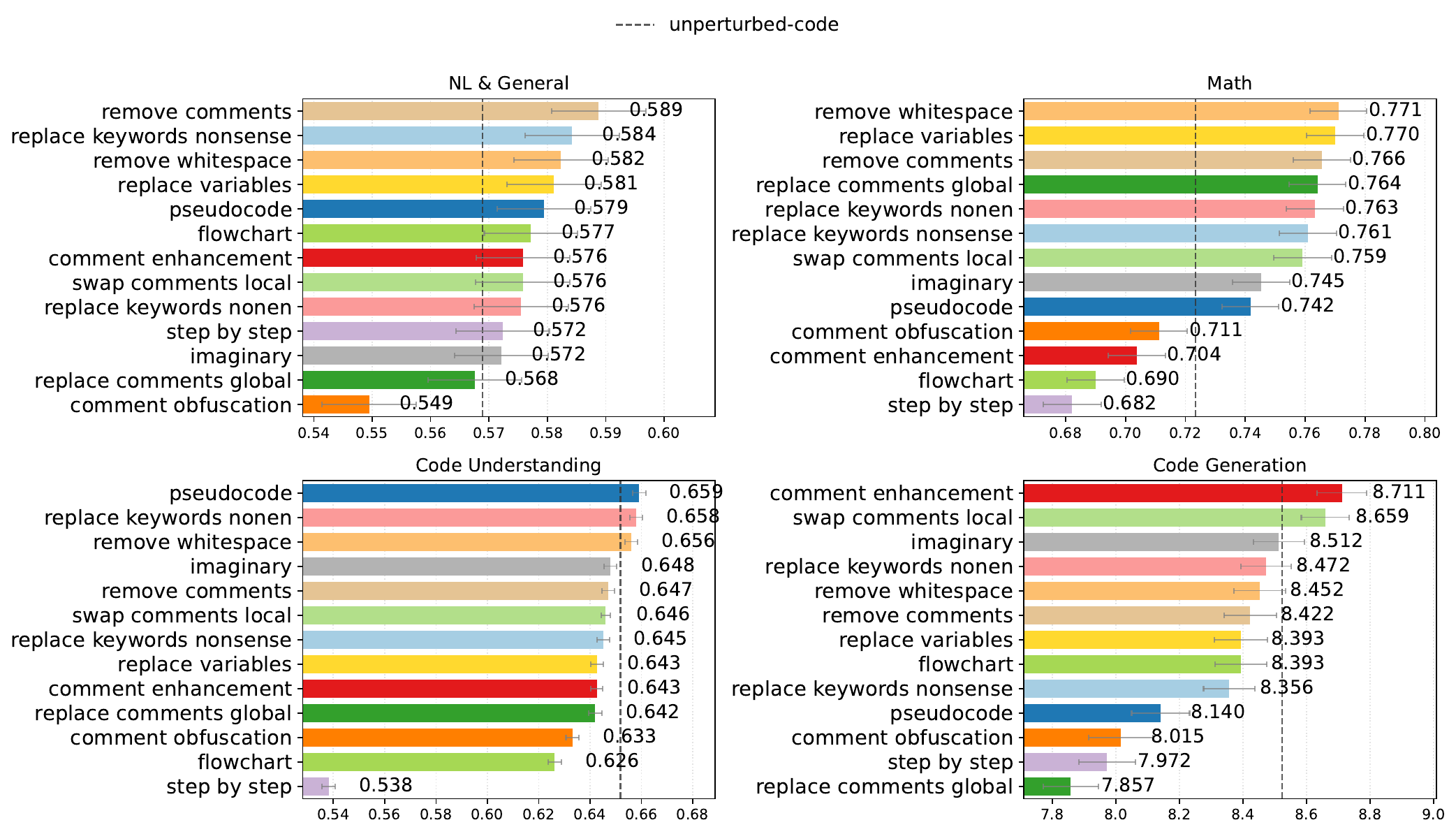}

    \caption{All perturbations across Qwen3-Base models (0.6B (top), 1.7B (mid), 8B (bottom)).}
    \label{fig:qwen_rq2_all}
\end{figure}

\paragraph{Qwen3 model family results (individual perturbations)}
See performance of all perturbation configurations in Figure~\ref{fig:qwen_rq2_all}.

\begin{figure}[t]
  \centering

    \includegraphics[width=\columnwidth]{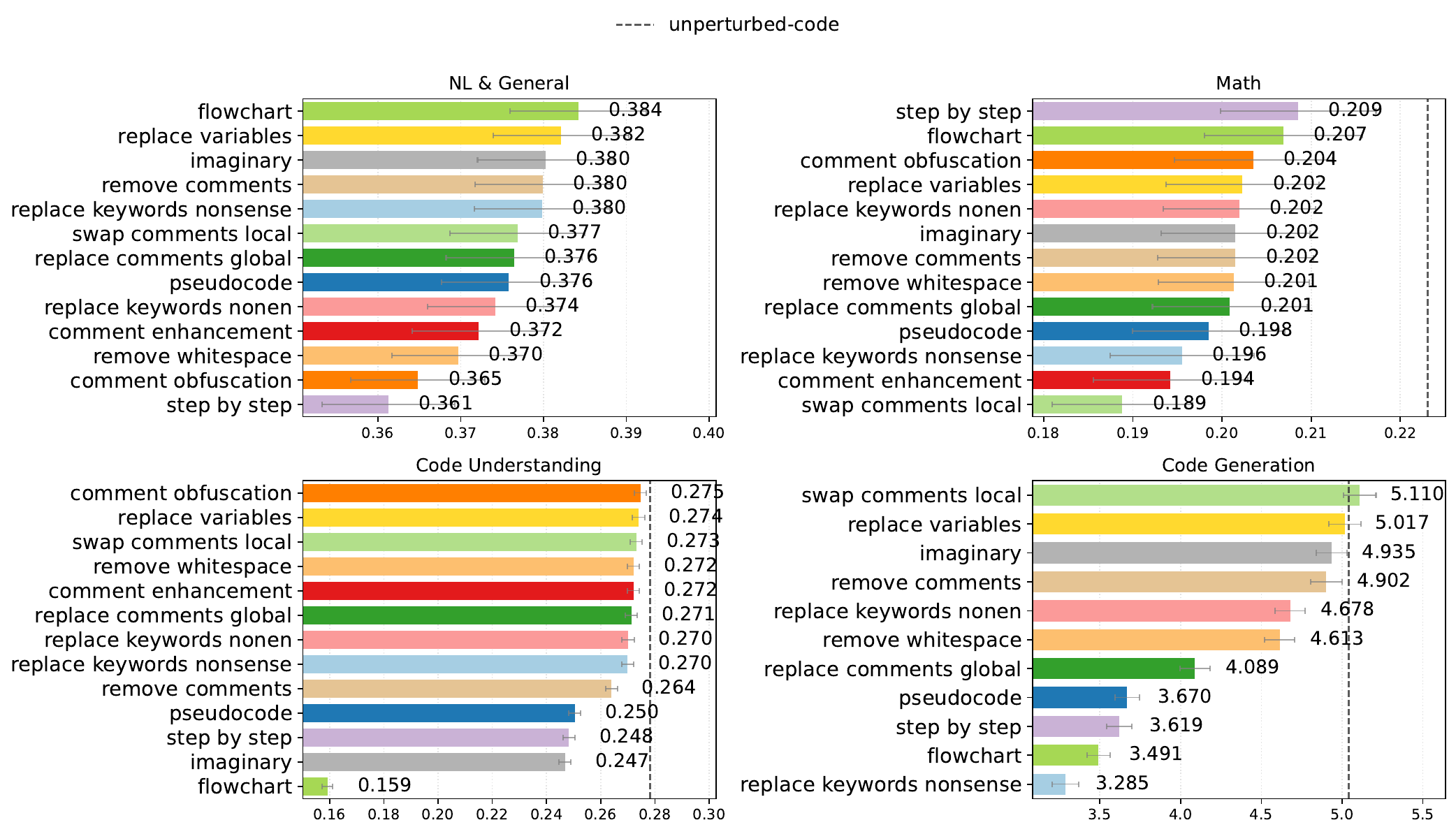}

    \includegraphics[width=\columnwidth]{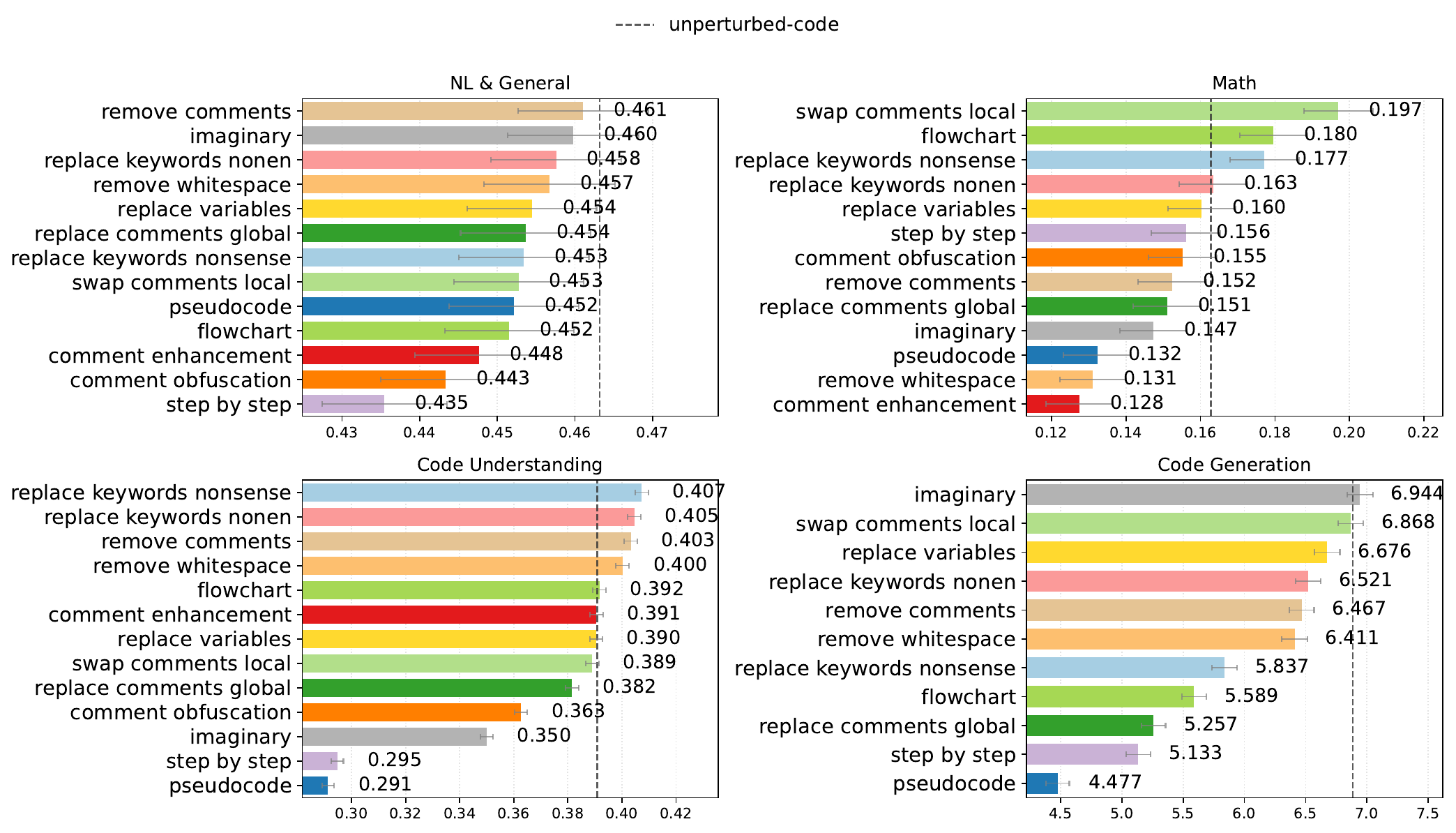}

    \caption{All perturbations across Llama-3.2 models (1B (top), 3B (bottom)).}
    \label{fig:llama_rq2_all}
 
\end{figure}

\paragraph{Llama-3.2 model family results (individual perturbations)}
See performance of all perturbation configurations in Figure~\ref{fig:llama_rq2_all}.

\begin{figure}[t]
  \centering

    \includegraphics[width=\columnwidth]{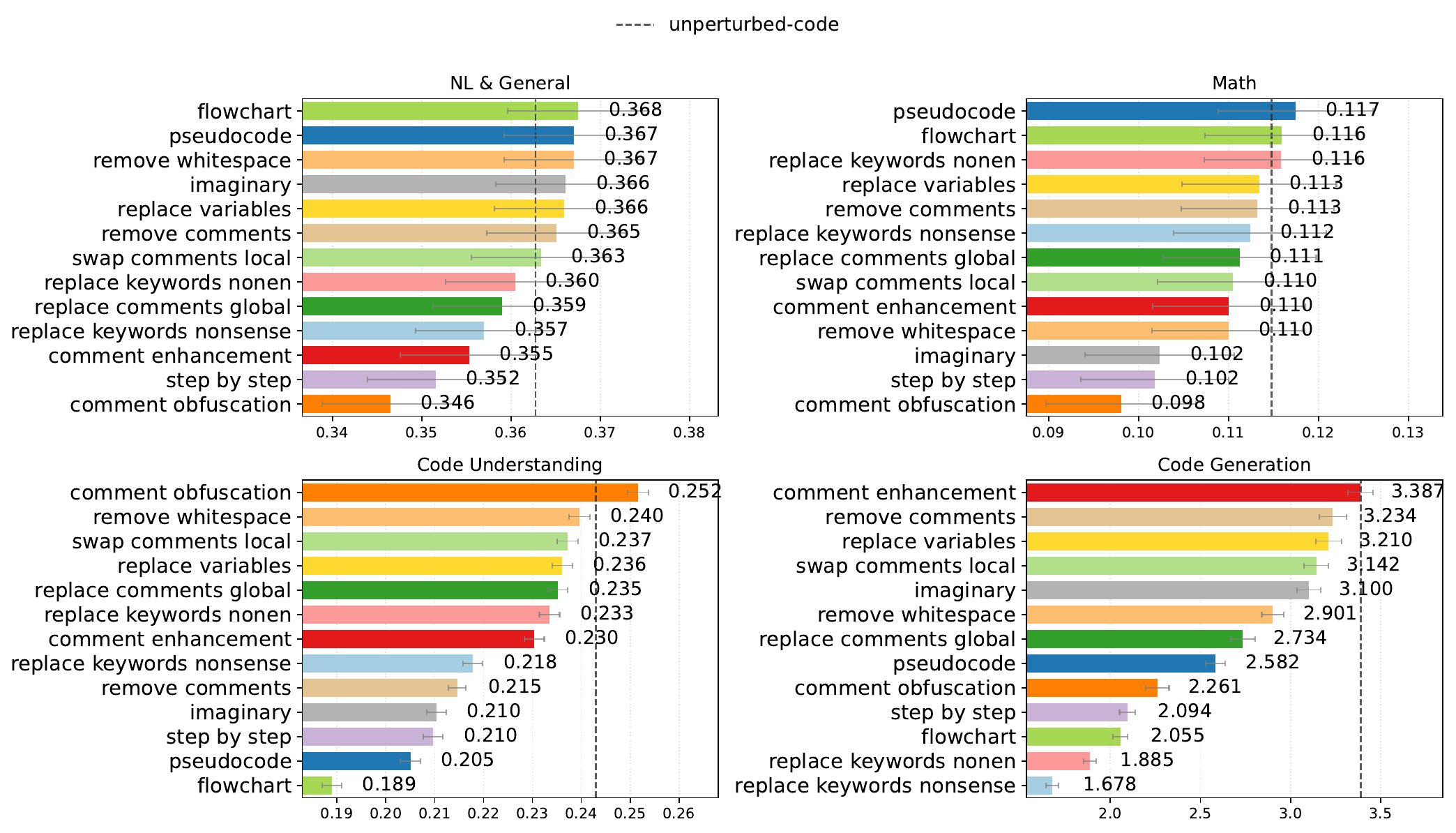}

    \includegraphics[width=\columnwidth]{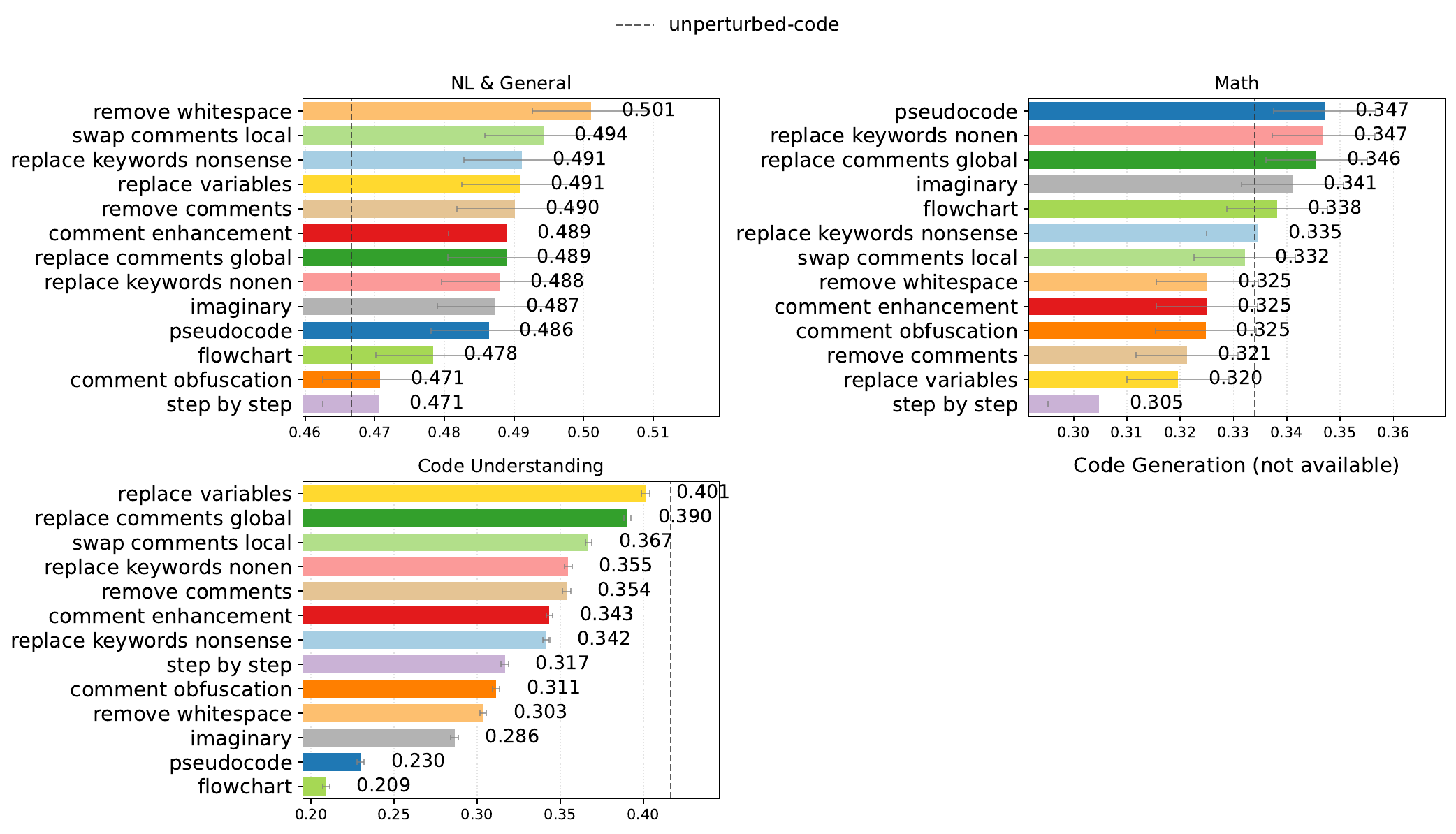}

    \caption{All perturbations across Gemma-3 models (1B (top), 4B (bottom)).}
    \label{fig:gemma_rq2_all}
\end{figure}

\paragraph{Gemma-3 model family results (individual perturbations)}
See performance of all perturbation configurations in Figure~\ref{fig:gemma_rq2_all}.

\begin{figure}[t]
  \centering
 
    \includegraphics[width=\columnwidth]{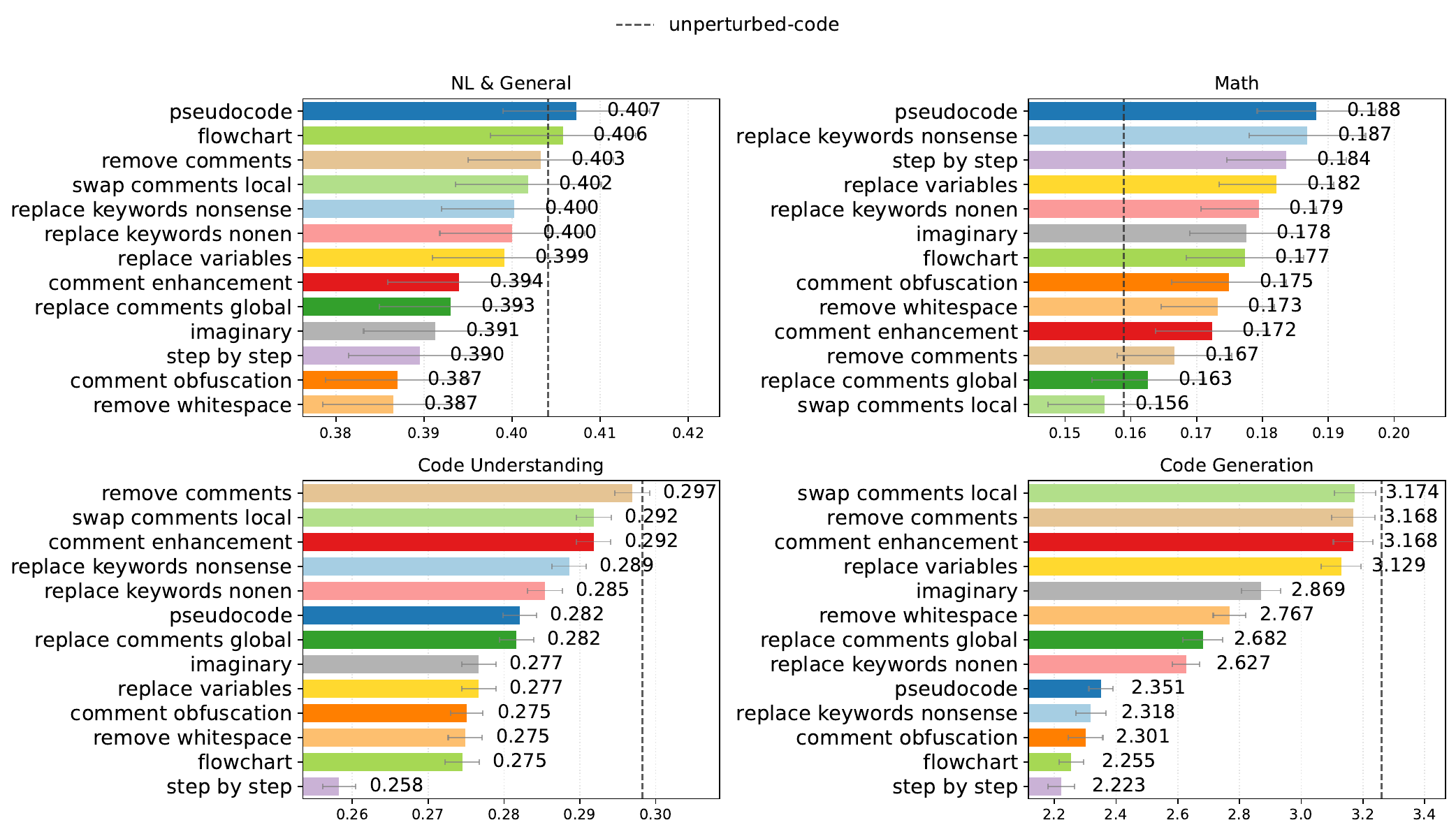}
    \caption{OLMo-2-0425-1B with all perturbations.}
    \label{fig:olmo_rq2_all}

\end{figure}

\paragraph{OlMo-2 model family results (individual perturbations)}
See performance of all perturbation configurations in Figure~\ref{fig:olmo_rq2_all}.

\begin{figure}[t]
  \centering

    \includegraphics[width=\columnwidth]{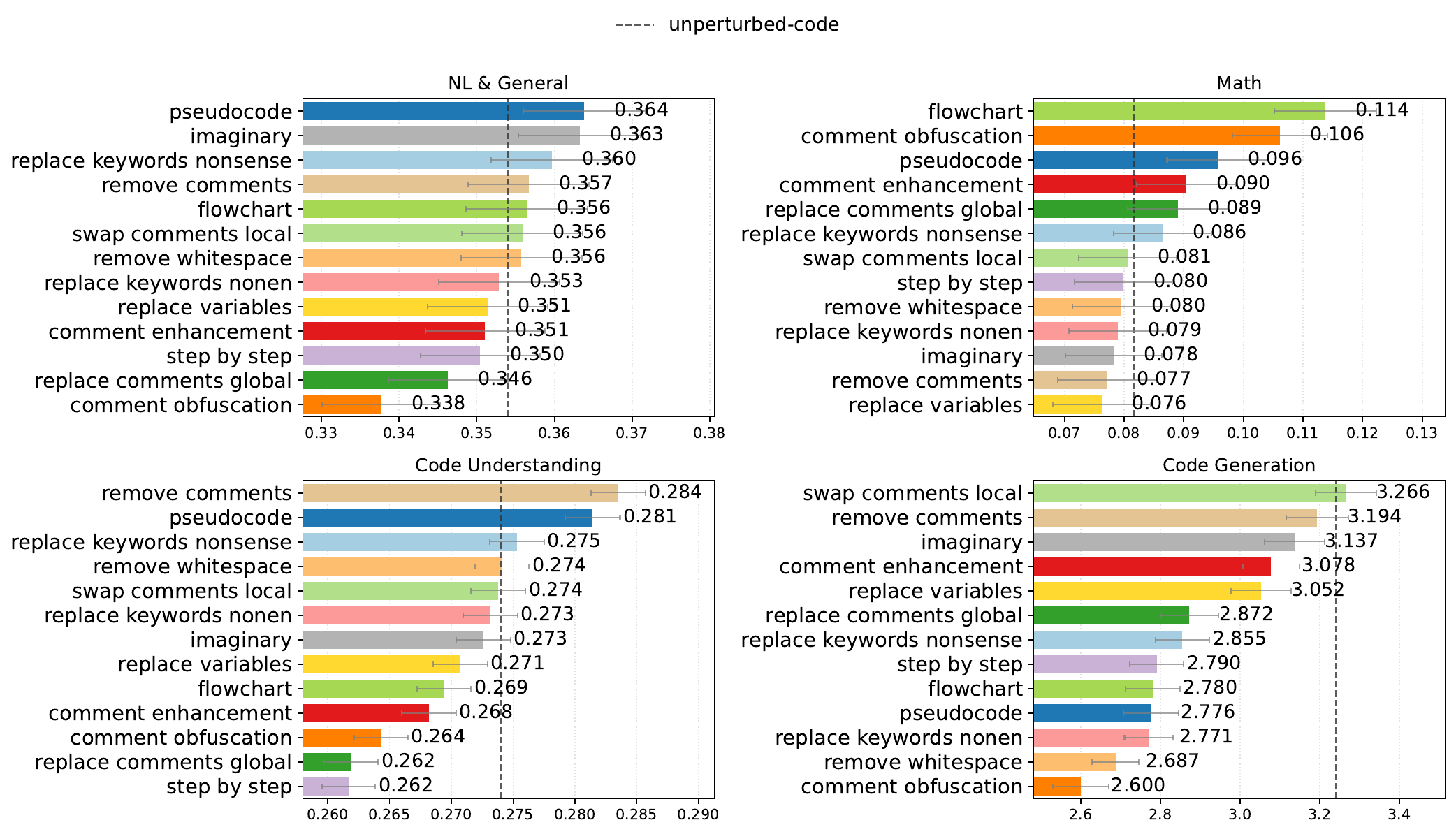}

    \includegraphics[width=\columnwidth]{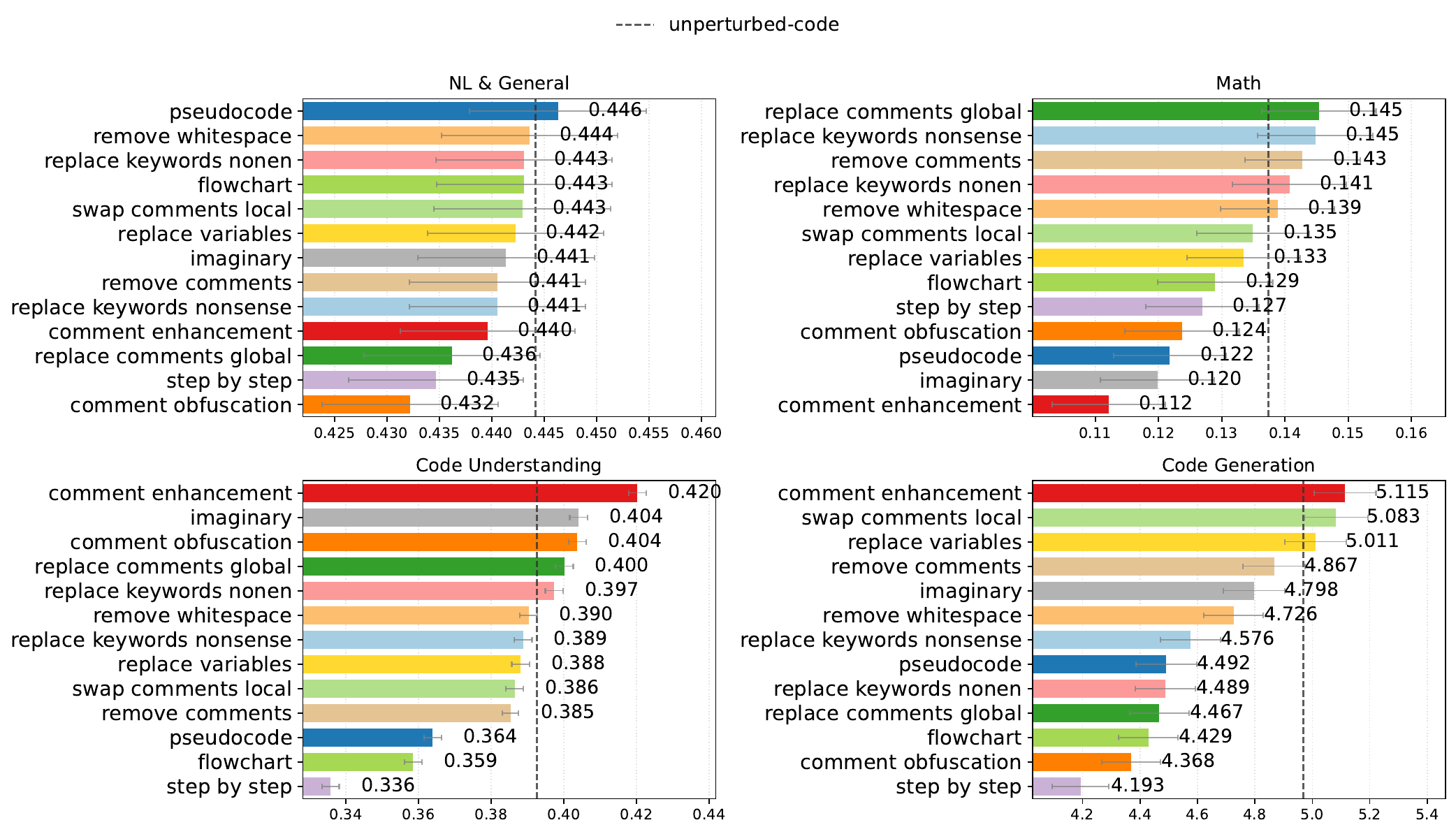}

    \caption{All perturbations across SmolLM2 models (360M (top), 1.7B (bottom)).}
    \label{fig:sml_rq2_all}
\end{figure}

\paragraph{SmolLM2 model family results (individual perturbations)}
See performance of all perturbation configurations in Figure~\ref{fig:sml_rq2_all}.

\subsubsection{Task performance with different programming languages (RQ3)}
\begin{figure}[t]
  \centering

  % Top: groups
  \begin{subfigure}{\columnwidth}
    \centering
    \includegraphics[width=\columnwidth]{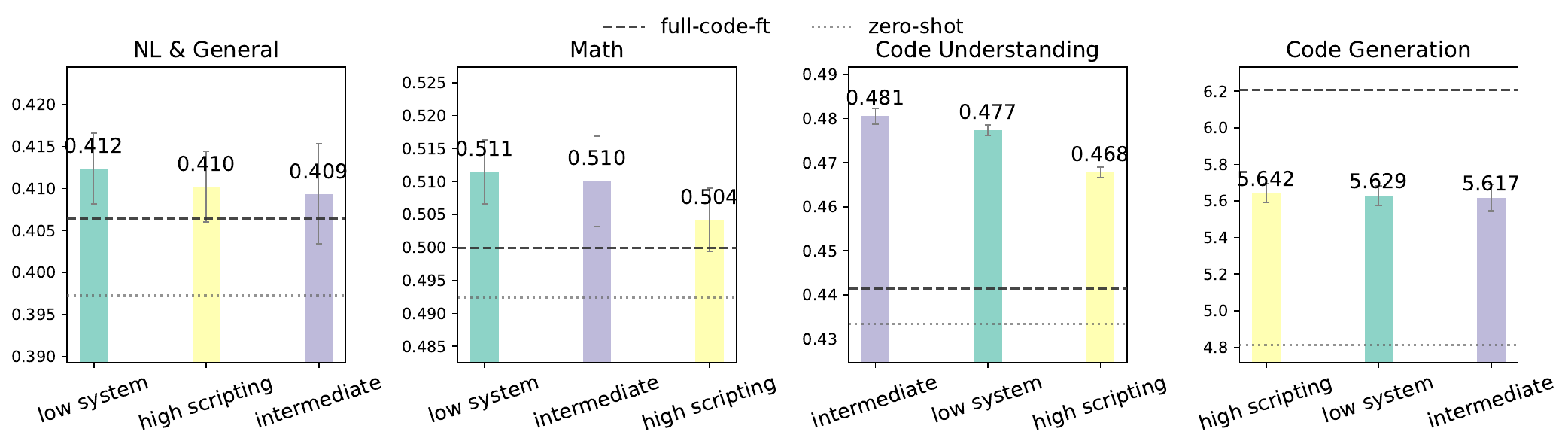}
    \caption{Qwen3-0.6B-Base}
    \label{fig:qwen3-0.6bb_rq3_groups}
  \end{subfigure}

  \begin{subfigure}{\columnwidth}
    \centering
    \includegraphics[width=\columnwidth,clip, trim={0 0 0 40}]{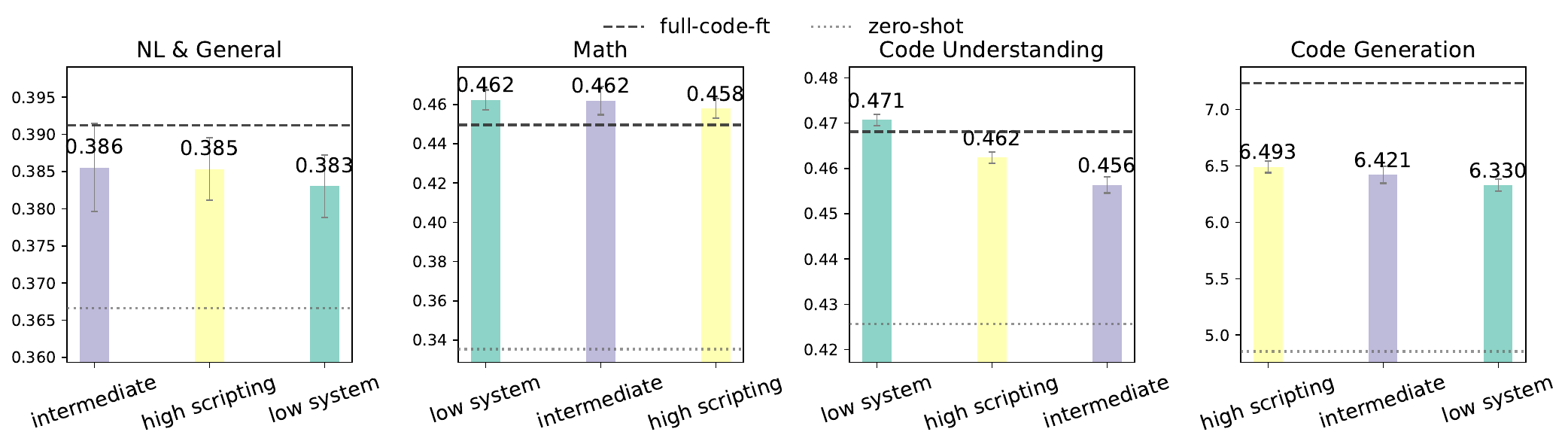}
    \caption{Qwen3-0.6B}
    \label{fig:qwen3-0.6b_rq3_groups}
  \end{subfigure}

  % Bottom: langs
  \begin{subfigure}{\columnwidth}
    \centering
    \includegraphics[width=\columnwidth,clip, trim={0 0 0 40}]{assets/plots_no_hellaswag/RQ3/Qwen3-1.7B_rq3_groups.pdf}
    \caption{Qwen3-1.7B}
    \label{fig:qwen3-1.7b_rq3_groups}
  \end{subfigure}

  \caption{Grouped performance of Qwen-3 family under low-system, intermediate, and high-scripting programming languages.}
  \label{fig:qwen_rq3_groups_combined}
\end{figure}

\begin{figure}[t]
  \centering

  % Top: groups
  \begin{subfigure}{0.9\columnwidth}
    \centering
    \includegraphics[width=0.9\columnwidth]{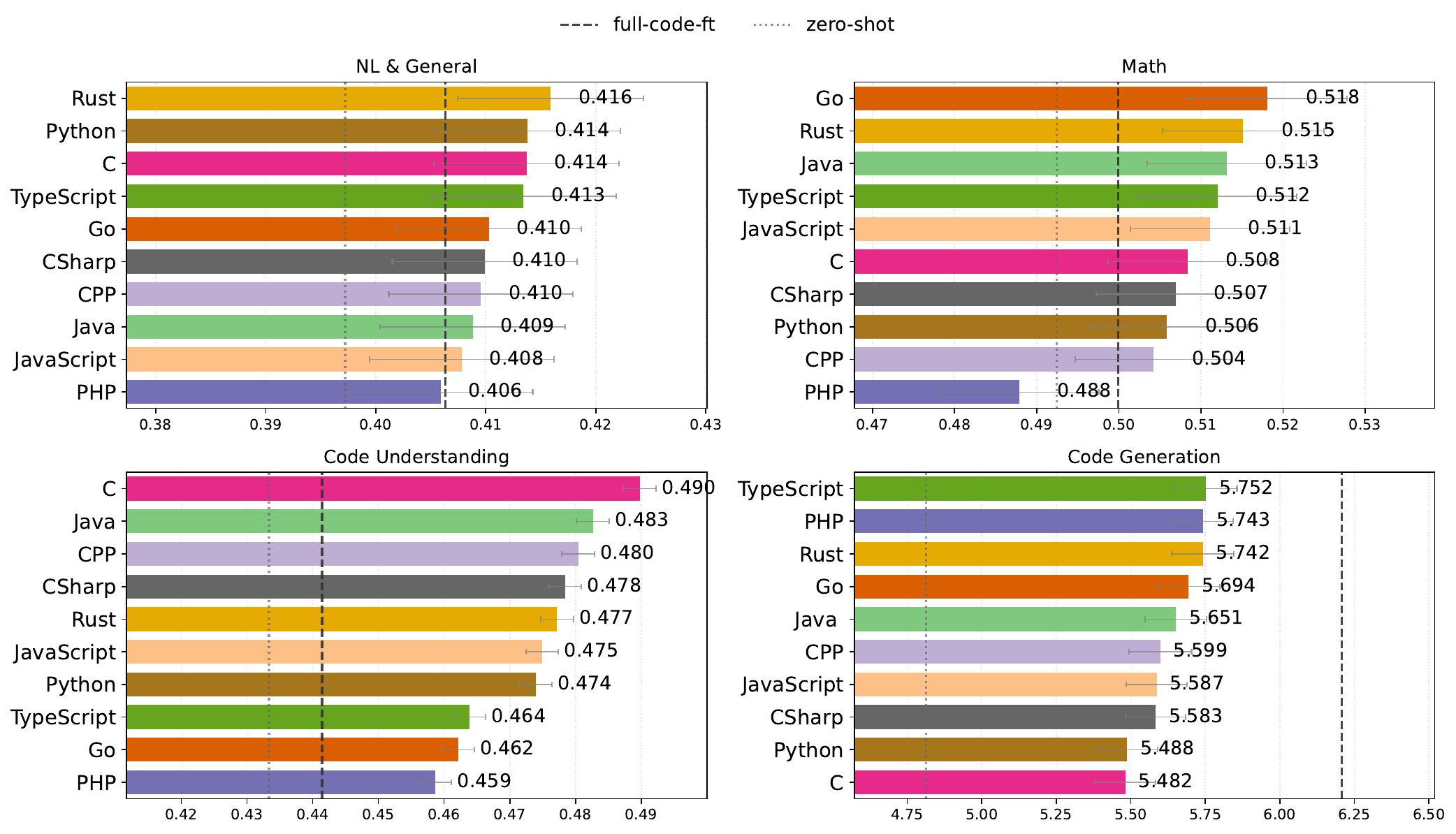}
    \caption{Qwen3-0.6B-Base}
    \label{fig:qwen3-0.6bb_rq3_langs}
  \end{subfigure}

  \begin{subfigure}{\columnwidth}
    \centering
    \includegraphics[width=0.9\columnwidth]{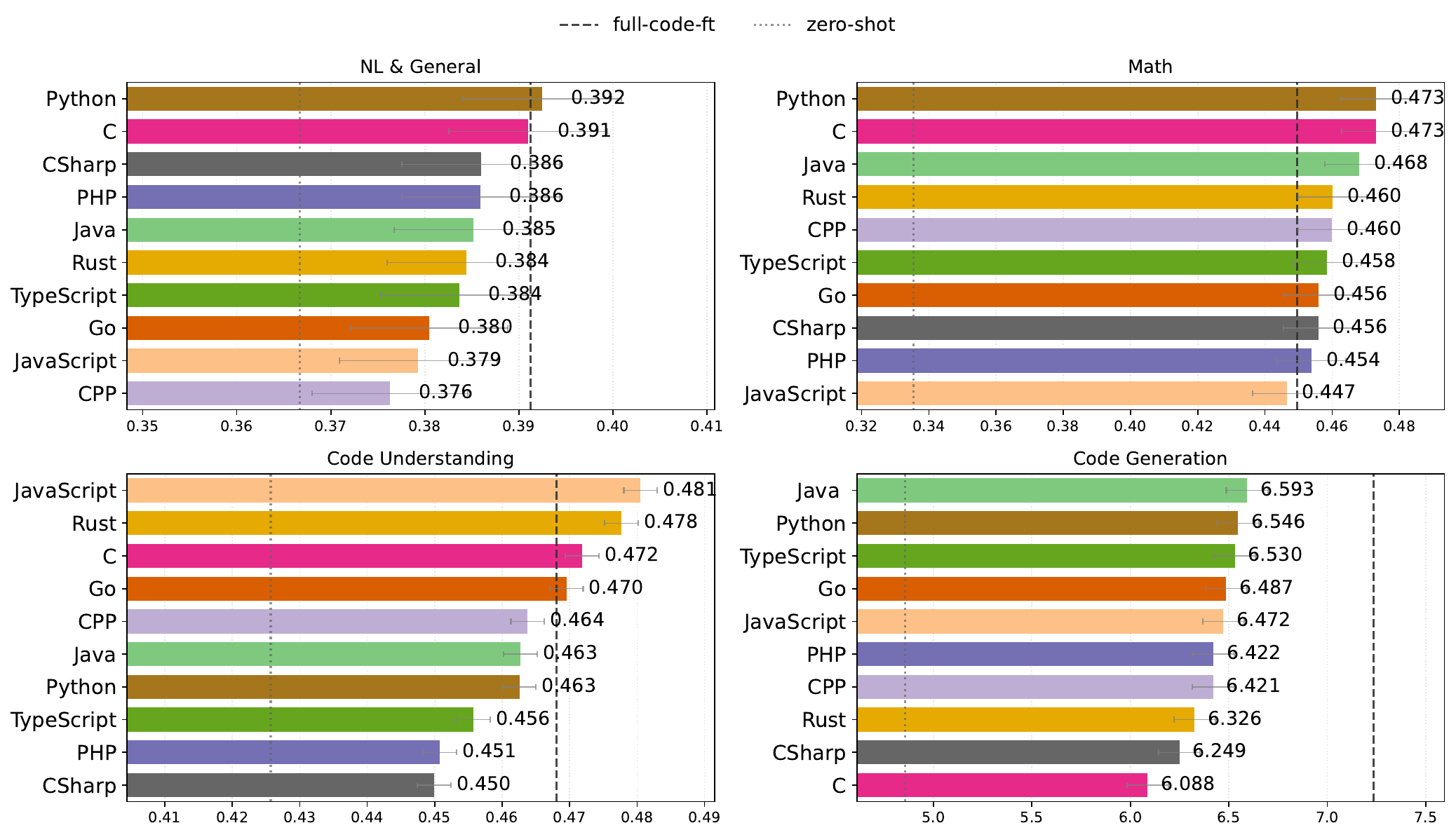}
    \caption{Qwen3-0.6B}
    \label{fig:qwen3-0.6b_rq3_langs}
  \end{subfigure}

  % Bottom: langs
  \begin{subfigure}{\columnwidth}
    \centering
    \includegraphics[width=0.9\columnwidth]{assets/plots_no_hellaswag/RQ3/Qwen3-1.7B_rq3_langs.pdf}
    \caption{Qwen3-1.7B}
    \label{fig:qwen3-1.7b_rq3_langs}
  \end{subfigure}

  \caption{All programming language specific performance of Qwen-3 family.}
  \label{fig:qwen_rq3_langs_combined}
\end{figure}

\paragraph{Qwen3 model family results}
See performance of grouped performance and individual programming languages in Figure~\ref{fig:qwen_rq3_groups_combined} and Figure~\ref{fig:qwen_rq3_langs_combined}, respectively.

\begin{figure}[t]
  \centering

  % Top: groups
  \begin{subfigure}{\columnwidth}
    \centering
    \includegraphics[width=\columnwidth]{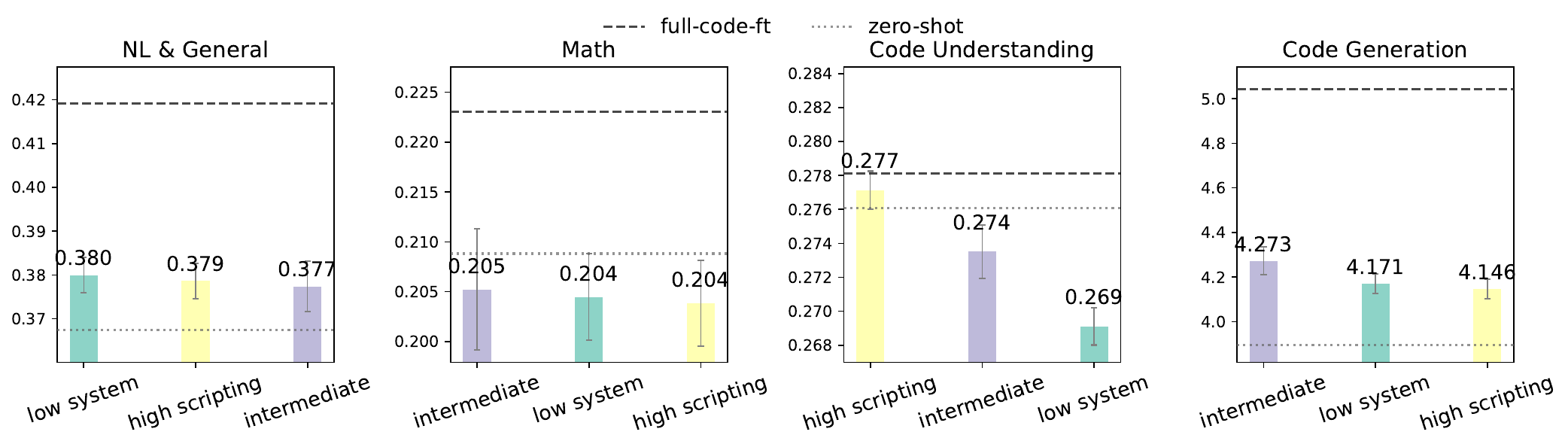}
    \caption{Grouped results (low-system, intermediate, high-scripting)}
    \label{fig:llama_rq3_groups}
  \end{subfigure}

  % Bottom: langs
  \begin{subfigure}{\columnwidth}
    \centering
    \includegraphics[width=\columnwidth]{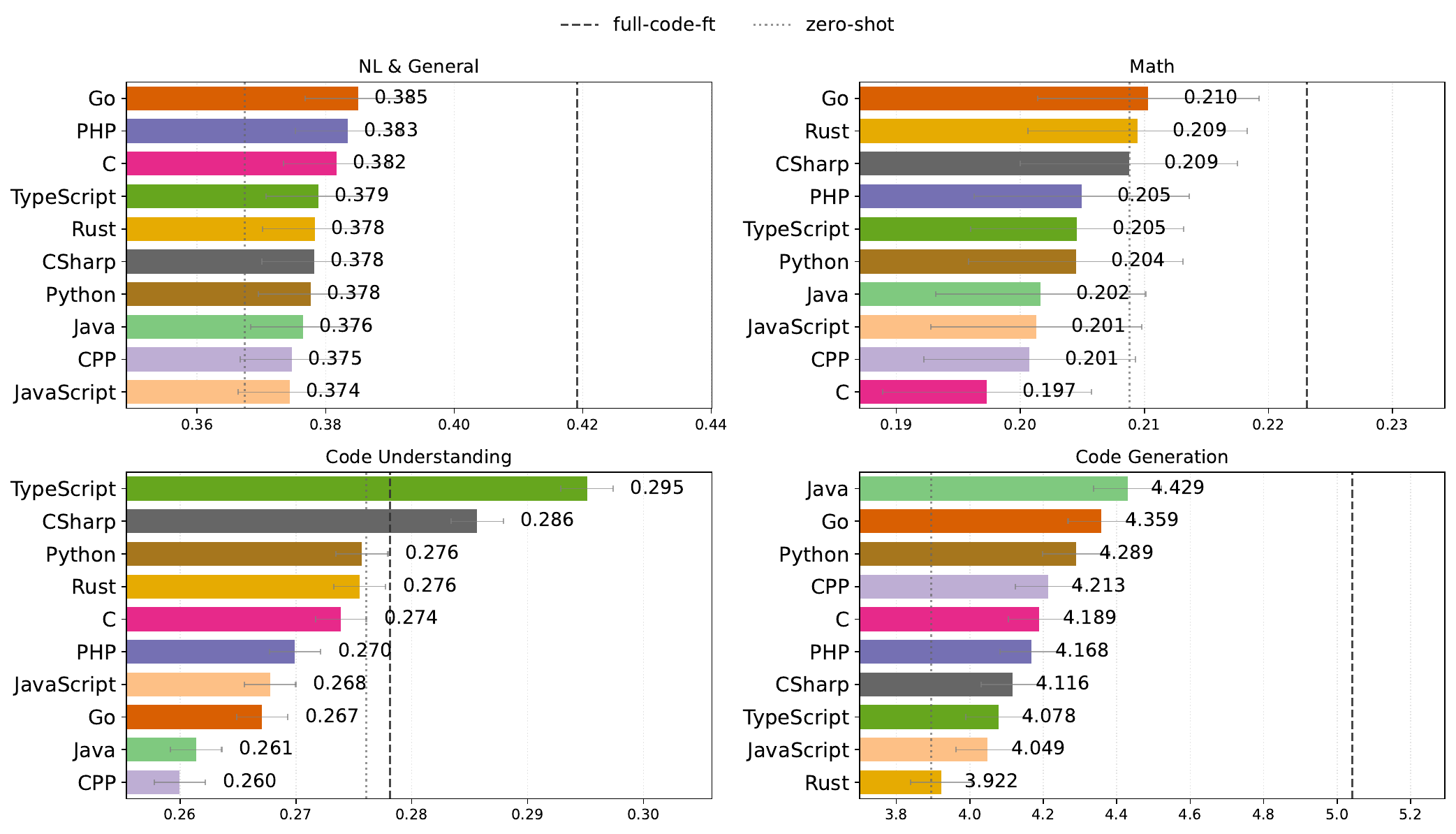}
    \caption{Per-language results}
    \label{fig:llama_rq3_langs}
  \end{subfigure}

  \caption{Performance for Llama-3.2-1B. 
  (a) Programming language groups, (b) individual languages.}
  \label{fig:llama_rq3_combined}
\end{figure}

\paragraph{Llama-3 model family results}
See performance of grouped performance and individual programming languages in Figure~\ref{fig:llama_rq3_combined}.

\begin{figure}[t]
  \centering

  % Top: groups
  \begin{subfigure}{\columnwidth}
    \centering
    \includegraphics[width=\columnwidth]{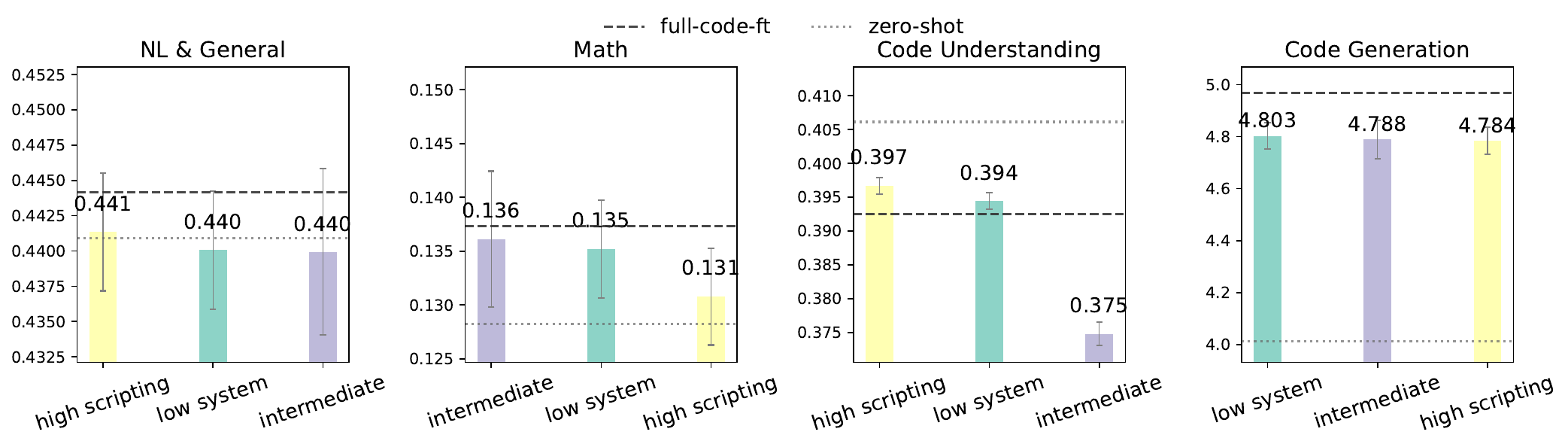}
    \caption{Grouped results (low-system, intermediate, high-scripting)}
    \label{fig:sml_rq3_groups}
  \end{subfigure}

  % Bottom: langs
  \begin{subfigure}{\columnwidth}
    \centering
    \includegraphics[width=\columnwidth]{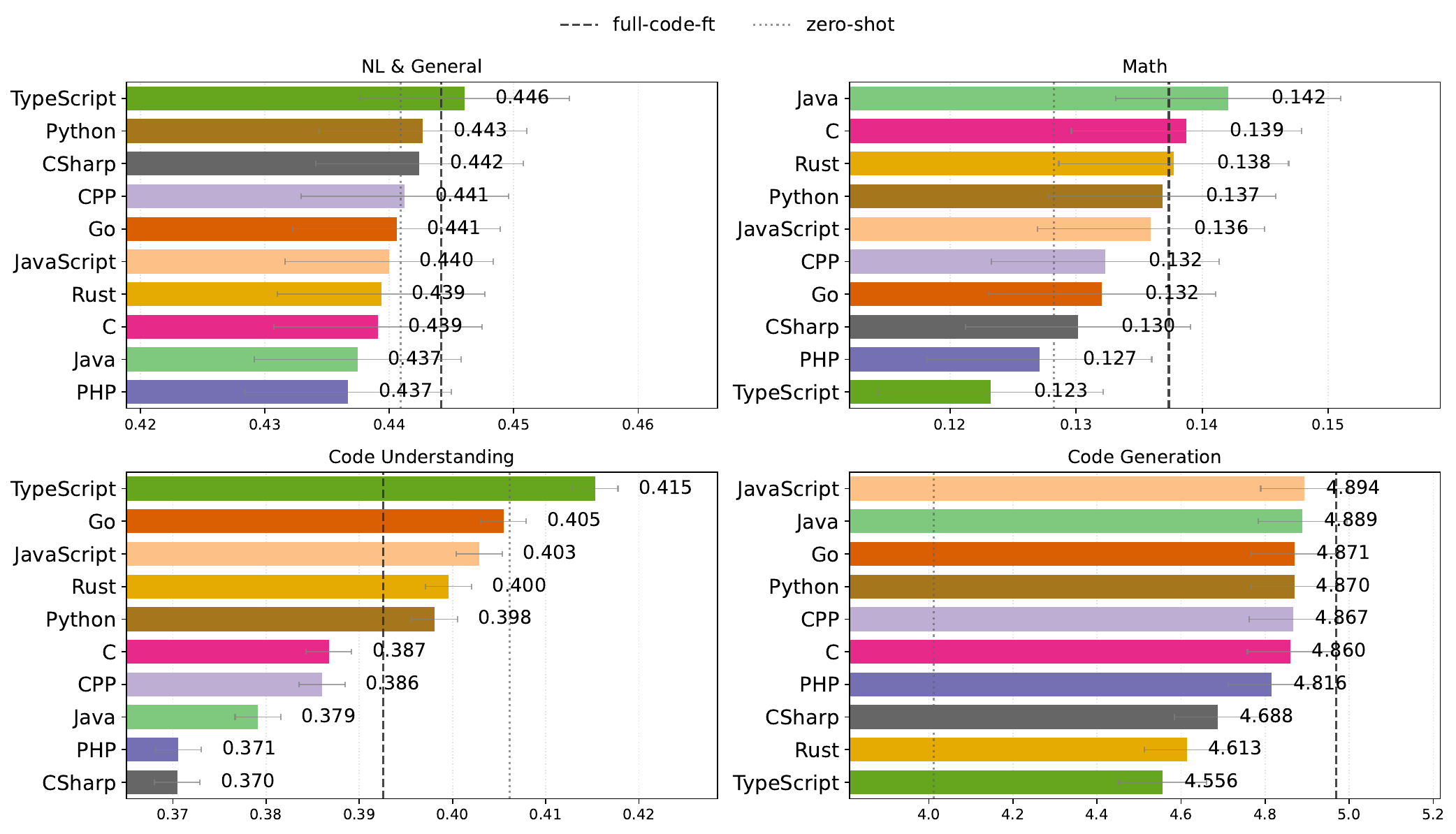}
    \caption{Per-language results}
    \label{fig:sml_rq3_langs}
  \end{subfigure}

  \caption{Performance for SmolLM2-1.7B. 
  (a) Programming language groups, (b) individual languages.}
  \label{fig:sml_rq3_combined}
\end{figure}

\paragraph{SmolLM2 model family results}
See performance of grouped performance and individual programming languages in Figure~\ref{fig:sml_rq3_combined}.

\end{document}